\documentclass[10pt,twocolumn,letterpaper]{article}

\usepackage{iccv}
\usepackage{times}
\usepackage{epsfig}
\usepackage{graphicx}
\usepackage{amsmath}
\usepackage{amssymb}
\usepackage{booktabs}
\usepackage{makecell} 
\usepackage{booktabs,multirow,array}
\usepackage[dvipsnames, svgnames, x11names]{xcolor}  
\usepackage{colortbl}
\usepackage{rotating}
\usepackage{pifont}
\usepackage{utfsym}
\usepackage{wrapfig}
\usepackage{tabularx}
\usepackage{xspace}
\usepackage{makecell}
\usepackage[font={small}]{caption}
\usepackage[hang,flushmargin]{footmisc} 
\definecolor{antiquefuchsia}{rgb}{0.57, 0.36, 0.51}

\newcommand{\best}[0]{\cellcolor{Lavender} }
\newcommand{\worst}[0]{\cellcolor{lightgray}}

\definecolor{cite}{RGB}{65,105,225}
\usepackage[breaklinks=true,bookmarks=false]{hyperref}
\usepackage{balance}

\iccvfinalcopy 

\usepackage[capitalize]{cleveref}
\crefname{section}{Sec.}{Secs.}
\Crefname{section}{Section}{Sections}
\Crefname{table}{Table}{Tables}
\crefname{table}{Tab.}{Tabs.}

\begin{document}

\title{RenderMe-360: A Large Digital Asset Library and Benchmarks Towards High-fidelity Head Avatars}

\author{
    Dongwei Pan$^{1,2}$
    \quad
    Long Zhuo$^{1*}$
    \quad
    Jingtan Piao$^{2,4*}$
    \quad
    Huiwen Luo$^{1*}$ \\
    \quad
    Wei Cheng$^{1,2*}$ 
    \quad
    Yuxin Wang$^{1*}$ 
    \quad
    Siming Fan$^{2}$ 
    \quad
    Shengqi Liu$^{2}$ \\
    \quad
    Lei Yang$^{2}$
    \quad
    Bo Dai$^{1}$
    \quad
    Ziwei Liu$^{3}$
    \quad
    Chen Change Loy$^{3}$ 
    \quad
    Chen Qian$^{1}$ \\
    \quad
    Wayne Wu$^{1}$
    \quad
    Dahua Lin$^{1,4\dagger}$
    \quad
    Kwan-Yee Lin$^{1,4\dagger}$ \\
    $^{1}$ Shanghai AI Laboratory
    \quad
    $^{2}$ SenseTime
    \quad
    $^{3}$ S-Lab, NTU
    \quad
    $^{4}$ The Chinese University of Hong Kong
    \\
    {\tt\small junyilin@cuhk.edu.hk, dhlin@ie.cuhk.edu.hk}
}

\twocolumn[{
  \renewcommand\twocolumn[1][]{#1}
  \maketitle
  \begin{center}

  \includegraphics[width=0.975\textwidth]{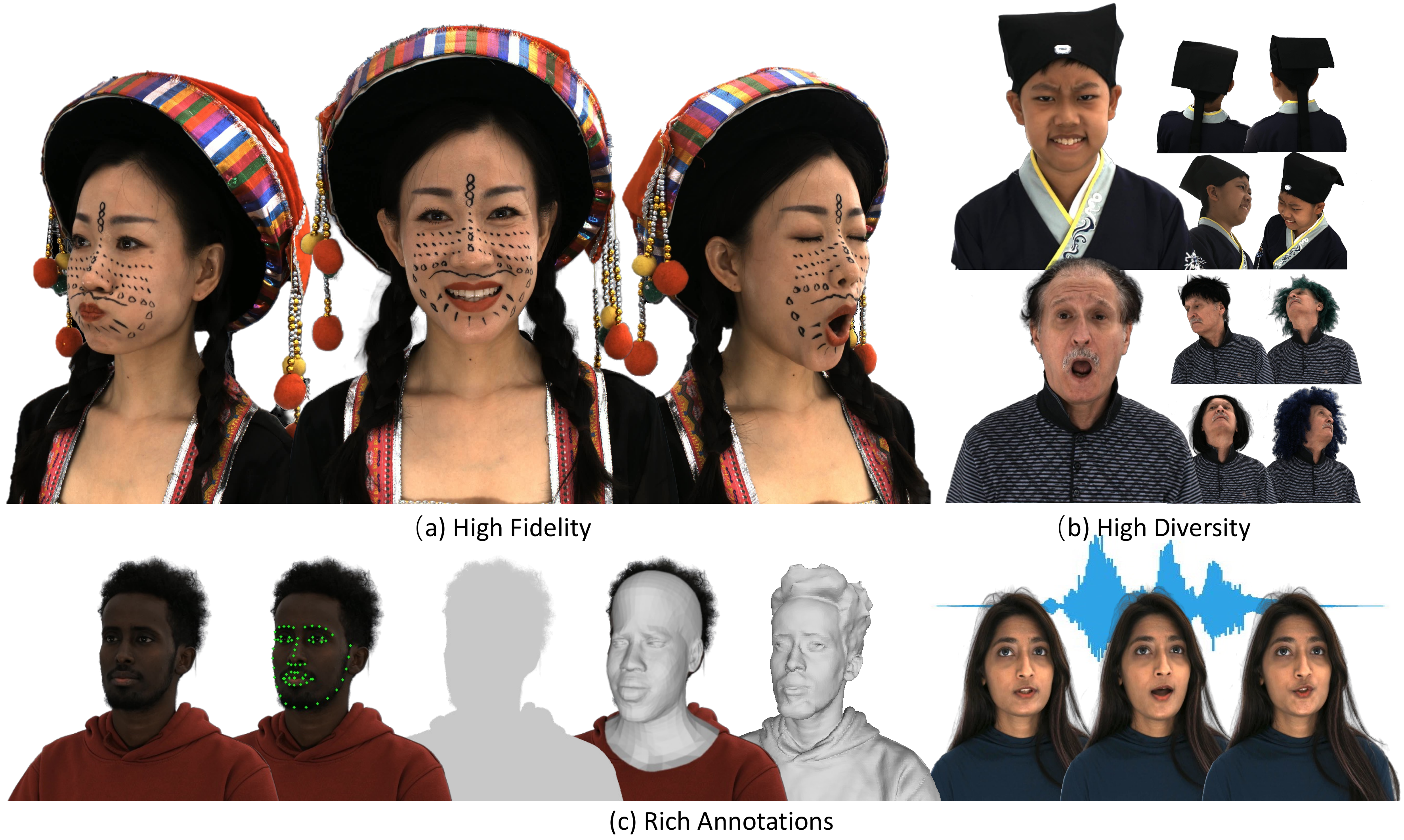}
  \vspace{-0.4cm}
  \captionsetup[figure]{hypcap=false}
  \captionof{figure}{\textbf{Overview of RenderMe-360's Core Features.} We present a large digital asset library RenderMe-360 for synthesizing high-fidelity head avatars. It has the characteristics of (a) high fidelity and (b) high diversity. Also, our dataset comes with (c) rich annotations of landmarks, matting, 3D parameterized head model, scan, and synchronized audio-video {\textit{etc}}. RenderMe-360 is proposed to facilitate the development of human head avatar related advanced research, such as head rendering and generation.}
\vspace{-0.3cm}
  \label{fig:intro-large}
  \end{center}
}]


\begin{abstract}
Synthesizing high-fidelity head avatars is a central problem for many applications on AR, VR, and Metaverse. While head avatar synthesis algorithms have advanced rapidly, the best ones still face great obstacles in real-world scenarios. One of the vital causes is the inadequate datasets  -- $1)$ current public datasets can only support researchers to explore high-fidelity head avatars in one or two task directions, such as viewpoint, head pose, hairstyle, or facial expression; $2)$ these datasets usually contain digital head assets with limited data volume, and narrow distribution over different attributes, such as expressions, ages, and accessories. In this paper, we present {\textbf{RenderMe-360}}, a comprehensive 4D human head dataset to drive advance in head avatar algorithms across different scenarios. RenderMe-360 contains massive data assets, with $243+$ million complete head frames of over $800k$  video sequences from $500$ different identities captured by synchronized HD multi-view cameras at $30$ FPS. It is a large-scale digital library for head avatars with three key attributes: 1) High Fidelity: all subjects are captured by $60$ synchronized, high-resolution $2K$ cameras to collect their portrait data in $360$ degrees. 2) High Diversity: The collected subjects vary from different ages, eras, ethnicities, and cultures, providing abundant materials with distinctive styles in appearance and geometry. Moreover, each subject is asked to perform various dynamic motions, such as expressions and head rotations, which further extend the richness of assets. 3) Rich Annotations: the dataset provides annotations with different granularities: cameras' parameters, background matting, scan, 2D as well as 3D facial landmarks, FLAME fitting, and text description.

Based on the dataset, we build a comprehensive benchmark for head avatar research, with $16$ state-of-the-art methods performed on five main tasks: novel view synthesis, novel expression synthesis, hair rendering, hair editing, and talking head generation. Our experiments uncover the strengths and weaknesses of state-of-the-art methods, showing that extra efforts are needed for them to perform in such diverse scenarios. RenderMe-360 opens the door for future exploration in modern head avatars. All of the data, code, and models will be publicly available at \url{https://RenderMe-360.github.io/}.

\end{abstract}

\section{Introduction}
Digitalizing human replicas is a perennial topic in both research and commercial communities. It serves as the foundation of many advanced applications, {\textit{e.g.,}}VR/AR, gaming, and metaverse. Among various tasks, human head avatar synthesis plays a crucial but difficult role. This is because the human head performs significant social functions with appearance, expression, speech, \textit{etc.}, in which even subtle differences between synthesized and real ones can be easily perceived by human eyes to trigger the uncanny valley effect. How to render, reconstruct, and animate a human head with realism reminds a great challenge.

Over decades, although numerous approaches have emerged and pushed forward the frontier of {\textit{facial}} reconstruction~\cite{tran2018nonlinear,tran2019learning} and animation~\cite{feng2021learning,richard2021meshtalk}, general full-{\textit{head}} level avatar synthesis~\cite{zheng2022avatar,3dmm,9577855} has only started to actively advance in recent years. Research efforts along human head avatar usually follow the flourishing of deep learning and neural rendering. Such formalizations require large-scale or dense multi-view training datasets to drive progress.

Unlike the efforts on 2D datasets~\cite{liu2015faceattributes,karras2019style}, which could utilize Internet-scale data to enhance the quantity and diversity, the path to constructing a 3D/4D repository is difficult. Thus, current human head-related datasets~\cite{wuu2022multiface,yang2020facescape,9577855,yu2020humbi,cudeiro2019capture,cosker2011facs} have significant limitations on {\textit{dataset scale}}, {\textit{sample diversity}}, {\textit{photorealism}}, {\textit{sensory modality}}, and {\textit{annotation granularity}}. For example, Multiface dataset~\cite{wuu2022multiface} contains only 13 subjects of facial data, VOCASET~\cite{cudeiro2019capture} only focuses on audio but ignores other facial functions, and HUMBI Face~\cite{yu2020humbi} has the resolution of only $2$ million pixels. The details of the limitations of the existing head-related datasets are shown in Table~\ref{tab:dataset}. These datasets are valuable to the research community, whereas they can only enable researchers to study a small set of problems. The progress of human head avatar algorithms also indicates the saturated performances on existing datasets, while the performance gap between standard datasets and real-world scenarios still remains. Moreover, human head avatar synthesis is a complex combination of many fundamental tasks (such as face/head reconstruction, expression animation, and hair modeling/animation), which requires a comprehensive digital asset library to support the exploration. In a nutshell, compared with 2D counterparts, the construction of 3D/4D human head repositories is impoverished. 

In this paper, we present \textbf{RenderMe-360}, a new publicly available large-scale 4D digital asset library with over $243$ million frames that features a wide range of downstream tasks, to boost the development of human head avatar creation. RenderMe-360 goes beyond previous datasets in several key aspects: 1) \textit{High Fidelity:} we set up a high-end data collection system, named POrtrait Large-scale hIgh-quality Capturing sYstem (POLICY), to capture high-resolution raw data of RenderMe-360. Within POLICY, all data is ensured to be captured by $60$ industrial cameras at $2448 \times 2048$ resolution (about $5$ megapixels) and $30$ FPS. 2) \textit{High Diversity:}  We collect $500$ different participants, who come from various countries with diverse ages and cultures (illustrated in Figure~\ref{fig:data_statistics}). Specifically, about $25\%$ of them are with designed makeup styles and wearing special decorations, such as ancient Chinese makeup styles with delicate hair accessories. These nature differences of the participants provide ample variety in both appearance and accent. For each subject, we capture $12$ expressions ($1$ neutral and $11$ peak), $42$ bilingual speeches, and $12$ hairstyles (at most). These collection protocols further enrich the diversity of motion, modality, and appearance. 
3) \textit{Rich Annotations:} We provide over $10$ types of annotations with different granularities (shown in Table~\ref{tab:dataset}), which ensure the compatibility of one single dataset to various tasks and methods. Specifically, we provide annotations in two levels-- per-frame annotations, and per-id annotations. The per-frame annotations refer to annotating every frame of the collected data that is tracked over capturing time. These per-frame annotations include camera parameters, matting, facial action units, and 2D/3D landmarks. The per-id annotations refer to annotating key frames for each identity in the fine-grained hierarchy, including appearance annotations, 3DMM-like models, 3D scans, and FLAME fitting, UV map, and text annotations. Our vast exploration space and massive data assets serve as the foundation to investigate the performance boundary of state-of-the-art head avatar algorithms.

Based on the proposed RenderMe-360 dataset, we set up benchmarks on five fundamental tasks, \textit{i.e.,} novel view synthesis, novel expression synthesis, hair editing, hair rendering, and talking head generation, with extensive experimental settings evaluated $16$ baseline methods (Table~\ref{tab:g-ft-capture}). We probe in detail how different factors{\footnote{For instance, how train-test distributions, training strategies, and the complexity of materials would affect novel view/expression synthesis. What data and representation can power the generalization? what is a good paradigm for hair rendering and editing?}} might introduce the influences to current baseline methods. Our experiments present many new observations, challenges, and possible new directions for the research community to catalyze future researches on the human head avatar. We hope RenderMe-360 could kickstart research efforts on related areas, and spur new opportunities not only from our formalized benchmarks, but also alternative ones that the community might come up with from our comprehensive, massive, and publicly available dataset.

\section{Related Works}
{
\newcommand{\zn}[0]{\textbf{\color{Red}N}}
\newcommand{\zs}[0]{\textbf{\color{Brown}S}}
\newcommand{\zm}[0]{\textbf{\color{SpringGreen}M}}
\newcommand{\zpf}[0]{\textbf{\color{Brown}PF}}
\newcommand{\zau}[0]{\textbf{\color{BlueViolet}AU}}
\newcommand{\cross}[0]{\color{Red}\usym{2717}}
\newcommand{\tick}[0]{\color{Green}\usym{2714}}

\begin{table*}[!t]
\setlength\tabcolsep{3pt}
\linespread{1.3}
\tiny
\begin{center}
\begin{tabular}{lccccccccccccc ccc cccccccccccc} 
\multirow{2}*{\textbf{~}} & \rotatebox[origin=lB]{80}{ID}  & \rotatebox[origin=lB]{80}{Age} & \rotatebox[origin=lB]{80}{Expression}  & \rotatebox[origin=lB]{80}{Sentence} & \rotatebox[origin=lB]{80}{Language} & \rotatebox[origin=lB]{80}{Frame}  & \rotatebox[origin=lB]{80}{Era} & \rotatebox[origin=lB]{80}{Ethnicity} & \rotatebox[origin=lB]{80}{Outfit} & \rotatebox[origin=lB]{80}{Accessory} & \rotatebox[origin=lB]{80}{HairStyle} & \rotatebox[origin=lB]{80}{Makeup} & \rotatebox[origin=lB]{80}{Motion}  & \rotatebox[origin=lB]{80}{Camera View} & \rotatebox[origin=lB]{80}{Resolution} & \rotatebox[origin=lB]{80}{FPS} & \rotatebox[origin=lB]{80}{Wig Style$\times$ Wig Color} & \rotatebox[origin=lB]{80}{Appearance Annotation} & \rotatebox[origin=lB]{80}{Phoneme-balanced Corpus} & \rotatebox[origin=lB]{80}{\zpf \space Face Lmk2d} & \rotatebox[origin=lB]{80}{\zpf \space Face Lmk3d} & \rotatebox[origin=lB]{80}{\zpf \space Matting} & \rotatebox[origin=lB]{80}{3DMM-like model} & \rotatebox[origin=lB]{80}{Scam} & \rotatebox[origin=lB]{80}{UV map} & \rotatebox[origin=lB]{80}{Text Annotation} & \rotatebox[origin=lB]{80}{\zpf  \space \zau}   \\ 
\hline
\tiny {\textbf{Dataset}} &\multicolumn{13}{|c|}{\textbf{Diversity}} & \multicolumn{3}{c|}{\textbf{Realism}} & \multicolumn{11}{c}{\textbf{Granularity}} \\
\hline
\tiny \textit{D3DFACS}~\cite{cosker2011facs}   & $10$ & - & $19$-$97$\zau & \cross & \cross & - & \cross & \cross & \cross & \cross & \cross & \cross & \tick & $6$(\zs) & \zm & $60$ & \cross & \cross & \cross & \cross & \cross & \cross & \cross & \tick & \tick & \cross & \tick  \\
\tiny \textit{HUMBI Face}~\cite{yu2020humbi}   & $772$ & (\textless $10$)-(\textgreater$60$) &  $20$ & \cross & \cross & $17.3$M & \cross & \tick & \tick & \cross & \tick & \cross & \tick & $68$  & $2$MP & $60$ & \cross & \cross & \cross & \tick & \tick & \cross & \tick & \cross & \tick & \cross & \cross      \\
\tiny \textit{Facescape}~\cite{yang2020facescape}  &  $847$ & $16$-$70$ & $20$ & \cross & \cross & $16.9$K \zm & \cross & \cross & \cross & \cross & \cross & \cross & \cross & $68$ & $4$-$12$MP & - & \cross & \cross & \cross & \tick & \cross & \cross & \tick & \tick & \tick & \cross & \cross        \\
\tiny \textit{i3DMM}~\cite{9577855}        & $64$ & $16$-$69$ & $10$ & \cross & \cross & \zn & \cross & \tick & \cross & \cross & \tick & \cross & \cross& $137$ & - & - & \cross & \cross & \cross & \cross & \cross & \cross & \cross & \tick & \cross & \cross & \cross    \\
\tiny \textit{VOCASET}~\cite{cudeiro2019capture}     & $12$ & - & \cross & $40$ & $1$ & - & \cross & \cross & \cross & \cross & \cross & \cross & \cross & $18$ & \zm & $60$ & \cross & \cross & \tick & \cross & \cross & \cross & \tick & \tick & \tick & \cross & \cross       \\
\tiny \textit{Multiface}~\cite{wuu2022multiface}    & $13$ & - & $65$/$118$ & $50$ & $1$ & $\approx15$M & \cross & \cross & \cross & \cross & \cross & \cross & \cross & $40$/$150$ & $3$MP & $30$ & \cross & \cross & \tick & \cross & \cross & \cross & \tick & \cross & \tick & \cross & \cross    \\
\hline
\tiny \textbf{\textit{RenderMe-360}} & $500$ & $8$-$80$ & $12$ & $25$/$42$ & $2$ & \textgreater$243$M & \tick & \tick & \tick & \tick & \tick & \tick & \tick & $60$ & $5$MP & $30$ & $7\times6$ & $127$ & \tick & \tick & \tick & \tick & \tick & \tick & \tick & \tick & \tick  \\
\hline
\end{tabular}
\caption{\textbf{Multi-view Head Dataset Comparison on Diversity, Realism and Granularity.} \zn: Dataset is not released, \zs: Scanner, \zm: Mesh, \zau: Action Unit, \zpf: Per-Frame. Outfit: A variety of clothes-related \& accessory-related designs. HairStyle: Manually designed or classify hairstyle, wigs with concrete style are included. Motion: head or body motion, but facial changes are not included.}
\vspace{-0.5cm}
\label{tab:dataset}
\end{center}
\end{table*}
}

\subsection{Human Head Centric Dataset} Data serve as the primary fuel to promoting the development of algorithms. Many valuable datasets are proposed for human head avatar creation, but only enable the research on a relatively small set of tasks. In contrast, RenderMe-360 is a comprehensive digital asset library that ensures the compatibility of evaluating multiple head avatar tasks in one single dataset. While there are many open-world unstructured 2D datasets ~\cite{ffhq,zhu2022celebv,liu2018celeba,huang2008labeled,kemelmacher2016megaface,parkhi2015deep} or synthetic ones, we focus on those real human heads with structured data.  

\noindent\textbf{Multi-View Head Dataset.} Collecting 3D/4D data is essential for head avatar research in both training and evaluating aspects. In the early days of computer vision, researchers mainly focused on 3D face reconstruction/tracking from data sources that included multi-view cues. In $1999$, Blanz and Vetter~\cite{DBLP:conf/siggraph/BlanzV99} used a laser scan to capture 3D faces, and proposed to model a morphable model (\textit{i.e.,}3DMM) from the database. As such a piece of equipment is not suitable for dynamic motion tracking, Zhang and Snavely~\cite{DBLP:journals/tog/ZhangSCS04} present a multi-camera active capturing system with six video cameras and two active projectors to ensure spacetime stereo capturing. Later on, Paysan {\textit{et al.}~\cite{bfm20093d} collect 3D faces by ABW-3D system. D3DFACS~\cite{cosker2011facs} introduces a dynamic 3D stereo camera system to capture 4D high-quality scans of $10$ performers with different Action Unit annotations.
Upon D3DFACS, Li {\textit{et al.}~\cite{FLAME:SiggraphAsia2017} additionally integrate 4D scans from CAESAR dataset~\cite{CAESAR} and self-captured ones (from the industrial multi-camera active stereo system -- 3dMD LLC, Atlanta). These datasets are mediocre in texture resolution and quality. Recently, Facescape~\cite{yang2020facescape} is proposed to fulfill the raw data quality, in which 3D faces are collected from a dense $68$-camera array with $847$ subjects performing specific expressions. Whereas, these research efforts are limited to supporting facial shape and expression learning.

To take a step further on modeling the entire head, Yenamandra {\textit{et al.}} propose i3DMM~\cite{9577855} dataset with $64$ subjects captured by a multi-view scanning system, called Treedys. Since Treedys is not specifically designed for head-scale capture, the authors apply post-process to capture data by cropping the head meshes based on the 3D landmarks, and removing the rest part of the upper body.  HUMBI~\cite{yu2020humbi} is a large-scale multi-view dataset, which contains different body part collections. As the systems for these two datasets are not customized to best fit head-level capture, they are limited in resolution.
Multiface~\cite{wuu2022multiface} contains head-oriented collections and detailed annotations, but only a small part of data ($13$ subjects) are publicly available. 

To facilitate multisensory modeling, VOCASET~\cite{cudeiro2019capture} is proposed. It is a 4D speech-driven scan dataset with about $29$ minutes of 4D scans and synchronized audio from $12$ speakers. Although VOCASET allows training and testing of speech-to-animation geometric models and can generalize to new data, it is limited in the extremely narrow diversity of subjects and onefold task. The other alternative is audio-visual data. These datasets are widely used in audio-visual learning tasks, like lip reading~\cite{chung2016lip, chung2017lip, DBLP:conf/bmvc/ChungZ17}, speaker detection~\cite{roth2020ava, tao2021someone} and talking head generation~\cite{wang2020mead, zhou2019talking}. For example, GRID~\cite{cooke2006audio} and MEAD~\cite{wang2020mead} are sparse multi-view datasets (four and eight respectively), which are characterized by consistent shooting conditions, carefully designed identity, and corpus distribution. However, the sparsity leaves these datasets more often be used in 2D methods.  

In contrast, our RenderMe-360, is a large-scale multi-view dataset for high-fidelity head avatar creation research. It is under a head-oriented, and high-resolution data capture environment. It contains diverse data samples (with $500$ subjects performing various activities, \textit{e.g.,} expressions, speeches, and hair motions), multi-sensory data, and rich annotations. A comparison between RenderMe-360 and other related datasets is shown in Table~\ref{tab:dataset}. 

\subsection{Neural Rendering for Head Avatar}

\noindent\textbf{Representations.} How to effectively represent and ren- der 3D scenes has been a long-term exploration of com- puter vision. The research efforts can be roughly classified into four categories at high-level: surface rendering, image-based rendering, volume rendering, and neural rendering. For surface rendering, the general idea is to first explicitly model the geometry, and then apply shading. For the geometry representation, polygonal meshes~\cite{baumgart1975polyhedron} are the most popular geometry representations for their compact and efficient nature with modern graphic engines. Other alternatives like point clouds~\cite{pfister2000surfels}, parametric surfaces~\cite{osserman2013survey}, volumetric occupancy~\cite{laine2010efficient,sitzmann2019deepvoxels}, and constructive solid geometry~\cite{foley1996computer} are less convenient. Implicit functions (\textit{e.g.,} signed distance field (SDF)) have better flexibility in complex geometry modeling. Upon these representations, researchers have proposed various shading models to render images~\cite{wang2021neus,muller2022instant,zheng2022pointavatar,mvp,sitzmann2019deepvoxels}. Whereas, all of these representations are better suited to surface reconstruction, rather than photo-realistic rendering, due to their inherent shortages in expressiveness. Traditional image-based rendering (IBR) methods~\cite{kang1998survey,shum2000review,mcmillan1995plenoptic} are texture-driven counterparts. They focus on rendering images by using representations like multi-plane images (MPI)~\cite{mildenhall2019local,tucker2020single,zhou2018stereo} or sweep plane~\cite{wang1986geometric,farias2000zsweep}. The core idea behind these representations is to leverage depth images and layers to obtain the discrete representations of light fields. Whereas, the view ranges are typically subjected to narrow view interpolations. Volume rendering ~\cite{mildenhall2021nerf,lombardi2019neural, mvp} has great ability in modeling inhomogeneous media such as clouds, and allows rendering in full viewpoints when images are dense. The core idea behind volume rendering is accumulating the information along the ray with numerically approximated of integral. With the emergence of coordinate-based neural networks, neural rendering pops up and becomes a powerful complementarity of classic representations. Such a methodology combines the advantages of differential rendering and neural networks.  For instance, neural surface rendering~\cite{yariv2020multiview,kato2020differentiable,wang2021neus}, and neural volume rendering~\cite{mildenhall2021nerf,lombardi2019neural} ensure novel views of the target scene can be rendered by arbitrary camera pose trained by dense multi-view images. These methods achieve photo-realistic rendering and smooth view transition results in creating free-viewpoint videos compared to traditional ones. The follow-up researches lie on the directions of model efficiency~\cite{muller2022instant,sun2022direct,yu2021plenoctrees}, dynamic scene~\cite{pumarola2021d,tretschk2021non,gao2021dynamic}, large-scene compatibility~\cite{turki2022mega,xiangli2021citynerf}, class-specific robustness~\cite{hong2022headnerf,wang2022morf}, multi-modal extensiveness~\cite{guo2021ad,wang2022clip}, or generalizablitiy~\cite{yu2021pixelnerf,wang2021ibrnet,chen2021mvsnerf,mihajlovic2022keypointnerf,cheng2022generalizable,lin2023visionnerf}.

\noindent\textbf{Head Rendering.} Researchers usually emphasize utilizing head or face priors to condition the neural fields. Such a philosophy can help either improve the robustness or create controllable avatars. 
Priors like parametric model~\cite{blanz2003face,FLAME:SiggraphAsia2017} coefficients, key points, and explicit surface mesh/point clouds are popular ones to be integrated into the framework. For example, 
NHA~\cite{grassal2022neural} presents a framework to learn vertex offsets and attached textures from fitted FLAME surface via coordinate-based MLPs embedded on the surface.  NerFACE~\cite{Gafni_2021_CVPR} and IM Avatar~\cite{zheng2022avatar} use FLAME~\cite{FLAME:SiggraphAsia2017} model expression coefficient to condition the neural field and learn to create an animatable head avatar from monocular video. MofaNeRF~\cite{zhuang2022mofanerf} takes similar inspiration, while only focusing on the face region. Taking multi-view images as input condition, Neural Volume~\cite{lombardi2019neural} models dynamic 3D content with a volumetric representation. MVP~\cite{mvp} replaces the single volume with a mixture of multiple predefined volumetric primitives which improves the resolution and efficiency of volume rendering. To increase the flexibility of re-rendering the avatar in new environments (\textit{e.g.,} novel expression and lighting), PointAvatar~\cite{zheng2022pointavatar} presents a paradigm of utilizing point-based representation which achieves fast model convergence by coarse-to-fine optimization. For generalization{\footnote{There is another trend for generalization, which leverages the power of both neural radiance fields and deep generative models~\cite{chan2022efficient,hong2022eva3d,gu2021stylenerf}. On the one hand, 3D-aware mechanisms could help toward lifting 2D to 3D. On the other hand, deep generative models allow large-scale head dataset~\cite{hong2022headnerf}, synthetic data~\cite{galanakis20233dmm}, or in-the-wild data~\cite{ffhq,liu2018celeba} could be utilized to increase diversity. We leave this direction for future discussion.}} KeypointNeRF~\cite{mihajlovic2022keypointnerf} synthesizes free viewpoints of human heads via multi-view image features and 3D keypoints. In addition, some researchers use cross-domain data such as audio or text to condition the neural fields. For instance, ADNeRF~\cite{guo2021ad} presents a 3D-aware alternative to the 2D talking face pipelines (unfolded in Section~\ref{ghm}) by conditioning the radiance field with both head poses and audio fragments. 

\noindent\textbf{Hair Reconstruction.} High-fidelity hair reconstruction has been a long-standing challenging task since the early age of computer vision and graphics. Human hair is difficult to render due to its tremendous volume of strands, great diversity among different identities, and micro-scale structure. Dynamic hair rendering and animation are even more difficult, since complex motion patterns and self-occlusions need to be additionally considered. When the underlying hair geometry is known, classic hair modeling paradigms like Kajiya-Kay~\cite{kajiya1989rendering}, Modified Marschner~\cite{marschner2003light,d2013importance}, and Double Cylinder shading models~\cite{DBLP:journals/tog/YanTJR15} provide the foundation of hair rendering. Most of the time in real-world scenarios, reconstructing hair (both geometry and color) from multi-view images is expected. A naive solution is applying classic multi-view stereo methods (\textit{e.g.,} COLMAP~\cite{schoenberger2016sfm}). Whereas, the results are usually coarse and noisy. For dynamic motion animation, various physics-based simulations~\cite{DBLP:conf/sca/KugelstadtS16,DBLP:journals/tog/JiangGT17,DBLP:journals/tog/Daviet20} are proposed to solve the problem via different hair collision assumptions. Some later research efforts utilize deep neural networks to extract temporal features of hair motion~\cite{DBLP:journals/tog/YangSZZ19}, infer 3D geometry~\cite{DBLP:conf/siggrapha/KuangCF0Z22}, or localize valid mask region~\cite{2021human_hair_inverse_rendering}. With the blooming of neural rendering, recent works make notable progress in both static and dynamic hair reconstruction. For example, to render the high-fidelity hair strand in real-time, Neural Strand~\cite{rosu2022neuralstrands} introduces a neural rendering framework for jointly modeling both hair geometry and appearance. For dynamic hair modeling, general dynamic scene rendering methods such as ~\cite{2021nsff, 2021nrnerf,
lombardi2019neural, mvp, wang2022hvh} could be directly applied to the task. These methods have been proven as powerful tools to model the motion and interaction of hair strands. Upon the~\cite{mvp}, HVH~\cite{wang2022hvh} designs a special volumetric representation for hair, and models the dynamic hair strands as the motion of the volumetric primitives.

\subsection{Generative Models for Head Manipulation}\label{ghm}

\noindent\textbf{Hair Editing.} In addition to hair rendering and animation, finding a neat solution to support hairstyle or hair color editing is also an exciting research problem. Related methods could be categorized into image-based editing~\cite{tan2020michigan,xiao2021sketchhairsalon, saha2021LOHO} and text-based editing~\cite{wei2022hairclip,Patashnik_2021_ICCV}. 
The general ideas behind the two trends follow a similar pipeline -- $(1)$ first, encode hair appearance, shape, and structure information from prompts. For image-based methods~\cite{DBLP:conf/cvpr/Lee0W020,xiao2021sketchhairsalon, DBLP:journals/tog/TanCC0CYTY20}, the prompts could be masks, well-drawn sketches, or reference images. For text-driven ones, the core prompt is text descriptions. $(2)$ The second step is style mapping, in which the input conditions are mapped into the corresponding latent code changes. Image-based methods utilize sophisticated conditional generative module~\cite{tan2020michigan, xiao2021sketchhairsalon} or modulate conditions into the prior space of a pre-trained generative model~\cite{saha2021LOHO}( \textit{e.g.,} StyleGANv2~\cite{Karras2019stylegan2}) via inversion strategies (\textit{e.g.,} ~e2e\cite{tov2021designing}, PTI~\cite{roich2021pivotal}, ReStyle~\cite{alaluf2021restyle}, and HyperStyle~\cite{alaluf2021hyperstyle}). As a more flexible complementarity, text-driven methods graft the power of CLIP~\cite{radford2021learning} to guide/regularize target attribute manipulation. StyleCLIP~\cite{Patashnik_2021_ICCV} is a general text-driven image manipulation framework and can be directly applied to hair editing. It provides a basic solution to tailor text information into latent optimization and mapper. Following StyleCLIP, HairCLIP~\cite{wei2022hairclip} designed specific latent mappers for hairstyle and hair color editing based on both reference images and text prompts. 

\begin{figure}[t]
    \centering
    \includegraphics[width=1.0\linewidth]{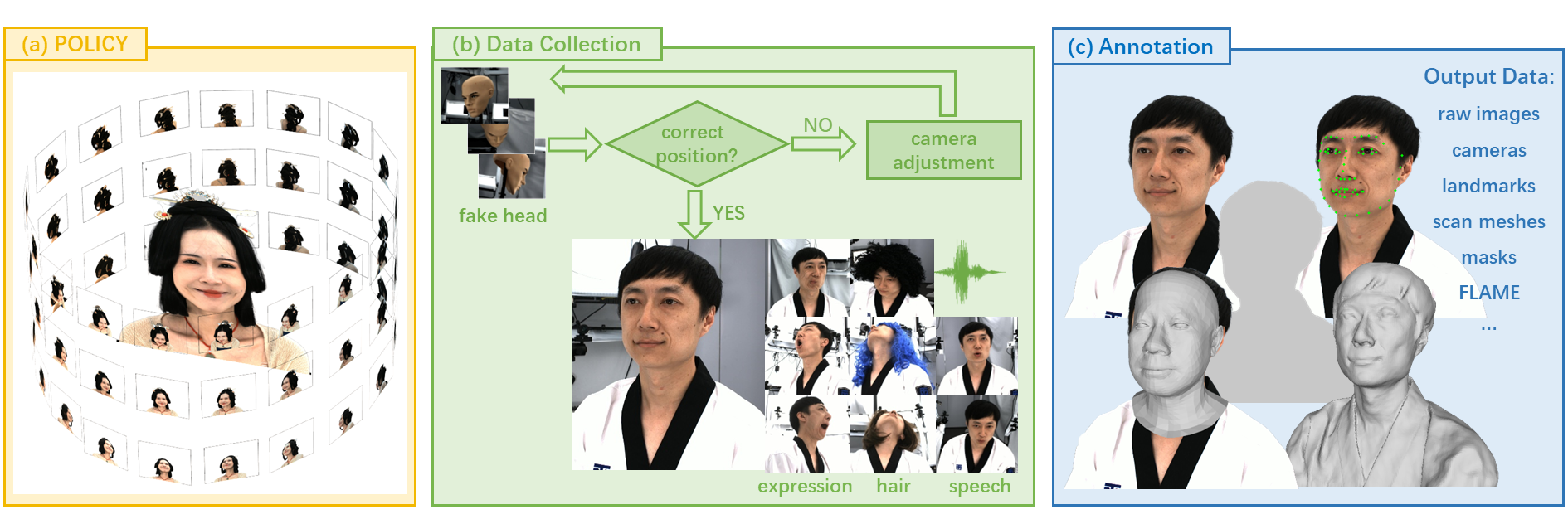}
    \caption{\textbf{Overview of Our Data Collection Pipeline.} Our system captures subjects in 60 different views. Camera adjustment is completed per day to ensure human's head is always at image's center. There are three different kinds of video sequences captured, which are expressions, wigs and speeches. Except for raw images and audio, we also provide rich annotations for future tasks.}
    \label{fig:process}
    \vspace{-0.5cm}
\end{figure}

\noindent\textbf{Talking Head Generation.} This task also known as face reenactment, aims to synthesize realistic human face videos according to the given source facial clips and the driving materials. The face animation task can be divided into two categories by the driving modality: image-driven face animation~\cite{thies2016face2face, wu2018reenactgan, siarohin2019first, zakharov2020fast, burkov2020neural, wang2021one, hong2022depth} and audio-driven face animation~\cite{suwajanakorn2017synthesizing, thies2020neural, jamaludin2019you, vougioukas2019realistic, chen2019hierarchical, zhou2019talking, Zhu2020ArbitraryTF, ji2021audio-driven, guo2021ad, liu2022semantic}. 
The major challenge for this task is to control the expressions and head pose of the synthesized video according to the driving materials while reserving the identity information of the source images. For the expression control, several methods used facial landmarks~\cite{siarohin2019first, zakharov2020fast}, latent feature space~\cite{jamaludin2019you, zhou2019talking} or the parameters of the parametric head model~\cite{thies2016face2face, thies2020neural} to model the facial expressions, and then use these intermediate representations to guide the animated video generation. For the pose control, some methods first predicted the pose representations from the given images as pose descriptor~\cite{burkov2020neural}, depth map~\cite{hong2022depth}, or the explicit rotation and translation matrix~\cite{wang2021one}.
More recently, AD-NeRF~\cite{guo2021ad} and SSP-NeRF~\cite{liu2022semantic} condition the radiance field with audio fragments for the customized talking head generation. 
AD-NeRF trains two neural radiance fields for inconsistent movements between the head and torso without an explicit 3D face model, which is relatively efficient and accessible compared to the methods based on the 3D head model. SSP-NeRF~\cite{liu2022semantic} uses one unified neural radiance field for portrait generation with the introduced torso deformation module and semantic-aware ray sampling strategy. 

\section{RenderMe-360} 
\begin{figure}
    \centering
    \includegraphics[width=1.0\linewidth]{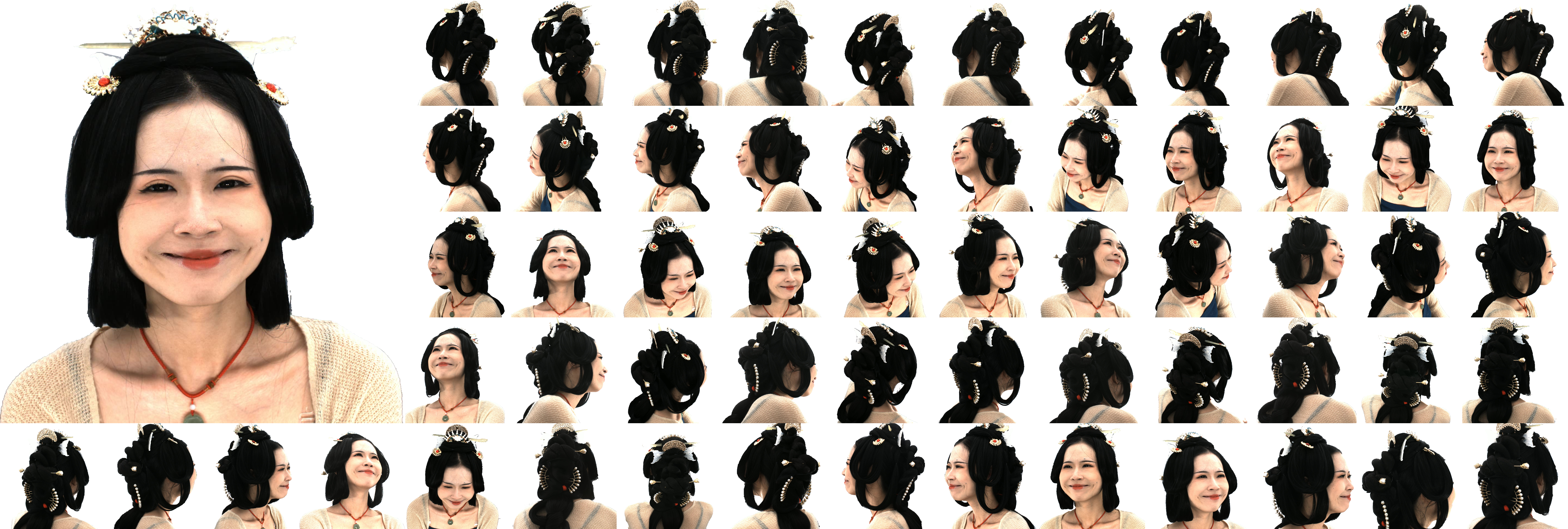}
    \caption{\textbf{Multi-view Head Data Sample.} The captured human head visual data encompass $60$ camera views with $360^\circ$ left-to-right, and $160^\circ$  up-to-down. }
    \label{fig:multi-view}
    \vspace{-0.5cm}
\end{figure}

In this section, we introduce RenderMe-360 dataset in detail. We start with the description of our capture system (Section~\ref{section:POLICY}), and move on to an overview process of the data collection (Section~\ref{section:data_collection}). Then, we present the data annotation pipeline (Section~\ref{section:data_annotation}). The whole process is visualized in Figure~\ref{fig:process}. {\textbf{Please also refer to our project page for a more vivid visual experience of data collection quality, and grasp the key features of this work.}

\subsection{Capture System} 
\label{section:POLICY}

As illustrated in Figure~\ref{fig:process}(a), we build a multi-video camera system, named POLICY, to record synchronized multi-view videos of human head performance. It contains $60$ industry cameras and covers a field of view of $360^\circ$ left-to-right and over $160^\circ$ up-to-down for video capture at the whole-head level (as shown in Figure~\ref{fig:multi-view}). To ensure encompassing fine details (\textit{e.g.,} hair strands, wrinkles, and freckles), we choose cameras with a high resolution at $2448 \times 2048$. The shutter speed of each camera is $30$ FPS to capture fine-grained motion changes. To capture multi-sensory information, a condenser microphone is collocated with the camera system, and the audio-vision synchronization is at the speed of $30$ Hz. Our POLICY achieves high-bandwidth data capturing with a speed of $90$ GB/s. Please refer to Appendix~\ref{sec:policy} for more system details.

\subsection{Data Collection}
\label{section:data_collection}
The data collection pipeline is illustrated in Figure~\ref{fig:process}(b). Specifically, to guarantee the valid rate of captured data, we first apply a trial collection to check on the operability of equipment and adjustment of camera positions before formal acquisition. We use a fake head as a recording target at the trial.  After this, the formal capture process starts, which consists of the following parts for per-person recording:

\noindent
1) \textit{Calibration Capture.} We capture camera calibration data before every round of recording. We use a chessboard and move it in front of the cameras at a fixed-order trajectory.

\noindent
2) \textit{Expression Capture.} We ask each subject to perform the same expression set, which includes $12$ distinctive facial expressions ($1$ natural and $11$ exaggerated expressions, as shown in Figure~\ref{fig:exp_capture} in the Appendix) defined in~\cite{9577855}. 

\noindent
3) \textit{Hair Capture.} To cover diverse hair materials and hair motions, we record more than $10$ video sequences (as illustrated in Figure~\ref{fig:hair_capture} in the Appendix) for each \textit{normal} subject, with different hairstyles under three levels --original hair, headgear, and wig captures. Specifically, the collected data includes one motion sequence for the subject's original hair, one for headgear that hides one's hair, and ten sequences for wearing different wigs with random styles and colors. In the wig collection step, we ask each subject to perform head rotation in a wide range in order to capture the large hair movement. For performers who originally wear unique hair accessories or hairstyles, the wig recordings are skipped. 

\noindent
4) \textit{Speech Capture.} We provide a rich corpus for the performers, which encompasses single words combined sentences, phonetically balanced protocols, and short paragraphs in two languages (Mandarin and English). For each subject, we randomly pick materials from the corpus and ask the subject to speak $25$ to $42$ phonetically balanced sentences. We do not require a standard mouthpiece but mispronunciation is not allowed. 

As shown in Figure~\ref{fig:data_statistics}, after the above processes finished, we obtain a large-scale dataset of over $800k$ recording videos from $500$ identities,  which is gender-balanced, includes multiple ethnicities ($217$ Asian, $140$ White, $88$ Black, and $55$ Brown), and spans ages from $8$ to $80$ with approximate normal distribution where teenagers and adults form the major part. More detailed description of our data collection process and related data statistics are discussed in Section~\ref{sec:colloction} and ~\ref{sec:statiscs} in the Appendix.

\begin{figure}
    \centering
    \includegraphics[width=1.0\linewidth]{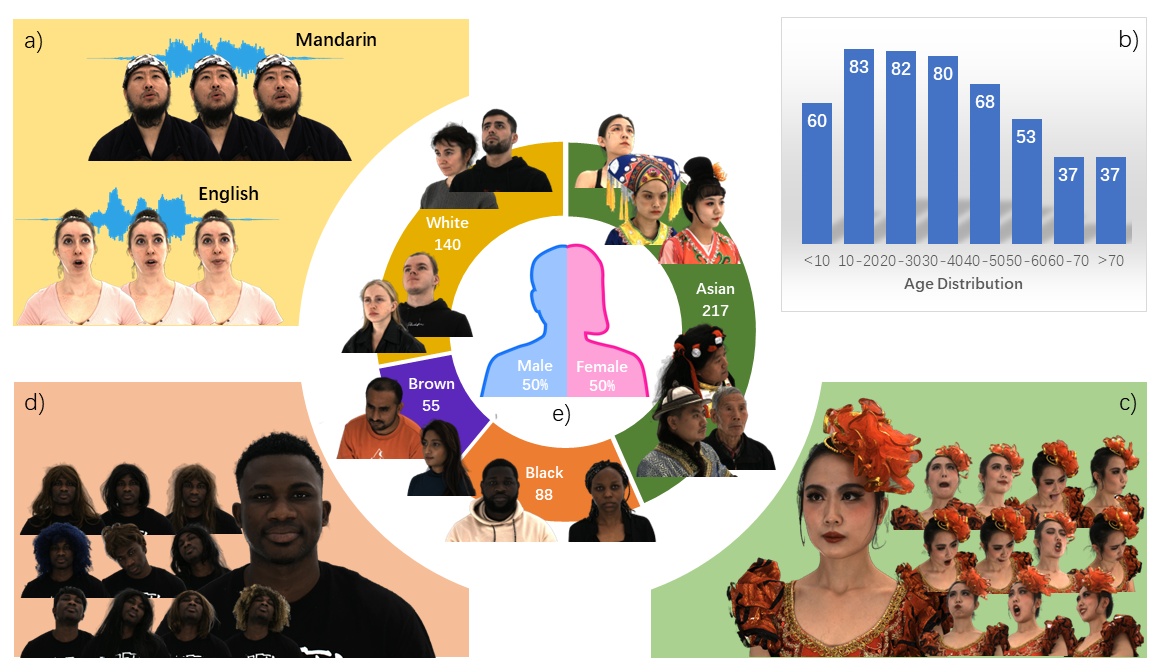}
    \caption{\textbf{Key Data Statistics.} a) $25$ to $42$ speeches are recorded per subject, Chinese sentences are designed for Chinese people, while others speak English. b) Age Distribution. c) $12$ expressions are captured for each subject. d) $8$ to $12$ wigs are randomly sampled and captured for subjects without head accessories. e) Distribution of gender and ethnicity. }
    \vspace{-0.5cm}
    \label{fig:data_statistics}
\end{figure}

\subsection{Data Annotation}
\label{section:data_annotation}

\begin{figure}
    \centering
    \includegraphics[width=1.0\linewidth]{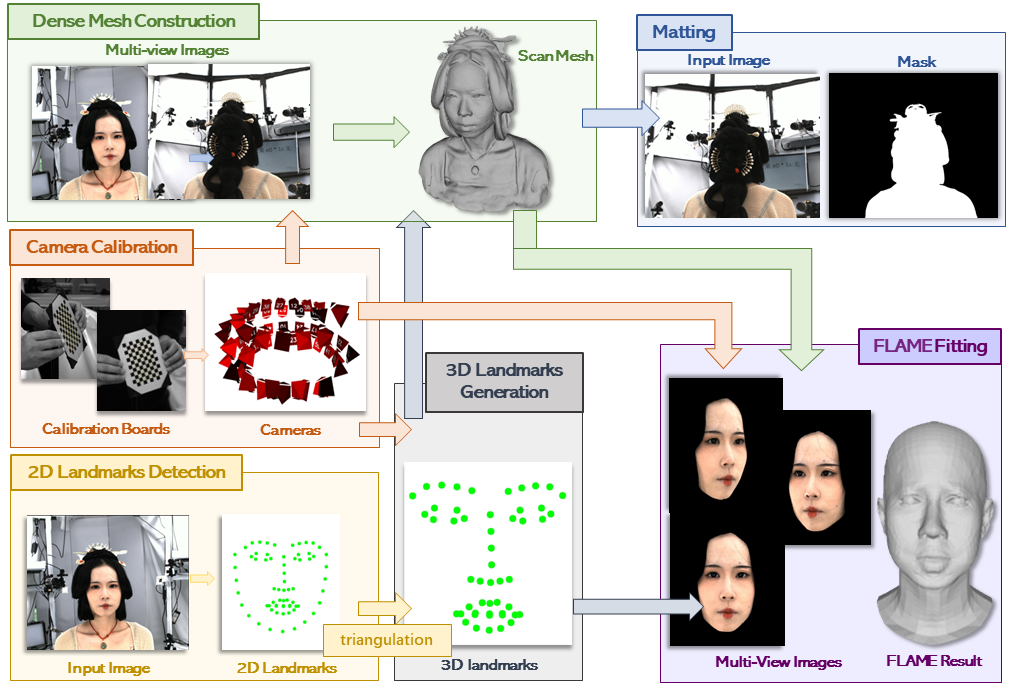}
    \caption{\textbf{Annotation Toolbox Pipeline Overview.} Camera calibration and 2D landmarks detection are processed in parallel. Based on calibrated cameras from multiple views, triangularization is applied to 2D landmarks to get 3D landmarks. Dense mesh reconstruction is supplied for better matting and FLAME results, calibrated cameras and 3D landmarks are taken as inputs to output scan mesh here. At last, all data generated before is well-prepared for FLAME fitting, and matting is taken scan mesh for refinement.}
    \label{fig:anno_pipeline}
    \vspace{-0.3cm}
\end{figure}

In addition to large-scale datasets, diverse and multi-granularity head-related annotations are also crucial for the research of human head avatar tasks. However, there is still a deficiency of an all-around head dataset with rich annotation in the research community. To facilitate the development of downstream tasks, we provide rich annotations on the captured data -- camera parameters, 2D/3D landmarks, dense mesh reconstruction, FLAME fitting, matting, and detailed text description that encompasses facial attributes and accessory materials. We also provide a toolbox to  automatically label most of the annotations, as shown in Figure~\ref{fig:anno_pipeline}. We unfold the key information in this section. For more details, please refer to Section~\ref{sec:annote} in Appendix.

\noindent\textbf{Camera Parameters.} We estimate the extrinsic matrix and rectify the intrinsic matrix for each camera via a fine calibration pipeline ~\cite{xrprimer}. The process includes chessboard detection, intrinsic calibration, and extrinsic calibration with multi-view bundle adjustment. To ensure the quality of calibrated data, we additionally applied fast novel view synthesis via Instant-NGP~\cite{muller2022instant}, and facial landmark reprojection on multi-view single-frame to eliminate unqualified estimation on camera parameters. Then, we loop the pipeline in everyday frequency to obtain fine estimations. 

\noindent\textbf{2D \& 3D Facial Landmarks.} 2D landmarks are detected per frame via an enhanced version of ~\cite{wu2018look} on selected frontal views, which range from $60^\circ$ left to $60^\circ$ right. With calibrated cameras and 2D landmarks from multiple views, RANSAC~\cite{ransac} triangulation is applied to obtain 3D landmarks. In order to guarantee the accuracy of 3D landmarks, low-quality 2D ones are filtered out with spatial and temporal constraints, and 3D results with large re-projection errors are also filtered out. For the frames that are neither precisely calculated on 2D landmarks nor 3D landmarks, we manually label the 2D landmarks and re-run the triangulation.  

\noindent\textbf{Dense Mesh Reconstruction.} Traditional MultViewStereo algorithms based on feature points extraction and geometric optimization, such as~\cite{wang2020patchmatchnet}, can only generate irregular point clouds, and have low-quality results in areas of texture missing, such as black hair and dark skin. Therefore,  we additionally apply NeuS \cite{wang2021neus}, which uses neural representation for signed-distance-function and optimizes with surface-based rendering results to do multi-view reconstruction and dense mesh extraction. For video sequences, the first frame is optimized from scratch, then the following frames are fine-tuned on the optimized neural representation to accelerate the speed of convergence.

\noindent\textbf{Matting.} Reasonable foreground segmentation for human heads is challenging. Since diverse hairstyles and accessories form the long-tail problem. Therefore, we develop a united pipeline that combines video-based matting and scan mesh information to improve the performance of matting. Specifically, we capture the background prior to each round of recordings, and apply RVM~\cite{DBLP:conf/wacv/LinYSS22}, a video-based convolutional neural network, to estimate the rough matting result in the first step. As it cannot handle detailed accessories and white boundaries between hats or clothes and the background, we additionally blend the depth-aware mask via Z-buffer during rasterization \cite{liu2019soft} on scanned mesh to improve the matting quality. With multi-view information, the background ambiguity can be distinguished from other views. We use Gaussian Mixture Model to blend the estimations from two models for each pixel\cite{rother2004grabcut}, since the former may lack generalization ability and the latter will be misaligned due to geometric errors. For extremely hard cases where both steps cannot output satisfactory results, we add human-in-the-loop to manually label the boundary.

\noindent\textbf{FLAME Fitting.} We apply a face parsing model based on ~\cite{BiSeNet} to get the face mask, in order to mainly focus on the facial region during fitting. We then take camera matrices, 2D/3D landmarks, scan mesh, and pixel information as inputs to generate FLAME models. Since only keyframes are attached with scans to save processing costs, we use two fitting methods in practice -- one is fitted with scan mesh, and the other is not. For frames with scan mesh, we use 3D landmarks to initialize FLAME parameters and optimize on point-to-point distance via ICP~\cite{arun1987least}. Only the face region is added to the optimization with a face mask from multi-view segmentation. Since scan geometry shows accurate facial shape in world space, fitting with it preserves better facial contour. For the ones without scan mesh, inspired by ~\cite{FLAME:SiggraphAsia2017}, personalization is designed for each subject. Specifically, the key idea is that we select neutral frames with corresponding scan geometries, and average across these fitting results to generate a personalized template for each subject. We keep the shape parameters constant during fitting the expression sequences which lack scans.

\noindent\textbf{Text Annotation.} To facilitate multi-modal research on human head avatars, we provide text descriptions for the captured videos at unprecedented granularity. These descriptions cover both static and dynamic attributes from four major aspects: $1)$  {\textit{static facial features}} of the subjects, where over $90$ attributes at general facial appearance, detailed appearance, and lighting condition levels are described; $2)$ {\textit{static information of non-facial regions}}, where the texture, material, and shape attributes of subject's accessories (such as necklace, earrings, and hairpin) and hairstyle are defined; $3)$ {\textit{dynamic facial actions}}, fine-grained action units (AUs) descriptions based on the FACs system\cite{ekman1978facial} are given; $4)$ \textit{dynamic video activity descriptions}, where full-sentence annotations of global action sequence descriptions for each captured video are provided.

\section{Benchmark}
{
\newcommand{\zmi}[0]{\textbf{\color{OrangeRed}M}}
\newcommand{\zsi}[0]{\textbf{\color{PineGreen}S}}
\newcommand{\zcc}[0]{\textbf{\color{RedViolet}C}}
\newcommand{\zv}[0]{\textbf{\color{Bittersweet}V}}
\newcommand{\zs}[0]{\textbf{\color{BlueViolet}S}}
\newcommand{\zi}[0]{\textbf{\color{LimeGreen}I}}
\newcommand{\zl}[0]{\textbf{\color{Melon}L}}
\newcommand{\zp}[0]{\textbf{\color{Purple}P}}
\newcommand{\zkp}[0]{\textbf{\color{Brown}K}}
\newcommand{\zk}[0]{\textbf{\color{Periwinkle}K}}
\newcommand{\zn}[0]{\textbf{\color{Brown}N}}
\newcommand{\zf}[0]{\textbf{\color{Blue}F}}
\newcommand{\zr}[0]{\textbf{\color{Apricot}R}}
\newcommand{\zc}[0]{\textbf{\color{Salmon}S}} 
\newcommand{\zg}[0]{\textbf{\color{SpringGreen}G}}
\newcommand{\zst}[0]{\textbf{\color{CadetBlue}S}}
\newcommand{\zdy}[0]{\textbf{\color{Brown}D}}
\newcommand{\zfe}[0]{\textbf{\color{Brown}F}}
\newcommand{\zpb}[0]{\textbf{\color{Melon}P}}
\newcommand{\cmark}{\ding{51}}%
\newcommand{\xmark}{\ding{55}}

\begin{table}[!t]
\vspace{-4ex}
\setlength\tabcolsep{1pt}
\linespread{1.2}
\begin{center}
\small
\resizebox{0.475\textwidth}{!}{
\begin{tabular}{llccccccc}
Task & Method & \rotatebox[origin=lB]{80}{Required Data} & \rotatebox[origin=lB]{80}{Representation} & \rotatebox[origin=lB]{80}{Static/Dynamic} & \rotatebox[origin=lB]{80}{Conditioning} & \rotatebox[origin=lB]{80}{Face Priors} & \rotatebox[origin=lB]{80}{3D Consistency} & \rotatebox[origin=lB]{80}{Generalizability}\\
\hline
\multirow{7}{*}{\begin{tabular}{l}  Novel View \\ Synthesis \end{tabular}} & 
Instant-NGP~\cite{muller2022instant}      & \zmi+\zcc & \zv & \zst & \zi & \xmark & \zr & \zc \\
~ & NeuS~\cite{wang2021neus}         & \zmi+\zcc & \zs & \zst & \zi & \xmark & \zf & \zc \\
~ & NV~\cite{lombardi2019neural}           & \zmi+\zcc & \zv & \zdy & \zi & \xmark & \zf & \zc \\ 
~ & MVP~\cite{mvp}          & \zmi+\zcc & \zv & \zdy & \zi & \zp    & \zr & \zc \\ 
~ & IBRNet~\cite{wang2021ibrnet}      & \zmi+\zcc      & \zv & \zst & \zi & \xmark    & \zr & \zg \\
~ & KeypointNeRF~\cite{mihajlovic2022keypointnerf}    & \zmi+\zcc & \zv & \zst & \zi & \zkp    & \zr & \zg \\
~ & VisionNeRF~\cite{lin2023visionnerf}    & \zmi+\zcc   & \zv & \zst & \zi & \xmark    & \zr & \zg \\

\hline
\multirow{3}{*}{\begin{tabular}{l} Novel Expression \\ Synthesis \end{tabular}} & 
    NeRFace~\cite{Gafni_2021_CVPR}      & \zsi+\zcc & \zv & \zdy & \zl & \zp    & \zr & \zc \\
~ & IM Avatar~\cite{zheng2022avatar}    & \zsi+\zcc & \zv & \zdy & \zl & \zp    & \zr & \zc \\
~ & PointAvatar~\cite{zheng2022pointavatar}    & \zsi+\zcc & \zpb & \zdy & \zl & \zp    & \zr & \zc \\

\hline
\multirow{2}{*}{\begin{tabular}{l} Hair Rendering \end{tabular}} & 
NSFF~\cite{2021nsff}        & \zmi+\zcc & \zv & \zdy & \zi & \xmark & \zr & \zc \\
~ & NR-NeRF~\cite{2021nrnerf}     & \zmi+\zcc & \zv & \zdy & \zi & \xmark & \zr & \zc \\

\hline
\multirow{2}{*}{\begin{tabular}{l} Hair Editing \end{tabular}} & 
HairCLIP~\cite{wei2022hairclip}        & \zsi & \zfe & \zst & \zl & \zfe & \zn & \zg \\
~ & StyleCLIP~\cite{Patashnik_2021_ICCV}     & \zsi & \zfe & \zst & \zl & \zfe & \zn & \zg \\

\hline
\multirow{2}{*}{\begin{tabular}{l} Talking Head \end{tabular}} & 
ADNeRF~\cite{guo2021ad}          & \zsi+\zcc & \zv & \zdy & \zl & \zp & \zr & \zc \\
~ & SSPNeRF~\cite{liu2022semantic}          & \zsi+\zcc & \zv & \zdy & \zl & \zp & \zr & \zc \\
\hline

\end{tabular}
}
\end{center}
\vspace{-1.5ex}
\caption{\small{\textbf{Methods for RenderMe-360 Benchmarks.} We construct the benchmark with five vital tasks, the underlying attributes are also listed. 
\zmi: Multi-view images,: \zsi: Single-view images, \zcc: Camera Calibration, 
\zv: Neural Volumetric, \zs: Neural SDF, \zpb: Point-based Representation, \zfe: Feature Space,
\zst: Static, \zdy: Dynamic,
\zi: Images Conditioning, \zl: Latent Codes,
\zp: Parametric Models \zkp: Face Keypoints,  
\zr: Radiance Field-based, \zf: SDF-based, \zn: Convolution-based, 
\zc: Case specific, \zg: Generalizable. }}
\vspace{-0.5cm}
\label{tab:g-ft-capture}
\end{table}
}

As our RenderMe-360 dataset is a large-scale, diverse, all-round, and multi-granularity head-centric digital repository, it provides various potentials in new research directions and applications for human avatar creation. 
Here, we build a comprehensive benchmark upon RenderMe-360 dataset, with $16$ representative methods on five vital tasks of human head avatars (summarized in Table~\ref{tab:g-ft-capture}). These tasks range from static head reconstruction and dynamic synthesis, to generation and editing. For each task, we set up several experimental protocols to probe the performance limits of current state-of-the-art methods under different settings. We present the key insights in the main paper, and highlight the best/worst results in lavender/gray colors. Please refer to Section~\ref{sec:benchmark} in Appendix for more implementation details, experiments, qualitative/quantitative results, and discussions.

\subsection{Novel View Synthesis}
Here, we present novel view synthesis (NVS) benchmark of both case-specific (\textit{i.e.,}Single-ID NVS in sub-section~\ref{sing-nvs}) and generalizable (sub-section~\ref{ID nvs}) tracks. 

\label{section:Novel View Synthesis Benchmar}
\subsubsection{Single-ID NVS}\label{sing-nvs}

This case-specific track refers to the setting of training on a single head with multi-view images, which originates from NeRF~\cite{mildenhall2020nerf}'s de facto setting, to evaluate the robustness of {\textit{static}} multi-view head reconstruction. We study four representative methods with two protocols -- $1)${\textit{\#Protocol-1}} for exploring methods' robustness to different appearance or geometry factors. The dataset is split into three categories, \textit{i.e.,} Normal Case, With Deformable Accessory, and With Complex Accessory, according to the complexity of appearance and geometry. We discuss the protocol in the main paper; $2)${\textit{\#Protocol-2}} for probing methods' robustness to different camera number and distributions. We discuss this protocol in Section~\ref{camera ablation-single} in the Appendix. 

\noindent
\textbf{Settings.} We select $20$ identities from the three categories to evaluate the methods. For the training-testing split, we uniformly sample $22$ views from all $60$ views as test views, and use the rest camera views to train each model. A visualization of camera distribution is shown in Figure~\ref{fig:case-specific-novel-view} in the Appendix, noted as $Cam 1$. The four methods for comparison are: Instant-NGP~\cite{muller2022instant}, NeuS~\cite{wang2021neus}, NV~\cite{lombardi2019neural} and MVP~\cite{lombardi2021mixture}. The first three methods
are originally designed for general-purpose case-specific NVS, and the last one is designed for human head avatar reconstruction. We compute PSNR, SSIM, and LPIPS for rendered novel view images against ground-truth. The quantitative results are listed in Table~\ref{tab:case-specific-novel-view}, and the qualitative ones are listed in Figure~\ref{fig:case-specific-novel-view} in the Appendix.

\noindent
\textbf{Results.} We observed three key phenomena under {\textit{\#Protocol-1}}: $1)$ From the data complexity perspective, all methods tend to drop the performance lengthways along the Table~\ref{tab:case-specific-novel-view}, with the level of accessory complexity increasing. This phenomenon reflects the status quo that we do not yet have one strong paradigm for robust case-specific human head multi-view reconstruction. $2)$ NeuS yields the best performance on average in terms of three metrics. There are two possible underlying reasons. First, NV and MVP are dynamic methods while not emphasizing temporal-consistency constraints. Thus, when comparing these methods with static ones under static measurement, the perturbation of data sequences would affect these two methods' construction on dynamic fields to certain degrees. Second, by associating the quantitative results with the qualitative ones, we can find that NeuS performs well in global shape reconstruction with almost no surrounding noise due to its surface representation property, but has a much smoothing surface appearance. In contrast, Instant-NGP and MVP can recover better high-frequency details. MVP uses multiple-primitive representation with different networks to render, equipping the model with a larger representative capacity. Whereas, they produce more surrounding noise. Neural Volume renders images mostly with artifacts. We could draw the idea that surface representation helps the novel view reconstruction in a global shape-forming manner. $3)$ NV suffers from limited grid resolution (although it uses inverse warping to ease the problem) and inaccurate alpha value estimation. Thus, it introduces more artifacts than other methods, and is strenuous in reconstructing  high-frequency details.  

{

\begin{table}[htb]
\begin{center}
\resizebox{0.475\textwidth}{!}{
\begin{tabular}{c|c|cccc}
\toprule[1.5pt]
\textbf{Split} & \textbf{Metrics} & \textit{NGP}~\cite{muller2022instant} &\textit{NeuS}~\cite{wang2021neus} & \textit{NV}~\cite{lombardi2019neural} & \textit{MVP}~\cite{lombardi2021mixture} \\

\midrule
\multirow{3}{*}{\begin{tabular}{l} \textbf{Normal Case} \end{tabular}} &
   PSNR$\uparrow$      & 24.71 & \best{26.29} & \worst{19.61} & 23.65 \\
~  & SSIM$\uparrow$    & 0.848 & \best{0.927} & \worst{0.777} & 0.895 \\
~  & LPIPS$\downarrow$ & 0.28  & \best{0.11}  & \worst{0.29} & 0.14 \\
\midrule
\multirow{3}{*}{\begin{tabular}{l} \textbf{With Deformable} \\\relax \makecell{\\} \textbf{Accessory} \end{tabular}} & 
   PSNR$\uparrow$      & 23.06 & 23.53 & \worst{17.83} & \best{23.93} \\
~  & SSIM$\uparrow$    & 0.807 & \best{0.904} & \worst{0.703} & 0.893 \\
~  & LPIPS$\downarrow$ & 0.31  & 0.13  & \worst{0.34} & \best{0.12} \\
\midrule
\multirow{3}{*}{\begin{tabular}{l} \textbf{With Complex} \\\relax \makecell{\\} \textbf{Accessory} \end{tabular}} &
   PSNR$\uparrow$      & 20.54 & \best{22.89} & \worst{16.46} & 21.5 \\
~  & SSIM$\uparrow$    & 0.776 & \best{0.874} & \worst{0.598} & 0.83 \\
~  & LPIPS$\downarrow$ & 0.36  & \best{0.16}  & \worst{0.44} & 0.18 \\
\midrule
\multirow{3}{*}{\begin{tabular}{l} \textbf{Overall} \end{tabular}} &
   PSNR$\uparrow$      & 23.21 & \best{24.67} & \worst{18.56} & 23.1 \\
~  & SSIM$\uparrow$    & 0.819 & \best{0.906} & \worst{0.723} & 0.876 \\
~  & LPIPS$\downarrow$ & 0.31  & \best{0.13}  & \worst{0.33} & 0.15 \\
\bottomrule[1.5pt]
\end{tabular}
}
\caption{\textbf{Single ID Novel View Synthesis (\textit{\#Protocol-1}).} We evaluate four methods on this task under three subsets with levels of complexity.}
\vspace{-0.5cm}
\label{tab:case-specific-novel-view}
\end{center}
\end{table}

\begin{table*}[htb]
\begin{center}
\resizebox{0.95\linewidth}{!}{
\begin{tabular}{c|c|c|ccc|ccc|ccc|ccc} \hline
\multirow{2}*{\textbf{Training Setting}} & \multirow{2}*{\textbf{Testing Setting}}  
& \multirow{2}*{\textbf{Methods}}
& \multicolumn{3}{c|}{\textbf{Normal Case}}
& \multicolumn{3}{c|}{\textbf{With Deformable Accessories}}
& \multicolumn{3}{c|}{\textbf{With Complex Accessories}}
& \multicolumn{3}{c}{\textbf{Overall}}\\
& & & \multicolumn{1}{c}{PSNR$\uparrow$} & \multicolumn{1}{c}{SSIM$\uparrow$} & \multicolumn{1}{c|}{LPIPS$\downarrow$*}
& \multicolumn{1}{c}{PSNR$\uparrow$} & \multicolumn{1}{c}{SSIM$\uparrow$} & \multicolumn{1}{c|}{LPIPS$\downarrow$*}
& \multicolumn{1}{c}{PSNR$\uparrow$} & \multicolumn{1}{c}{SSIM$\uparrow$} & \multicolumn{1}{c|}{LPIPS$\downarrow$*}
& \multicolumn{1}{c}{PSNR$\uparrow$} & \multicolumn{1}{c}{SSIM$\uparrow$} & \multicolumn{1}{c}{LPIPS$\downarrow$*}\\ \hline
\multirow{6}*{Fixed Views} & \multirow{3}*{Fixed Views} &
IBRNet~\cite{wang2021ibrnet}& 23.36 & 0.918 & 144.17 & 20.82 & 0.849 & 197.85 & 20.33 & 0.827 & 187.57 & 21.97 & 0.878 & 168.44 \\
& & VisionNeRF~\cite{lin2023visionnerf} &23.57 & 0.905 & 139.52 & 20.42 & 0.846 & 186.04 & 20.89 & 0.835 & 189.60 & 22.11 & 0.873 & 163.67 \\
& & KeypointNeRF~\cite{mihajlovic2022keypointnerf} & 19.59 & 0.898 & 127.06 & 17.42 & 0.805 & 213.43 & 16.54 & 0.760 & 205.82 & 18.29 & 0.840 & 168.34 \\
\cline{2-15}
& \multirow{3}*{Random Views} &
IBRNet~\cite{wang2021ibrnet}& 24.34 & 0.924 & 140.21 & 20.81 & 0.85 & 189.57 & 20.45 & 0.832 & 179.57 & 22.485 & 0.883 & 162.39 \\
& & VisionNeRF~\cite{lin2023visionnerf} & 25.79 & 0.914 & 148.70 & 21.43 & \best0.883 & \best148.90 & 20.37 & 0.87 & 159.50 & 23.345 & 0.895 & 151.45\\
& & KeypointNeRF~\cite{mihajlovic2022keypointnerf} & \worst16.96 & \worst0.871 & \worst170.66 & \worst16.07 & \worst0.775 & \worst256.83 & \worst14.64 & \worst0.714 & \worst270.33 & \worst16.16 & \worst0.808 & \worst217.12 \\
\hline
\multirow{6}*{Random Views} & \multirow{3}*{Fixed Views} &
IBRNet~\cite{wang2021ibrnet}& 23.37 & 0.918 & 144.21 & 20.82 & 0.8487 & 197.85 & 19.79 & 0.803 & 182.85 & 21.84 & 0.872 & 167.28 \\
& & VisionNeRF~\cite{lin2023visionnerf} &23.05 & 0.905 & 135.20 & 21.42 & 0.864 & 167.00 & 20.28 & 0.835 & 165.01 & 21.95 & 0.877 & 150.60\\
& & KeypointNeRF~\cite{mihajlovic2022keypointnerf} & 19.74 & 0.902 & 113.5 & 18.05 & 0.817 & 183.66 & 17.02 & 0.778 & 182.08 & 18.64 & 0.850 & 148.19 \\
\cline{2-15}
& \multirow{3}*{Random Views} & 
IBRNet~\cite{wang2021ibrnet}& 24.38 & 0.924 & 139.71 & 21.02 & 0.850 & 190 & 20.91 & 0.837 & 175.14 & 22.67 & 0.884 & 161.14 \\
& & VisionNeRF~\cite{lin2023visionnerf} & \best28.08 & \best0.943 & \best97.32 & \best23.86 & 0.882 & 150.6 & \best23.08 & \best0.873 & \best133.2 & \best25.78 & \best0.910 & \best119.61\\
& & KeypointNeRF~\cite{mihajlovic2022keypointnerf} & 18.65 & 0.897 & 124.33 & 17.60 & 0.813 & 192.04 & 16.61 & 0.779 & 186.67 & 17.88 & 0.847 & 156.84 \\
\hline
\end{tabular}
}
\vspace{-0.5cm}
\end{center}
\caption{\textbf{Benchmark Results on Unseen Expression NVS (\textit{\#Protocol-1}).} We train generalizable methods given different source view settings. The results are evaluated on unseen expressions of sampled training identities. (LPIPS* denotes LPIPS $\times$ 1000)}
\label{tab:seen_id_unseen_exp}
\end{table*}
}

\begin{table*}[htb]
\begin{center}
\resizebox{0.95\linewidth}{!}{
\begin{tabular}{c|c|c|ccc|ccc|ccc|ccc} \hline
\multirow{2}*{\textbf{Training Setting}} & \multirow{2}*{\textbf{Testing Setting}}  
& \multirow{2}*{\textbf{Methods}}
& \multicolumn{3}{c|}{\textbf{Normal Case}}
& \multicolumn{3}{c|}{\textbf{With Deformable Accessories}}
& \multicolumn{3}{c|}{\textbf{With Complex Accessories}}
& \multicolumn{3}{c}{\textbf{Overall}}\\
& & & \multicolumn{1}{c}{PSNR$\uparrow$} & \multicolumn{1}{c}{SSIM$\uparrow$} & \multicolumn{1}{c|}{LPIPS$\downarrow$*}
& \multicolumn{1}{c}{PSNR$\uparrow$} & \multicolumn{1}{c}{SSIM$\uparrow$} & \multicolumn{1}{c|}{LPIPS$\downarrow$*}
& \multicolumn{1}{c}{PSNR$\uparrow$} & \multicolumn{1}{c}{SSIM$\uparrow$} & \multicolumn{1}{c|}{LPIPS$\downarrow$*}
& \multicolumn{1}{c}{PSNR$\uparrow$} & \multicolumn{1}{c}{SSIM$\uparrow$} & \multicolumn{1}{c}{LPIPS$\downarrow$*}\\ \hline
\multirow{6}*{Fixed Views} & \multirow{3}*{Fixed Views} &
IBRNet~\cite{wang2021ibrnet}& 22.25 & 0.895 & 157.96 & 18.42 & 0.824 & 213.55 & 17.97 & 0.744 & 255.98 & 20.22 & 0.840 & 196.36 \\
& & VisionNeRF~\cite{lin2023visionnerf} &21.01 & 0.866 & 146.44 & 18.00 & 0.801 & 216.22 & 17.35 & 0.734 & 262.60 & 19.34 & 0.817 & 192.93 \\
& & KeypointNeRF~\cite{mihajlovic2022keypointnerf} & 18.85 & 0.866 & 148.13 & 15.93 & 0.789 & 205.04 & 16.14 & 0.734 & 231.89 & 17.44 & 0.814 & 183.30 \\
\cline{2-15}
& \multirow{3}*{Random Views} &
IBRNet~\cite{wang2021ibrnet}& 22.54 & 0.897 & 154.06 & 18.72 & 0.831 & 198.75 & 18.12 & 0.751 & 249.35 & 20.48 & 0.844 & 189.06\\
& & VisionNeRF~\cite{lin2023visionnerf} & 24.01 & \worst0.818 & 150.04 & 18.15 & 0.857 & 198.46 & 19.33 & 0.796 & 197.42 & 21.38 & 0.822 & 173.99\\
& & KeypointNeRF~\cite{mihajlovic2022keypointnerf} & \worst17.03 & 0.841 & \worst187.19 & \worst14.79 & \worst0.76 & \worst244.94 & \worst15.46 & \worst0.715 & \worst273.00 & \worst16.08 & \worst0.789 & \worst223.08 \\
\hline
\multirow{6}*{Random Views} & \multirow{3}*{Fixed Views} &
IBRNet~\cite{wang2021ibrnet}& 22.24 & 0.895 & 157.95 & 18.42 & 0.824 & 213.55 & 18.01 & 0.746 & 256.81 & 20.23 & 0.840 & 196.57 \\
& & VisionNeRF~\cite{lin2023visionnerf} &21.92 & 0.889 & 139.90 & 18.43 & 0.833 & 176.12 & 18.35 & 0.773 & 223.04 & 20.16 & 0.846 & 169.74\\
& & KeypointNeRF~\cite{mihajlovic2022keypointnerf} & 18.96 & 0.868 & 138.21 & 16.15 & 0.800 & 185.43 & 16.12 & 0.744 & 230.09 & 17.55 & 0.820 & 172.99 \\
\cline{2-15}
& \multirow{3}*{Random Views} & 
IBRNet~\cite{wang2021ibrnet}& 22.53 & 0.897 & 154.05 & 18.75 & 0.830 & 195.12 & 18.10 & 0.749 & 250.72 & 20.48 & 0.843 & 188.49 \\
& & VisionNeRF~\cite{lin2023visionnerf} & \best24.77 & \best0.918 & \best110.4 & \best20.22 & \best0.858 & \best149.30 & \best19.35 & \best0.797 & \best196.90 & \best22.28 & \best0.873 & \best141.75\\
& & KeypointNeRF~\cite{mihajlovic2022keypointnerf} & 18.02 & 0.865 & 145.30 & 15.75 & 0.794 & 194.16 & 16.15 & 0.747 & 227.49 & 16.99 & 0.818 & 178.06 \\
\hline
\end{tabular}
}
\vspace{-0.5cm}
\end{center}
\caption{\textbf{Benchmark Results on Unseen ID NVS (\textit{\#Protocol-2}).} We train generalizable methods given different source views during training/testing time, and the reported numbers are evaluated on unseen identities in each test split. (LPIPS* denotes LPIPS $\times$ $1000$)}
\vspace{-0.3cm}
\label{tab:unseen_person}
\end{table*}

\begin{table}[htb]
\begin{center}
\resizebox{0.98\linewidth}{!}{
\begin{tabular}{c|c|l} \hline
\textbf{Training Setting} & {\textbf{Testing Setting}}  & {\textbf{Explanation}}\\ \hline
\multirow{4}*{Fixed Source Views} & \multirow{2}*{Fixed Source Views} & The model is trained given fixed source camera \\
& & views and tested with the same source view indexes. \\ \cline{2-3}
 & \multirow{2}*{Random Source Views} & The model is trained given fixed source camera \\
 & & views and tested with random source view indexes. \\ \hline
\multirow{5}*{Random Source Views} & \multirow{2}*{Fixed Source Views} & The model is trained given random source camera \\
& & views and tested with the fixed source view indexes. \\ \cline{2-3}
 & \multirow{3}*{Random Source Views} & The model is trained given random source camera \\
 & & views and tested with re-random selected source \\ & & view indexes. \\ \hline
\end{tabular}
}
\end{center}
\vspace{-0.5cm}
\caption{\textbf{Explanation for Training-Testing Settings in Generalizable NVS.} All settings are evaluated on the same camera split of target views, and source views are selected apart from the target views. Tested random views are constrained under a certain angle range. At inference, three source views are provided.}
\vspace{-0.5cm}
\label{setting details}
\end{table}

\subsubsection{Generalizable NVS}\label{ID nvs}
This track refers to the setting of training across multiple human heads, and testing on unseen{\footnote{Note that the adjective ~`seen' refers to the sample that is used in training, and ~`unseen' means the sample is not used as training data.} human heads (\textit{i.e.,} new identities) or unseen motions (\textit{e.g.,} expressions) with {\textit{conditioning on one or few input images}} (as source views). It allows us to evaluate the network's effectiveness in learning priors, and the ability to adapt priors. We investigate three methods under two protocols in the main paper -- $1)$ {\textit{\#Protocol-1}} for investigating methods' generalization ability on geometry deformation via evaluating the generalization ability to unseen expressions on seen identities. $2)$ {\textit{\#Protocol-2}} for probing methods' capability in learning category-level human head priors via evaluating the generalization ability to unseen identities. We also name this protocol as {\textit{Unseen ID NVS}}; This setting is challenging as it requires the model to generalize to both new appearances and geometries. To further reveal the factors that might have influences on generalization performance, we enrich both protocols with four sets of training-testing view settings and three data subsets under different complexity.

\noindent
\textbf{Settings.}  We study three generalizable methods: IBRNet~\cite{wang2021ibrnet}, VisionNeRF~\cite{lin2023visionnerf}, and KeypointNeRF~\cite{mihajlovic2022keypointnerf} in terms of PSNR/SSIM/LPIPS metrics. For the training-testing identity split, we select a subset from RenderMe-360,  with $160$ identities for training and $20$ for serving as unseen identities. The selected identities are evenly sampled from the three data subsets. We select $7$ out of $60$ camera views as novel views{\footnote{\small{In our  Generalizable NVS setting, these {\textit{camera indexes}} of novel views are under certain probability to be sampled in training split as the target views. In other words, the views might not be novel to seen identities but are novel to unseen identities. We design such a strategy to facilitate the models to learn the prior that encompass the feature mapping information under view transformation. There is an alternative setting in the research community that excludes novel views for both seen and unseen identities~\cite{lin2023visionnerf}. We also encourage readers to evaluate their methods under the alternative setting for more ample performance analysis.}}, the others are used as the source view training candidates. During training, we select three camera views from candidates as source views and use them as the image conditions to train the models. The criterion of source view selections differs among the four training-testing view settings. Table~\ref{setting details} presents the explanations.} Note that,
we calculate the metrics in {\textit{\#Protocol-2}} on all $12$ expressions. As a consequence, $10$ expression structures attached with trained identities are covered in the training set, and the rest $2$ expression structures are unseen. Such an evaluation strategy provides the feasibility for researchers to analyze their methods' generalization ability on appearance and geometry in both entangled and disentangled aspects.

\begin{figure}
    \centering
    \includegraphics[width=1\linewidth]{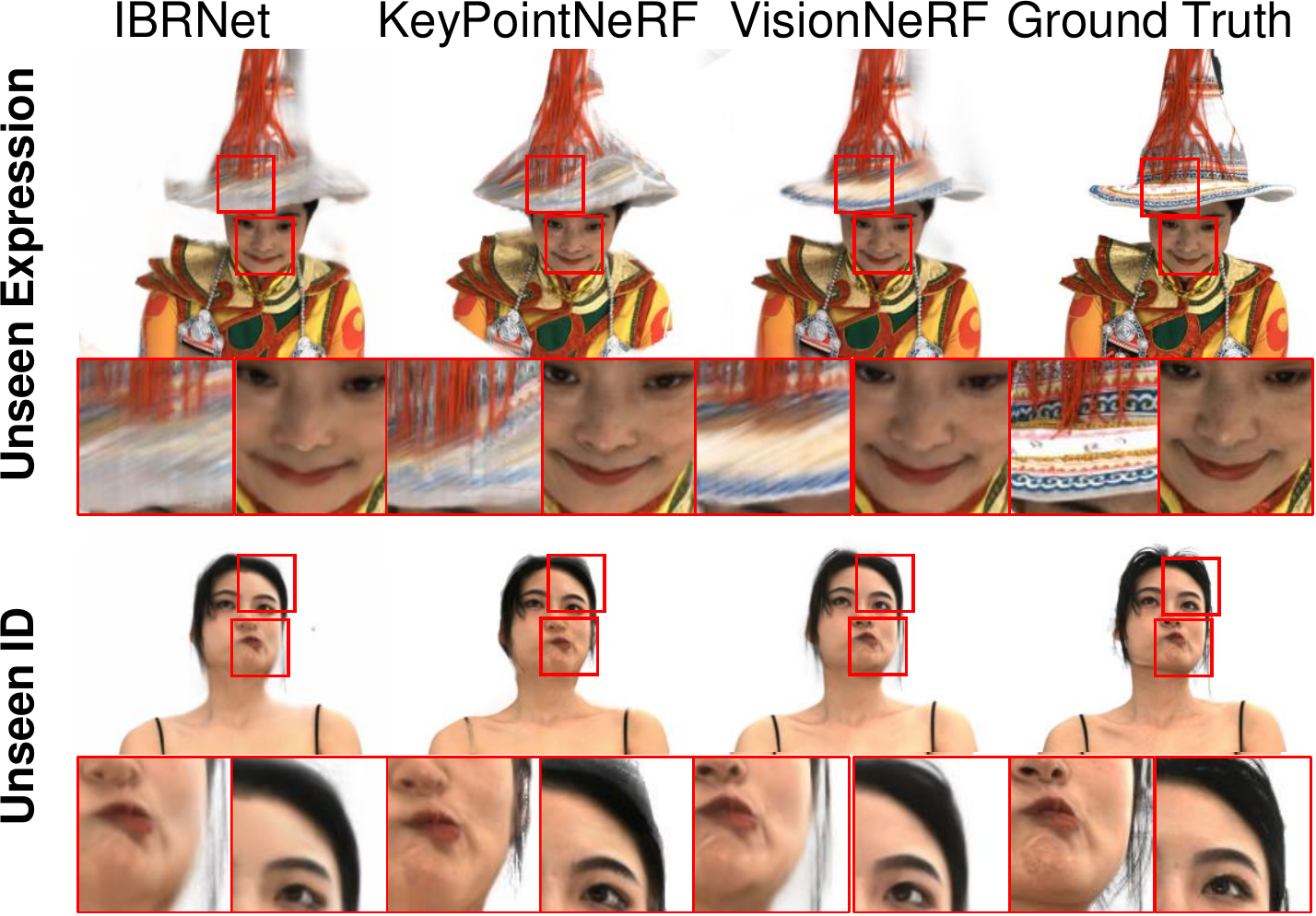}
    \caption{\textbf{Qualitative Results of Generalizable NVS (\textit{\#Protocol-1\&2}).} We illustrate four generalizable methods, \textit{i.e.,}IBRNet, KeypointNeRF, and VisionNeRF in two different settings. The regions in red boxes are zoomed in for better visulization. }
    \label{fig:main-generlization}
\vspace{-0.5cm}
\end{figure}

\noindent
\textbf{Results.}
The quantitative results are shown in Table~\ref{tab:seen_id_unseen_exp} and~\ref{tab:unseen_person}. From per method perspective, we draw the consistent conclusions that: $1)$ random view training could help enhance the model's robustness on both unseen expression and identity tasks; $2)$ the performance declines in terms of most metrics with the complexity of human head's appearance/geometry increase; $3)$ the Unseen ID NVS task introduces larger performance drop rate than Unseen Expression NVS. These two phenomena suggest that these generalizable methods could learn priors like the information of ~`minimal-accessory' mean head, and local geometry transformation on a certain level, while still struggle with more diverse scenarios that are long-tail distributed~\footnote{Please associate the Tables with Figure~\ref{fig:att_annot}and~\ref{fig:att_annot2}.}. In addition, there are several interesting observations when comparing the three methods: $1)$ VisionNeRF\cite{lin2023visionnerf} achieves the best results on average. The robustness might come from its large capacity of learnable variables from a transformer-based structure on image features and the multi-resolution based encoder. $2)$ IBRNet\cite{wang2021ibrnet} results in blurry synthesis even under the train and test settings on fixed views. $3)$ KeypointNeRF~\cite{mihajlovic2022keypointnerf} falls behind for most of the scenarios, but is in the lead on LPIPS on average. In other words, KeypointNeRF benefits in perceptual measurement like LPIPS while suffering from pixel-wise measurements. We infer the possible reason behind the contradictory metric performances is that -- the modules driven by triangulated keypoints provide better feature and view alignments in an explicit manner to help reconstruct the radiance fields. Whereas, such a key insight is a double-edged sword for full human head tasks. Since only the facial region could be well guaranteed with accessible facial landmarks. As a consequence, non-facial regions, like the hat in Figure~\ref{fig:main-generlization}, are more blurry than the facial region and distorted in the geometry aspect.  Moreover, KeypointNeRF only renders the intersected frustum regions from source views in practice, which aggravates the performance problem from the full-head measurements. The results turn better when we only calculate regions that KeypointNeRF could render, as shown in Tab.~\ref{tab:mask_region} in the Appendix. 

\subsection{Novel Expression Synthesis}
This task refers to the setting of reconstructing a 4D facial avatar based on {\textit{monocular}} video sequences{\footnote{Note that, differing from unseen expression nvs protocol, the novel expression should be synthesized under the guidance of {\textit{non-target person's image}} prompts, such as facial expression parameters. The main focus of this setting is to evaluate methods' effectiveness in {\textit{dynamic changes}} of the surface of a face.}}. We study three representative methods with different expression settings -- $1)$ {\textit{\#Protocol-1}} for investigating the interpolation/extrapolation abilities of training on intentional expression structures and testing on novel ones. We discuss this setting in main paper; $2)$ {\textit{\#Protocol-2}} for exploring the robustness of training on normal conversation sequences, then testing on both new conversations and intentional expression structures. The normal conversation scenarios include subtle expression changes. They can help to verify a method's reconstruction on local motion transformation. The intentional expression structures provide the challenges of reconstructing 4D information in high-frequence texture/geometry, and multi-scale motion changes (Figure~\ref{fig:exp_capture}). We unfold this protocol in Section~\ref{nes-supp} in the Appendix. 

\begin{figure}
    \centering
    \includegraphics[width=0.975\linewidth]{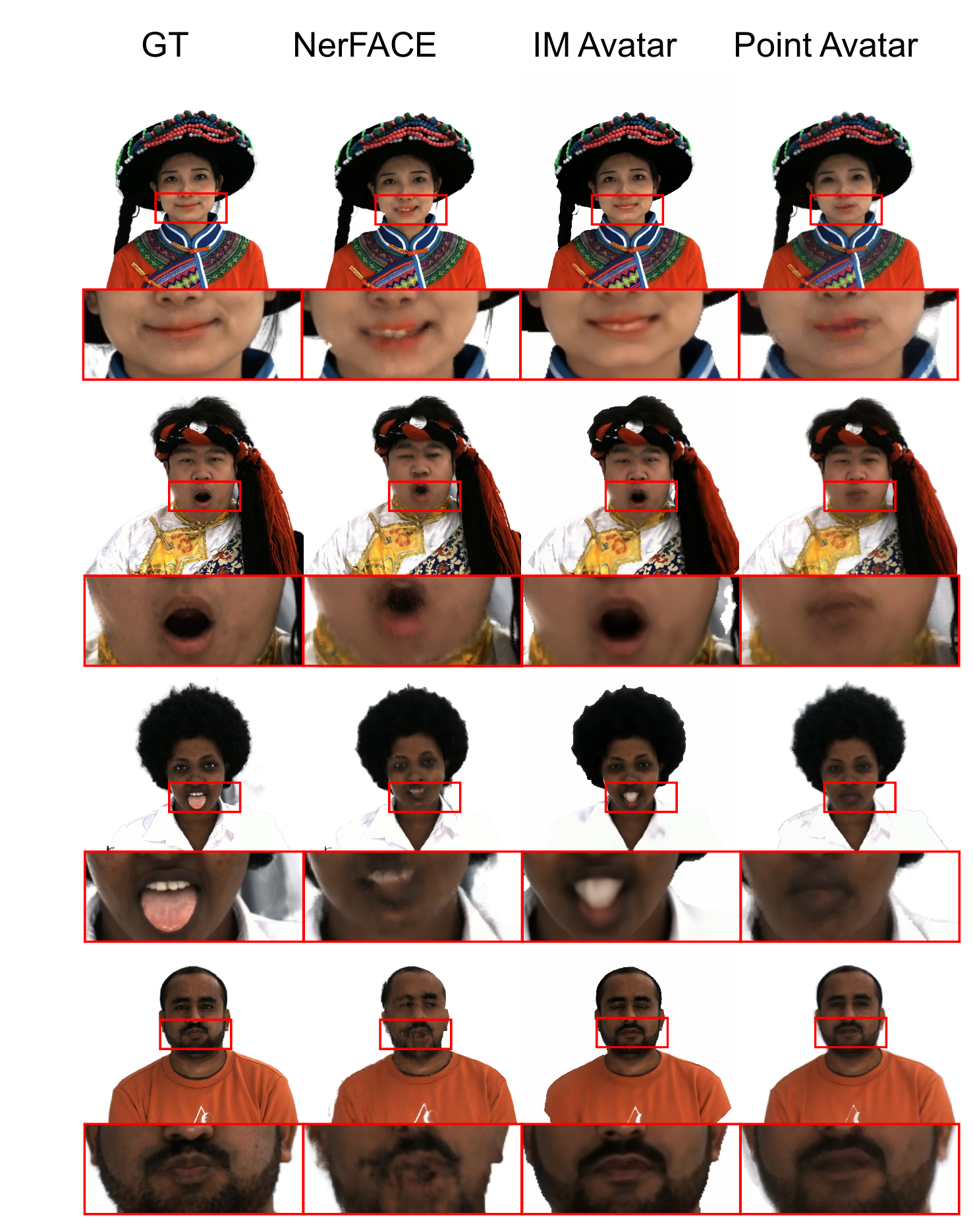}
    \caption{\textbf{Illustration of Novel Expression Synthesis (\textit{\#Protocol-1}).} We showcase four samples from both normal expression and hard expression splits.}
    \vspace{-0.5cm}
    \label{fig:novel_exp_exp-main}
\end{figure}

\noindent
\textbf{Settings.} We study three case-specific, deformable head avatar methods: NeRFace~\cite{Gafni_2021_CVPR}, IM Avatar~\cite{zheng2022avatar}, and Point Avatar~\cite{zheng2022pointavatar}. These methods showcase different paradigms of leveraging neural implicit representations for dynamic head avatars. The official implementation of IM Avatar suffers from unstable training when not using specific GPU {\footnote{This problem is frequently raised in GitHub Issues,~\textit{e.g.,} \url{https://github.com/zhengyuf/IMavatar/issues/3}, of the official release version.}}  We find one of the sensitive factors might relate to the FLAME parameters. We follow the official released data preprocessing pipeline of IM Avatar, where the FLAME parameters are initialized from DECA~\cite{DECA:Siggraph2021} and refined with single-view facial keypoints{\footnote{We abbreviate the preprocessing pipeline as DECA in the follow up sections with less rigorous.}}. In order to obtain relatively stable results (shown in Table ~\ref{tab:Case_Specific_Novel_Pose_Synthesis-main}), we also compare the results from DECA and our optimized FLAME parameters, which are shown in Table~\ref{tab:Case_Specific_Novel_Pose_Synthesis} and Figure~\ref{fig:nes} in the Appendix. All methods are evaluated in terms of  PSNR, SSIM, LPIPS, and L1 Distance, similar to ~\cite{zheng2022pointavatar}. For {\textit{\#Protocol-1}}, we select $20$ identities from the three categories (\textit{i.e.,} Normal, With Deformable/Complex Accessories) to form the benchmark data. We use $6$ expression sequences for per-identity training and the other $6$ expressions for testing.

\noindent
\textbf{Results.}
The quantitative result is presented in Table~\ref{tab:Case_Specific_Novel_Pose_Synthesis-main}. We split the novel expressions into normal and hard subsets according to their similarity to the training expression structures. We find PointAvatar outperforms the two implicit-based methods (IM Avatar and NerFace) on both splits under most of the metric measurements. The comparison suggests that combing explicit point-based representation with implicit one helps increase the robustness of new expression synthesis. This is reasonable since point cloud provides more flexibility and specificity in geometry deformation than pure implicit ones. But such a merit does not always exist. The granularity of points limits PointAvatar's performance on subtle motions (\textit{e.g.,}~`pout' in the last row of Figure~\ref{fig:novel_exp_exp-main}). In addition, we observe that all methods suffer from out-of-distribution cases like the ~`tongue out' in the third  row of the Figure. Moreover, from the whole-head rendering aspect, we find that IM Avatar struggles with thin structures like twisted hair band and hair strands. This is because IM Avatar constrains reconstruction on the surface. NerFace has fine rendering results in a global manner, while facing problems in robustly modeling dynamic motion. 

\begin{table}[!ht]
\begin{center}
\vspace{-0.5em}
\resizebox{0.475\textwidth}{!}{
\begin{tabular}{c|c|cccc}
\toprule[1.5pt]
Method & Split & L$_{1}$ $\downarrow$ &  PSNR $\uparrow$ & SSIM $\uparrow$ & LPIPS $\downarrow$ \\
\midrule
IM Avatar~\cite{zheng2022avatar} & \textbf{N} & \worst0.047 & 22.61 & \best0.903 & 0.134 \\
~                 & \textbf{H} & \worst0.047 & 21.91 & 0.895 & \worst0.149 \\
\midrule
NerFace~\cite{Gafni_2021_CVPR} & \textbf{N} & 0.034 &20.46 &0.876 &0.114\\
~               & \textbf{H} &0.037 & \worst18.89 & \worst0.865 &0.121\\
\midrule
PointAvatar~\cite{zheng2022pointavatar} & \textbf{N} & 0.0057 & 24.57 & 0.878 & 0.089 \\
~                   & \textbf{H} & \best0.0055 & \best25.05 & 0.883 & \best0.086\\
\bottomrule[1.5pt]
\end{tabular}
}
\caption{\small{\textbf{Novel Expression Synthesis (\textit{\#Protocol-1}).}  We benchmark three methods on different splits of RenderMe-360. \textbf{N}: Normal Expression, \textbf{H}: Hard Expression.}}
\label{tab:Case_Specific_Novel_Pose_Synthesis-main}
\vspace{-2.0em}
\end{center}
\end{table}

\subsection{Hair Rendering}

\begin{table*}[htb]
\begin{center}
\resizebox{0.975\textwidth}{!}{
\begin{tabular}{c|c|ccc|ccc|ccc|ccc}
\toprule[1.5pt]
\multirow{2}{*}{\begin{tabular}{l} \textbf{Aspects} \end{tabular}} &
\multirow{2}{*}{\begin{tabular}{l} \textbf{Benchmarks} \end{tabular}} &
        \multicolumn{3}{c|}{\textbf{Short Hair}} & \multicolumn{3}{c|}{\textbf{Long Hair}} & \multicolumn{3}{c|}{\textbf{Curls}} & \multicolumn{3}{c}{\textbf{Over All}} \\
~  & ~  & PSNR$\uparrow$ & SSIM$\uparrow$ & LPIPS$\downarrow$ & PSNR$\uparrow$ & SSIM$\uparrow$ & LPIPS$\downarrow$ & PSNR$\uparrow$ & SSIM$\uparrow$ & LPIPS$\downarrow$ & PSNR$\uparrow$ & SSIM$\uparrow$ & LPIPS$\downarrow$ \\

\midrule[1.5pt]
\multirow{2}{*}{\begin{tabular}{l} \textbf{Static Rendering} \end{tabular}} &
    \textbf{Instant-NGP~\cite{muller2022instant}} & \best{25.53} & 0.848 & 0.274 & \best{24.99} & 0.834 & 0.29 & 21.06 & 0.789 & \worst0.355 & \best{23.75} & 0.822 & \worst0.309 \\

~  & \textbf{NeuS~\cite{wang2021neus}} & 23.54 & \best{0.851} & \best{0.108} & 21.05 & \worst0.746 & 0.239 & \best{21.3} & 0.789 & \best{0.261} & 21.76 & \worst0.787 & 0.214  \\

\midrule[1.5pt]
\multirow{2}{*}{\begin{tabular}{l} \textbf{Dynamic Rendering} \end{tabular}} &
    \textbf{MVP~\cite{mvp}} & 24.31 & 0.821 & 0.148 & 22.56 & \best{0.868} & \best{0.197} & 20.97 & \best{0.795} & 0.262 & 22.61 & \best{0.856} & \best{0.201} \\

~  & \textbf{NV~\cite{lombardi2019neural}} & \worst21.19 & \worst0.816 & \worst0.289 & \worst20.48 & 0.829 & \worst0.263 & \worst19.275 & \worst0.764 & 0.351 & \worst20.32 & 0.806 & 0.297 \\

\midrule[1.5pt]
\multirow{2}{*}{\begin{tabular}{l} \textbf{Time-Interpolation} \end{tabular}} &
    \textbf{NSFF~\cite{2021nsff}} & \best{27.98} & \best{0.856} & \best{0.094} & \best{28.27} & \best{0.867} & \best{0.094} & \best{28.231} & \best{0.846} & \best{0.112} & \best{28.19} & \best{0.858} & \best{0.098} \\

~  & \textbf{NR-NeRF~\cite{2021nrnerf}} & 27.14 & 0.851 & 0.114 & 27.62 & 0.865 & 0.122 & 27.825 & 0.84 & 0.136 & 27.563 & 0.854 & 0.124 \\

\bottomrule[1.5pt]
\end{tabular}
}

\caption{\textbf{Quantitative Results of Hair Rendering.} We study six methods for the hair rendering task under three settings. In static rendering and dynamic rendering, we evaluate the novel view synthesis result, while we render the image of the same camera view but evaluate an inter-novel time stamp in the time-interpolation part.}
\vspace{-0.5cm}
\label{tab:hair-rendering-result}
\end{center}
\end{table*}

This task refers to the setting of modeling accurate hair appearance across changes of viewpoints or dynamic motions. We focus on three sub-problems of hair rendering: $1)${\textit{\#Protocol-1}} for probing current methods' effectiveness on static hair reconstruction, in which methods are trained on multi-view images and tested on novel views; $2)$ {\textit{\#Protocol-2}} for evaluating the algorithms' capability on dynamic hair performance capture, in which methods are trained on multi-view video sequences and tested on the motion sequences under novel views; $3)$ {\textit{\#Protocol-3}} for investigating the methods' interpolation ability on dynamic hair motion, in which the methods are trained on frames sampled from a monocular video, and tested on the rest frames of the video.

\noindent
\textbf{Settings.} We select a subset from RenderMe-360 to form the benchmark for this task, with $20$ representative wig collections from $8$ randomly picked human subjects. This subset is further split into three groups,~\textit{i.e.,} short hair, long hair, and curls, according to the complexity of hair strand intersections. In total, we study six representative methods under the three mentioned protocol settings. The evaluation metrics are PSNR, SSIM, and LPIPS. Concretely, we discuss Instant-NGP \cite{muller2022instant} as well as NeuS~\cite{wang2021neus} for {\textit{\#Protocol-1}}. We train the models with $38$ camera views of a specific frame (the one with the largest motion magnitude in the video) and evaluate their performances with the rest $22$ views. The distribution of camera split is the same as the one in Section~\ref{section:Novel View Synthesis Benchmar}. For {\textit{\#Protocol-2}}, we study two dynamic neural rendering methods -- MVP~\cite{lombardi2021mixture} and NV~\cite{lombardi2019neural}. The methods are evaluated under $4$ held-out views of motion sequences. The four views are distributed around the front, double side, and back of the human head. For training, the other $56$ views of the motions are fed into the models. For {\textit{\#Protocol-3}}, we reveal the effectiveness of NSFF~\cite{2021nsff} and NR-NeRF~\cite{2021nrnerf}.  We take a camera from a frontal view as the monocular camera, and sample the input sequence in $10$ FPS. The rest frames are used as evaluation data. This strategy results in about $30$ frames for training per motion sequence and $60$ frames for testing. The training data volume is similar to the original papers, while the testing data volume is larger for a more comprehensive evaluation. Note that, hair rendering is a long-standing task, and there are many instructive methods. For example, state-of-the-art multi-view hair rendering methods like HVH~\cite{wang2022hvh}, and Neural Strand~\cite{rosu2022neuralstrands} are also valuable. However, most of the methods are not open-sourced, and difficult to be re-implemented with aligned performances claimed in the original papers. Also, there are various quantitative evaluation settings among the hair rendering research efforts, and these settings emphasize many different aspects. We discuss six neural rendering methods that are not customized for hair but representative in rendering, to explore their adaption ability and provide open-source baselines for this task.  We leave the exploration of more interesting and challenging scenarios upon RenderMe-360 dataset to the community for future work.

\noindent
\textbf{Result.} The quantitative results are shown in Table~\ref{tab:hair-rendering-result}. We observe several interesting phenomena. $1)$ For methods under the NVS tracks of static hair rendering and dynamic hair rendering, their performances all show a declining trend with the increasing complexity of hair geometry. Specifically, the ~`curls' scenario leads the methods to sharp performance drops under all metrics. This is reasonable, as curls data provides more challenges than the other two categories in terms of the difficulties in modeling more diverse intersections, complex motion situations, and high-frequency details.  $2)$ NSFF and NR-NeRF remain roughly flat performances under the time-interpolation synthesis protocol. NSFF models the dynamic scene as a continuous function with the utility of a time-dependent neural scene flow field, and optimizes the function with spatial and temporal constraints. Its design help to achieve robustness in different motion interpolation scenarios.  NR-NeRF has merits in dynamic reconstruction for disentangling dynamic motion into rigid and non-rigid parts. It 
introduces the ray-bending network to model the non-rigid motion, and a rigidity network to constrain the rigid regions. $3)$ From the hair motion aspect, long hair/curls scenarios contribute mostly to non-rigid deformation, whereas NSFF is superior to NR-NeRF in terms of three metrics. We infer that the deformation model of NR-NeRF has a flaw in capturing exact correspondences between images at different time steps, 
which leads to blur accumulated results along multiple frames. Figure~\ref{fig:hair_render}(b) in the Appendix also demonstrates the surmise from a qualitative perspective. Specifically, NSFF renders more fine detail hair strands than NR-NeRF. Moreover, as a side observation, the reconstructed face (where most part of the region is under the rigid-transformation across time) of the second subject is blurry,  which may show a relatively unstable time-interpolation ability of the learned latent code in NR-NeRF. $4)$ In the static rendering, Instant-NGP has overall better `PSNR' and `SSIM' than NeuS. From Figure~\ref{fig:hair_render}(a), we can also observe that Instant-NGP renders hair in better high-frequency patterns. We infer that the individual local-part reconstruction strategy in Instant-NGP helps in fine-detail pattern reconstruction. $5)$ MVP performs better in all three metrics compared to NV. Whereas, these two methods show more blur reconstruction than static methods (Figure~\ref{fig:hair_render}(a)). The phenomenon suggests the efforts of dynamic
 field designs should also be paid to the preservation of per-frame precision, rather than only focusing on deformation to new frames.

\subsection{Hair Editing}

Editing hair attributes, \textit{e.g.,} color, hairstyle, and hair position, is an interesting but challenging task. The operations could be done in 2D ~\cite{tan2020michigan,xiao2021sketchhairsalon, saha2021LOHO} or 3D ~\cite{wang2022hvh, rosu2022neuralstrands} manner with various conditions. Here, we showcase one sub-direction -- text-aware 2D hair editing, to give an example of the possible usages of our text annotation. This task refers to the setting of editing the hair attributes, given the source image and target text prompt. 

\noindent
\textbf{Settings.} For the evaluated data, we select $45$ representative head images from the neutral expression subset of RenderMe-360. These images consist of $30$ normal hairstyles, and $15$ identities with deformable head accessories. The data samples vary from each other with distinctive attributes, such as hair color, hairdo, skin tone and makeup. Upon the data, we present four configurations of possible ways to utilize our text annotation under the hair editing task. Concretely, we assemble two state-of-the-art text-based hair editing methods (\textit{i.e.,} HairCLIP~\cite{wei2022hairclip} and StyleCLIP~\cite{Patashnik_2021_ICCV} ) with popular inversion strategies~\cite{tov2021designing, roich2021pivotal, alaluf2021restyle, alaluf2021hyperstyle} to form the configurations. For the first configuration, we apply HairCLIP~\cite{wei2022hairclip}, which designs specific mappers for hair color and hairstyle editing,  based on text or image references. We follow the official implementation to test the capability of text-based editing after face alignment and e4e~\cite{tov2021designing} inversion. For the second configuration, we still focus on HairCLIP, but replace e4e~\cite{tov2021designing} with another inversion method, \textit{i.e.,} Restyle\_e4e~\cite{alaluf2021restyle}. Since Restyle\_e4e strategy theoretically has better identity preserving ability. For the other two configurations, we combine another famous text-based pre-trained model StyleCLIP~\cite{Patashnik_2021_ICCV}, with utilizing the other two inversion methods (PTI~\cite{roich2021pivotal} and HyperStyle~\cite{alaluf2021hyperstyle}). We choose StyleCLIP's global direction style editing for adapting arbitrary text references. Note that, HairCLIP trains the mapper network to predict the latent code change conditioned on the text prompt, and the latent code in the $\mathcal{W}$+ space comes from e4e inversion. The general idea behind e2e is picking more editable latent codes in $\mathcal{W}$+ space, which is not conceptually aligned with PTI-trend (which augments the manifold to include the image). Thus, we only showcase the combination of e4e/Restyle\_e4e inversion with the pre-trained HairCLIP model. For the evaluation metrics,  we follow the metrics used in HyperStyle~\cite{alaluf2021hyperstyle}: identity similarity score (ID-score\cite{deng2019arcface}), MS-SSIM, LPIPS, and pixel-wise L2 distance to evaluate the inversion results with the source images. 

\begin{table}[!ht]
\begin{center}
\resizebox{0.475\textwidth}{!}
{
\begin{tabular}{c|c|cccc}
\toprule[1.5pt]
Configuration & Split & ID-score $\uparrow$ & MS-SSIM $\uparrow$ &  LPIPS $\downarrow$ & L2 $\downarrow$ \\
\midrule[1.5pt]
\multirow{2}{*}{\begin{tabular}{l} e4e~\cite{tov2021designing} \end{tabular}}
~ & \textbf{N} & 0.57 & 0.81 & 0.16 & 0.032\\
~ & \textbf{H} & \worst0.52 & \worst0.74 & \worst0.23 & \worst0.069 \\
\midrule
\multirow{2}{*}{\begin{tabular}{l} Restyle\_e4e~\cite{alaluf2021restyle} \end{tabular}}
~ & \textbf{N} & 0.58 & 0.82 & 0.16 & 0.027 \\
~ & \textbf{H} & 0.54 & 0.76 & 0.22 & 0.059 \\
\midrule
\multirow{2}{*}{\begin{tabular}{l} PTI~\cite{roich2021pivotal} \end{tabular}}
~ & \textbf{N} & \best{0.88} & \best{0.95} & \best{0.06} & \best{0.003} \\
~ & \textbf{H} & 0.86 & 0.93 & 0.11 & 0.015 \\
\midrule
\multirow{2}{*}{\begin{tabular}{l} Hyperstyle~\cite{alaluf2021hyperstyle} \end{tabular}}
~ & \textbf{N} & 0.81 & 0.90 & 0.07 & 0.012 \\
~ & \textbf{H} & 0.77 & 0.87 & 0.10 & 0.033 \\

\bottomrule[1.5pt]
\end{tabular}
}
\end{center}
\caption{\small{\textbf{Different Inversions for Hair Editing.}  We showcase four inversion configurations for the hair editing task on the identities from our testset. \textbf{N} is short for normal cases. \textbf{H} is short for hard cases with deformable accessories.}}
\vspace{-0.5cm}
\label{tab:inversion}
\end{table}

\noindent
\textbf{Results.} Table ~\ref{tab:inversion} shows the quantitative results. Overall, all configurations function normally with our text annotation and data samples, which demonstrates the feasibility of utilizing our data in the hair editing domain. Among the four configurations, we could observe that PTI and HyperStyle show better quantitative results than the first two. 
The superiority is most significant in terms of identity preservation. From the aspect of methods' effectiveness on the out-of-distribution (OOD) samples, we can observe that PTI inversion is the most robust, while the performances of other methods decrease more from normal hairstyles to images with the deformable accessory. This is reasonable as high-quality datasets for training inversion methods are typically under the shortage of complex hair accessories, \textit{e.g.,} traditional high hats with ethnic characteristics. Additionally, the standard pre-processing requires cropped aligned faces, which often ignores partial hair and head accessories, as also been mentioned in ~\cite{yang2023styleganex}. This phenomenon reflects that there should be more research attention on the OOD problem, and the completeness regions that are associated with hair. In the Appendix, we present the qualitative results and analysis depicted in Figure~\ref{fig:hair_clip}.

\begin{table}[!ht]
\begin{center}
\resizebox{0.45\textwidth}{!}
{
\begin{tabular}{c|c|ccccc}
\toprule[1.5pt]
Method & Split & PSNR $\uparrow$ & SSIM $\uparrow$ &  LMD $\downarrow$ & Sync $\uparrow$ \\
\midrule[1.5pt]
\multirow{2}{*}{\begin{tabular}{l} AD-NeRF~\cite{guo2021ad} \end{tabular}}
~ & English & \best{18.44} & 0.83 & 2.29 & 2.75  \\
~ & Mandarin & 18.42 & \worst0.80 & \worst2.45 & \worst2.26  \\
\midrule
\multirow{2}{*}{\begin{tabular}{l} SSP-NeRF~\cite{liu2022semantic} \end{tabular}}
~ & English & \worst18.22 & \best{0.85} & \best{1.20} & 3.88  \\
~ & Mandarin & 18.31 & 0.81 & 0.95 & \best{4.20}  \\
\bottomrule[1.5pt]
\end{tabular}
}
\end{center}
\caption{\small{\textbf{Quantitative Evaluaction on the Talking Head Generation.} We benchmark AD-NeRF~\cite{guo2021ad} and SSP-NeRF~\cite{liu2022semantic} on two subsets of RenderMe-360.}}
\label{tab:talking_head}
\end{table}

\begin{figure}
    \centering
    \vspace{-0.5cm}
    \includegraphics[width=1.0\linewidth]{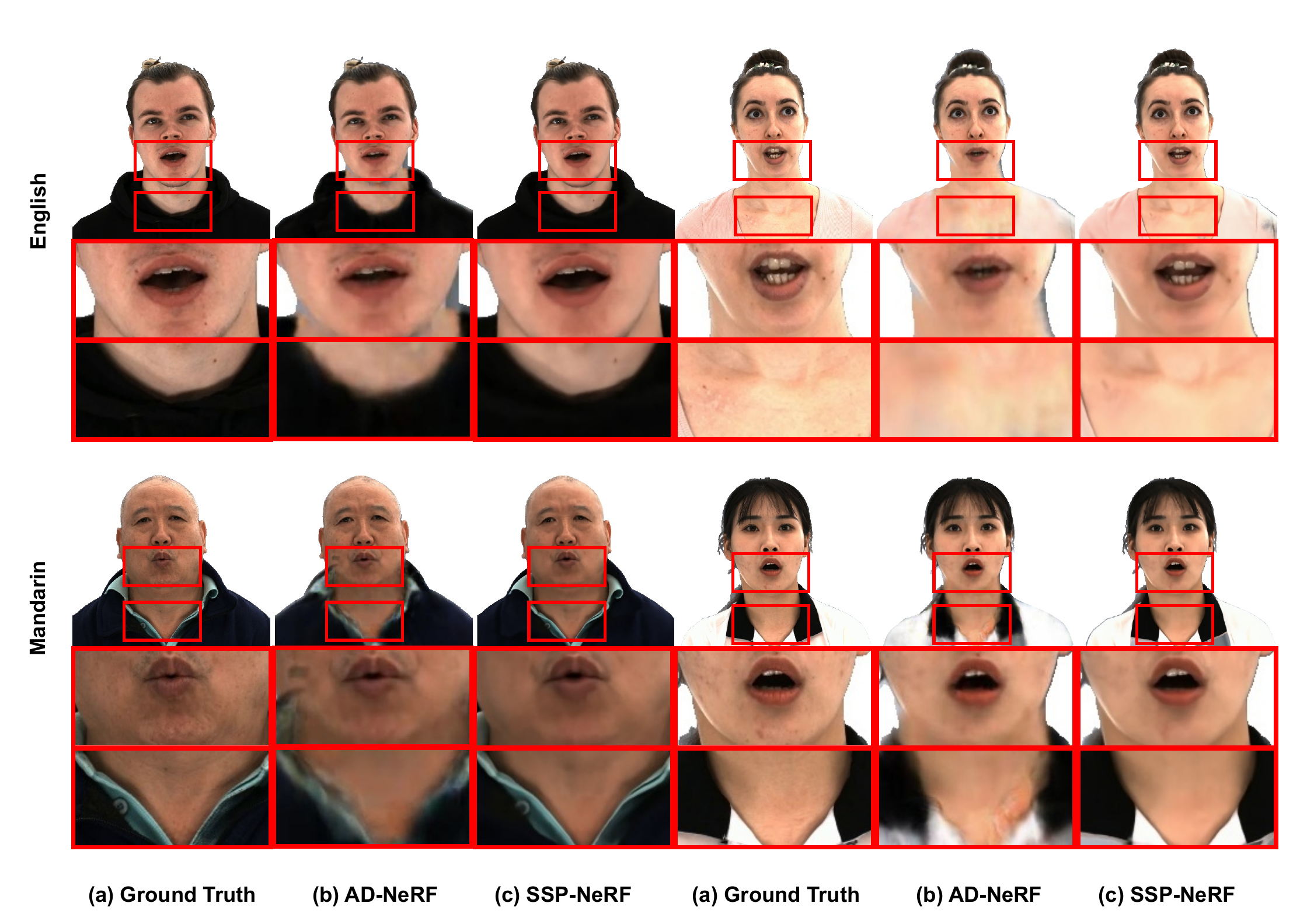}
    \caption{\textbf{Qualitative Illustration of Talking Head Generation.} We showcase results from AD-NeRF~\cite{guo2021ad} and SSP-NeRF~\cite{liu2022semantic} on four representative samples of RenderMe360. }

    \label{fig:talking_head}
\end{figure}

\subsection{Talking Head Generation}
With the phoneme-balanced corpus videos, our dataset can also serve as a standard benchmark for case-specific audio-driven talking head generation. This task refers to the setting of reenacting a specific person, with generating high-fidelity video portraits that are in sync with arbitrary speech audio as the driving source. We include two state-of-the-art talking-head methods to showcase the potential of our multi-sensory data. Previous approaches in this track mainly evaluate their performance on self-selected data. They manually extract several-minute video clips from TV programs or celebrity speeches for training and testing~\cite{guo2021ad,liu2022semantic,suwajanakorn2017synthesizing,thies2020neural}. Thus, there is a lack of unified selection criteria, and no benchmark agreement is achieved across different institutions yet. Additionally, some data sources (\textit{e.g.,} YouTube videos) may suffer from license issues. We hope our attempt could provide a standard benchmark for this task. 

\noindent
\textbf{Settings.} For evaluation data, we choose two subsets that cover two languages (\textit{i.e.,} English and Mandarin) from RenderMe-360. Each subset contains five distinctive identities, with six phoneme-balanced front-face videos per identity. 
Under this setting, we study two NeRF-based representative baselines, namely AD-NeRF~\cite{guo2021ad} and SSP-NeRF~\cite{liu2022semantic}. Compared with 2D generative model-based methods~\cite{jamaludin2019you,chen2019hierarchical, zhou2019talking} and explicit 3D mesh-aware ones ~\cite{thies2020neural}, these two methods bridge audio sources with implicit scene representation of neural radiance fields. Specifically, the two NeRF-based methods leverage pose and shape prior, along with audio information, to directly condition the semantic-aware NeRF. Such a methodology could theoretically help represent fine-scale head components (such as teeth and hair) with better photo-realistic synthesis quality. 
Following SSP-NeRF~\cite{liu2022semantic}, we utilize PSNR and SSIM metrics to evaluate image quality, while landmark distance (LMD) and SyncNet confidence (Sync)~\cite{Chung16a} are used to assess the accuracy of the lip movements.

\noindent
\textbf{Results.}
Table~\ref{tab:talking_head} and Figure~\ref{fig:talking_head} present the quantitative results and qualitative illustration of talking head models. From Table ~\ref{tab:talking_head}, AD-NeRF and SSP-NeRF exhibit similar PSNR and SSIM scores, but SSP-NeRF outperforms AD-NeRF in terms of LMD and Sync confidence. This phenomenon indicates that SSP-NeRF produces more accurate mouth shapes. The inference could be further supported by the qualitative results shown in Figure~\ref{fig:talking_head}, where SSP-NeRF's mouth shapes are closer to the ground truth. Additionally, the images generated by SSP-NeRF are clearer at the head and torso junctions. From the training language aspect, we can observe from Table~\ref{tab:talking_head} that,  there is no significant difference between the two splits in Mandarin and English. Both methods have similar support for these languages.  This reflects that even though the DeepSpeech model is used for extracting speech features that are primarily trained on non-Mandarin data, it still has good support for Mandarin due to its underlying word relationship capture ability. Moreover, the qualitative results are not ideal, if we compare models' performance to the test videos used in recent work~\cite{guo2021ad, liu2022semantic}. This demonstrates our dataset's potential as a new testset, uncovering more challenges for the case-specific audio-driven talking head generation. 

\section{Discussion}

\noindent
\textbf{Boarder impact and limitations.} The proposed RenderMe-360 dataset, together with the comprehensive benchmark, is expected to effectively facilitate modern head rendering and generation research. RenderMe-360 contains over $243$ million high-fidelity video frames and their corresponding meticulous annotations. However, as the field of human head avatar is consistently blooming, we could not include all of the related research topics, and all of the state-of-the-art methods at one time. Thus, we treat the construction of benchmarks based on RenderMe-360 as a long-standing mission of our team. We will construct more and more benchmarks on different topics unflaggingly, to support the sustainable and healthy development of the related research community. Also, we will build an open platform based on RenderMe-360. We sincerely encourage and welcome contributions to RenderMe-360 from the community, to boost the development of human head avatars together.

\noindent
\textbf{Conclusion.} We build a large-scale 4D human head dataset and relative benchmarks, RenderMe-360, for boosting the research on human head avatar creation. Our dataset covers $500$ subjects with diverse appearances, behaviors, and accents. We capture each subject with high-fidelity appearance, dynamic expressions, multiple hairstyles, and various speeches. Furthermore, we provide rich and accurate annotations, which encompass camera parameters, matting, 2D/3D facial landmarks, scans, FLAME fitting, and text descriptions. Upon the dataset, we conduct extensive experiments on the state-of-the-art methods to form a comprehensive benchmark study. The experimental results demonstrate that RenderMe-360 could facilitate downstream tasks, such as novel view synthesis, novel expression synthesis, hair editing, and talking head generation. We hope our dataset could unfold new challenges and provide the cues for future directions of related research fields. We will release all raw data, annotations, tools, and models to the research community.

{\small
\bibliographystyle{ieee_fullname}
\bibliography{egbib}
}

\clearpage
\appendix
\noindent
\textbf{\LARGE Appendix}
\vspace{5ex}

\setcounter{equation}{0}
\setcounter{figure}{0}
\setcounter{table}{0}
\setcounter{section}{0}
\makeatletter
\renewcommand{\theequation}{S\arabic{equation}}
\renewcommand{\thefigure}{S\arabic{figure}}
\renewcommand{\thetable}{S\arabic{table}}

In this supplementary material, we provide more information about the proposed RenderMe-360 dataset and additional experimental discussions for comprehensive benchmarking. Specifically, $(1)$ we introduce the dataset capturing process in detail (Section~\ref{sec:dataset}). The section includes three aspects: hardware construction, data collection, and data annotation of the proposed RenderMe-360 dataset. $(2)$ More comprehensive experiments are performed in multiple downstream tasks (Section~\ref{sec:benchmark}). We analyze the phenomena both qualitatively and quantitatively. $(3)$ We discuss some potential applications that can be benefited from our dataset, and list a toy example in the text-to-3D generation scenario, to show how to utilize our dataset in a flexible way (Section~\ref{sec:application}).

\section{Dataset Construction Details}
\label{sec:dataset}

In this section, we first introduce our physical capturing environment, namely POLICY (Section~\ref{sec:policy}). Second, we provide an elaborate data collection pipeline introduction(Section~\ref{sec:colloction}). Third, we present the detailed annotation processes regarding each annotated dimension (Section~\ref{sec:annote}). Finally, we analyze the data statistics of the proposed dataset in detail (Section~\ref{sec:statiscs}).

\subsection{Capture System: POLICY}
\label{sec:policy}
\noindent \textbf{Hardware Setup.}
We build a multi-video camera capture cylinder called POLICY to capture synchronized multi-view videos of the human head performance. The capture studio contains $60$ synchronous cameras with a resolution of $2448\times2048$. The sensor model is LBAS-U350-35C, and the shutter speed is at $30$ FPS for video capture. The cameras are arrayed in a cylindrical confined space, and they all point inward to the middle of the cylinder. We separate the camera array into four hierarchical layers. The first and the fourth layers use a large field of view to capture the overall head motion at a long distance, while the second and the third layers adopt a small field of view to capture more details of the head. 39 LED displays are used in the cylinder, where 6 are used to balance the lighting distribution in  front of the human face.

In addition, POLICY also contains five computers with high-performance CPUs and RAIDs, a network switch, eight frame grabbers, an extra camera, a time-code viewer, a condenser microphone, and fiber optic USB capture cables. The fiber optic USB capture cables are used to link the other devices. 

\noindent \textbf{Hardware Synchronization.}
It is a great challenge to achieve high-bandwidth capturing and synchronization in both visual portrait data collection from $60$ color cameras with different views, and audio-vision data collection from recording devices. We illustrate the structure design of POLICY in Figure~\ref{fig:link}, and show the reason why our POLICY can overcome the challenge in following paragraphs.

For visual data, POLICY connects every eight cameras to a frame grabber and a synchronization generator. Two frame grabbers are connected to a computer on the other end to achieve high-bandwidth transmission of the capturing data. A synchronization generator is connected in series to the next synchronization generator on the other end, and the first synchronization generator is linked to the first computer. During capturing visual data, the first computer controls all synchronization generators by launching a high-level trigger to achieve a microsecond error in the cameras' synchronization. 

For audio data, POLICY uses the extra camera to connect to a synchronization generator and the time-code viewer. A high-quality microphone is placed in front of the human head. The time-code viewer is linked to the microphone for the collection of the time stamp of the audio voice. The microphone and the extra camera are connected to a computer. During capturing audio data, the time code of the microphone and the synchronized signal from the extra camera enable the high-precise synchronization of audio-vision data.

All computers are connected to the network switch to synchronize the capturing operations and store the capturing data at high bandwidth. With the connection of these devices, POLICY achieves high-bandwidth capturing with the speed of 90 GB/s, multi-view synchronization, and audio-vision synchronization at the speed of $30$ Hz.

\begin{figure}[htb]
    \centering
    \includegraphics[width=1.0\linewidth]{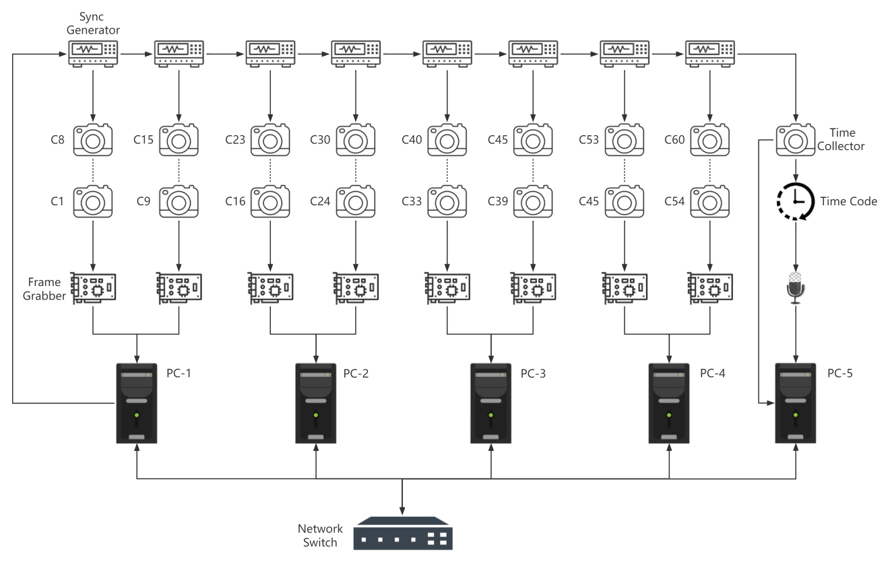}
    \caption{\textbf{The Structures of the POLICY.} 60 industrial high-definition cameras and a high-quality recording device are connected through synchronous generators, frame grabbers, five high-performance computers, and a network switch.}
    \label{fig:link}
\end{figure}

\subsection{Data Collection Details}
\label{sec:colloction}
\subsubsection{Criterion for Captured Attribute Design} We invite $500$ people to be our capture subjects. We require each subject to perform three different parts during the data capture, namely expression, hair, and speech. We will detail the collection process in Section~\ref{protocol}. In the current sub-section, we will describe the content design.

\noindent \textbf{Expression.} The design of expression collection is based on the standard proposed in i3DMM~\cite{9577855}, in which $10$ facial expressions are recorded as the train set and the other $5$ are used as the test set. We capture $1$ neutral expression and $11$ facial expression ($9$ for the train set and $2$ for the test set, if not specifically explained). It needs to be stressed that two of our design expressions (smile and mouth-open) are treated as the test expression, with the motivation that the smile and mouth-open are used to test extrapolation and interpolation of the benchmarks respectively. The expression capture example is visualized in Figure~\ref{fig:exp_capture}.

\noindent \textbf{Hair.} The design of the hair collection consists of three aspects -- original outfit capture, 3D face capture (with hair cap to hide hair), and wig capture.  Specifically, for the original outfit capture setting, each subject is captured with his/her original hairstyle. For performers dressed in different eras, the collection of 3D face capture and wig are skipped due to the inconvenience of wearing a wig or hair cap on the head with already wearing many different accessories. For the normal performers, one video of wearing the hair cap is captured and then the wig part follows. We prepare wigs with $7$ daily styles (~`Men's straight short hair', ~`Men's curly short hair', ~`Women's bobo hair', ~`Women's pear curls', ~`Women's long curls', ~`Women's long straight hair', and ~`Women's small curls'), and $6$ color tones (black, blue, brown, green, gold and yellow). During the collection, the subject is asked to turn around his head in a whole circle. Such a design can benefit the emphasizing of the dynamic motion that relates to the wig. Different wig styles, colors, and head motions are visualized in Figure~\ref{fig:hair_capture}.

\noindent \textbf{Speech.} Since the subjects consist of four ethnicities, we provide the speech corpus in two languages, Mandarin for Chinese and English for the others, and two versions. In the first version, each subject speaks $42$ sentences, which consist of sentences and short paragraphs. For Mandarin sentence design, we select $30$ phonetically balanced sentences from ~\cite{mandarin_corpus} as our main part, and $10$ sentences combined with single words from ~\cite{monosyllable_mandarin_chinese} in order to cover all the consonants, vowels, and tones. The composition of English sentences is similar to VOCASET~\cite{cudeiro2019capture}, whose main part is $40$ phonetically balanced sentences, the same as VOCASET. Two short paragraphs are both added to the Mandarin and English collections as a supplement for continuous long-time talking. Each subject has the same corpus in the first version. 

In the second version, we shorten the total number of sentences from $42$ to $25$ in order to speed up the collection. Moreover, we randomly sample the sentences from the corpus for each subject so as to improve differentiation. For Mandarin, we first cut the single words-combined sentences from $10$ to $5$ but still keep their coverage of consonants, vowels, and tones. Then the main part, $30$ phonetically balanced sentences, is shortened to $20$, which consists of $10$ fixed and $10$ flexible sentences. Finally, we randomly sample one paragraph from the original two. As a result, we get $26$ sentences in total for each subject. For English, the main part, $40$ phonetically balanced sentences, are shortened to $25$, which consists of $15$ fixed and $10$ flexible sentences, and the paragraph part is processed the same as in Chinese. Since we have $500$ identities in total, about $150$ Chinese and $150$ Foreigners are captured with the first version and the rest with the second version.

\begin{figure}
    \centering
    \includegraphics[width=1\linewidth]{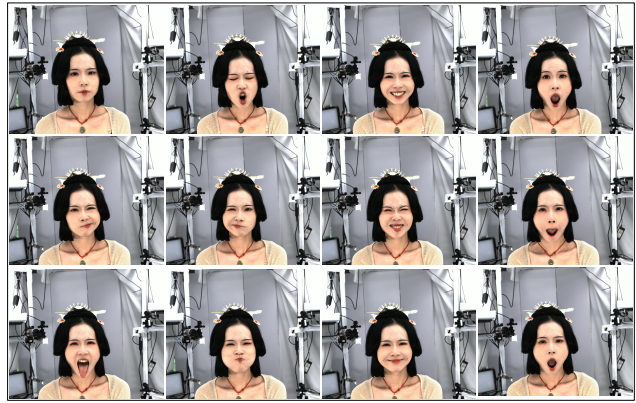}
    \caption{\textbf{Expression Capture.} We capture 12 expressions, containing 1 expressionless and 11 exaggerated expressions.}
    \label{fig:exp_capture}
\end{figure}

\begin{figure}[t]
    \centering
    \includegraphics[width=1\linewidth]{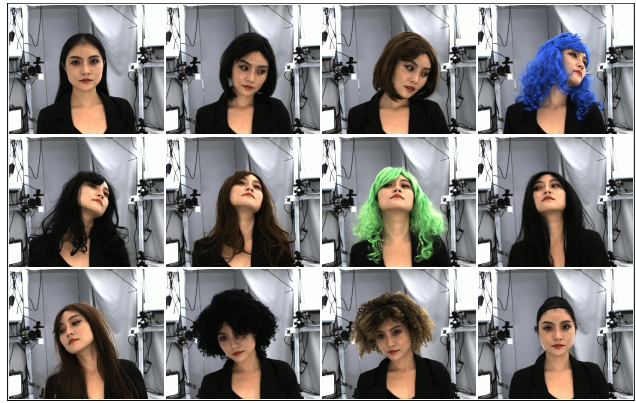}
    \caption{\textbf{Hair Capture.} We capture $12$ hairstyles for each subject, which includes one original hairstyle, one wearing mesh hood, and ten wearing wigs. The ten wigs are randomly picked from our wig set. We ask the participant to turn the head clockwise with different hairstyles.}
    \vspace{-0.5cm}
    \label{fig:hair_capture}
\end{figure}

\subsubsection{Collection Protocol}\label{protocol}
As the dataset collection spans over months, to guarantee the accuracy of data collection, we design a collection protocol and execute it before every capture. The protocol consists of three steps, \ie, pre-collection check, collection, and post-collection check. 

\noindent \textbf{Pre-Collection Check.}
To ensure proper operability of equipment and accurate camera position, two steps of inspection are applied:

\noindent
1) \textit{Hardware Check.} We manually check the status of all computers and cameras and make sure that all $60$ video stream is ready-to-work and synchronized by testing collection. We prepare backup cameras for the broken ones.

\noindent
2) \textit{Fake Head Capture.} We put a fake head in the middle of the view and keep it static, and then capture one frame of all $60$ cameras. Then we check all the frames, when the head offsets the imaging center, the pose of the correspondent camera needs to be fixed. The sharpness of the images is also checked in case one or part of the cameras are not focusing on the head.

\noindent \textbf{Collection.} The main collection consists of four parts:

\noindent
1) \textit{Camera Calibration.} A chessboard is held and turned around for $3$ circles, then every camera can capture data with the chessboard in various poses. The data is used for calculating the camera parameters (intrinsic and extrinsic).

\noindent
2) \textit{Expression Capture.} Each subject's expression metadata is collected with $12$ facial expressions. Each expression collection lasts about $3$ to $5$ seconds and the performer starts with the neutral expression, changes continuously to designated expressions, and then keeps the performance unchanged until this collection finish. Substandard or incorrect expressions will be discarded and re-recorded.

\noindent
3) \textit{Hair Capture.} The hair collection is separated into three parts: origin hair, hair cap, and wig capture. One video for the origin hair and one for the hair cap are captured for each subject. In these two parts, the subject always keeps still with eyes straight ahead. Then the wig part collection begins and we collect about $10$ videos for wigs with random hairstyles and colors. Generally each subject cover about $4$ wig styles and $3$ wig colors. In the wig collection, the performer starts with his head in the middle of the view and eyes straight ahead, then cranes his neck $360$ degrees, relaxing it as usual but with as much amplitude as possible. When finishing the whole process, the subject returns to the original status and waits for the end of this part. We'll record it again when insufficient head rotation appears.

\noindent
4) \textit{Speech Capture.} We prepare a large corpus in two languages (Mandarin and English) for each subject. The whole speech collection is split into $4$ or $6$ parts according to the number of sentences. In each collection, the performer is asked to read the sentences which are shown on a screen and the collection lasts about $30$ to $40$ seconds. We do not require a standard mouthpiece but mispronunciation is not allowed.

\noindent \textbf{Post-Collection Check.}
A script is applied to concatenate and visualize the multiview video synchronously. All the collected data is processed and checked manually to filter out source data issues. We demonstrate the necessity and importance of the data post-collection check with extensive trial and error experiences.

After the above processes finish, we obtain a large-scale dataset of $500$ identities. Each identity is guided to perform $12$ expressions, $25$ to $42$ sentences, and more than $10$ hair collections. 

\subsection{Data Annotation Details}
\label{sec:annote}
We obtain the raw data of RenderMe-360 from the collection pipeline with POLICY. Then, we annotate the data to get rich annotations with the processes described below.

\subsubsection{Camera Parameter Annotation}
Camera calibration is the basic step for fine-grained annotation in a multi-view capture system. The process in our pipeline is visualized in Figure~\ref{fig:anno_calib_lmk}. To make sure the availability and accuracy of the parameters, two checking procedures are performed besides basic camera pose estimation. First, we apply fast NeRF model training of Instant-NGP~\cite{muller2022instant} via feeding all the camera views. We render images with the same views and manually check for potentially unreasonable rendering results caused by wrong extrinsic parameters. Secondly, we perform the keypoint annotation process with the same frames and re-project the 3D facial landmarks to manually check for the out-of-face result. The unqualified results will loop in re-calibration process.

\begin{figure}
    \centering
    \includegraphics[width=1.0\linewidth]{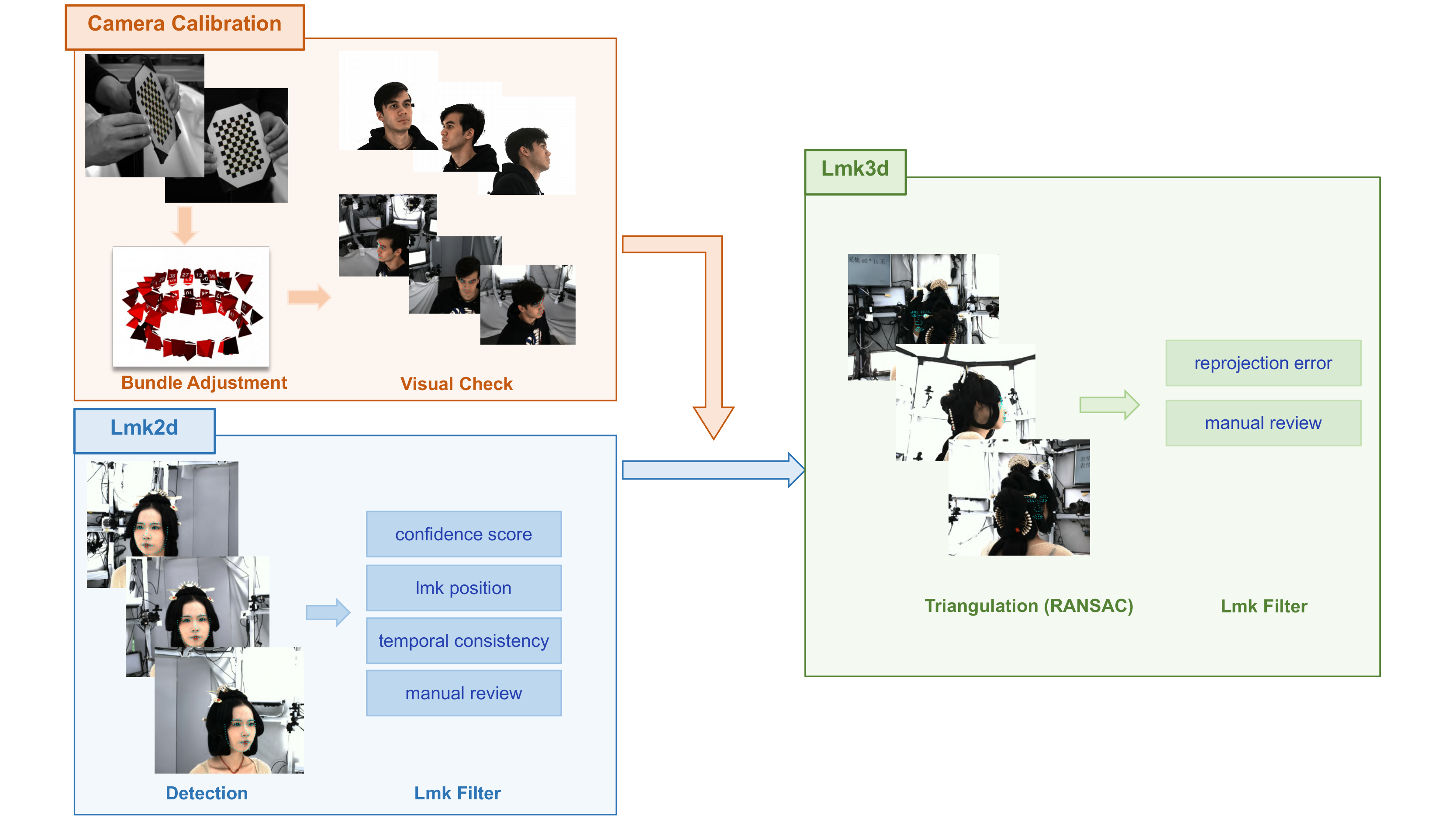}
    \caption{\textbf{Camera Calibration and Keypoint Detection.} The camera calibration process contains chessboard data collection, calibration with bundle adjustment, and visual check. After the detection and filtering of the multi-view 2D landmarks, the 2D landmarks result, together with the camera parameters, are utilized to triangulate for robust 3D landmarks. }
    \label{fig:anno_calib_lmk}
\end{figure}

\subsubsection{Facial Keypoint Anotation}
To filter out abnormal 2D landmarks and precisely triangulate to get robust landmark 3D, we apply the following rule-based and heuristic rules. $1)$ We use a enhanced version of facial landmark detection model~\cite{wu2018look}, and discard the result with a low confidence score. $2)$ Since some unqualified landmark results have an abnormal scale or location, we heuristically set thresholds for the largest distance between landmarks and the mean location. $3)$ As there is no large head motion in the expression capture stage,  we consider the temporal consistency of the detected landmarks and filter out the case with an overall offset of the keypoints. $4)$ We manually check the data to select inaccurate landmark results. We make sure that data of at least $3$ views are applied to do the triangulation, and check the reprojection error in all $60$ views. When a significant location error or an abnormal reprojected location is detected, we manually label all 2D landmarks and re-run the triangulation process for an accurate result. 

\begin{figure}[t]
    \centering
    \includegraphics[width=1.0\linewidth]{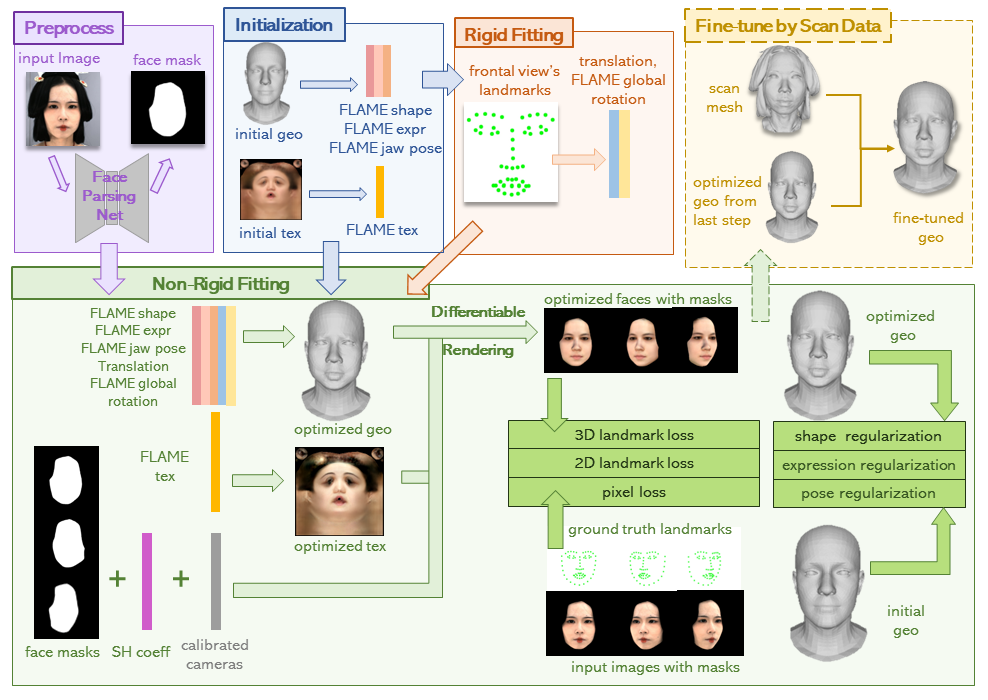}
    \caption{\textbf{FLAME Fitting.} The fitting pipeline is focused on the subject's face region, and face masks for each view are preprocessed. Rigid fitting aims to solve translation and rotation roughly with 2D and 3D landmarks, values are improved in the non-rigid fitting. Non-rigid fitting optimizes FLAME's other parameters as well, but mainly on shape, expression, jaw pose and texture parameters to ensure better identity likeness of final geometry. The last fine-tuned step is not necessary for all frames, frames without scan mesh are optimized based on frames with it.}
    \label{fig:flame_pipeline}
\end{figure}

\subsubsection{FLAME Fitting} 
The overall pipeline for FLAME fitting is illustrated in Figure \ref{fig:flame_pipeline}. Raw captured images are first processed via masking out the background and non-facial head regions, in order to avoid fitting distractions. Then, a rigid fitting is applied to get rough values of translation and global rotation. Concretely, the 2D and 3D facial landmarks are both involved in this process. We use $51$ facial landmarks due to the non-differentiable attribute of contour landmarks trajectory. 2D landmarks from the frontal views are used for rough estimation, and 3D landmarks are used for anchor 3D position. For the rigid fitting, the optimizing target can be viewed as 
\begin{align}
    \mathcal{L}_{\text{rigid}} = \|\text{lmk}_{2d}- {Proj}(R\cdot\text{lmk}_{\text{flame}}+t)\|
\end{align}
where $\text{lmk}_{2d}$ is the detected 2D landmarks, $\text{lmk}_{\text{flame}}$ is the marked corresponding landmarks on the FLAME model, and $R$,$t$ are the variables to be optimized, the loss is calculated through all frontal views and all $51$ facial landmarks.

Non-rigid fitting is further applied to improve translation/global rotation, FLAME shape, expression, jaw pose, and texture parameters. We utilize landmarks in both 2D and 3D to constrain the optimization. Since 3D landmarks provide one more dimension value (\textit{i.e.,} z value), while having a shortage of good face contour information. Thus, 2D landmarks around face contours are needed to improve shape. Moreover, with calibrated cameras, we are able to render geometry and texture in image space by using differentiable rendering and comparing pixel differences with input images. However, texture parameters only map to albedo map based on texture basis, and skin tone from input images is affected by environment lighting conditions. Thus, optimized spherical harmonics (SH) coefficients are needed to adjust rendered faces. To ensure the reasonability of optimized geometry,  we provide shape, expression, and pose regularizations to avoid broken geometry. Scan meshes show accurate facial shapes in world space, so a FLAME fitting process with scan can preserve better facial edges and corners, but not all frames are grouped with it. As shown in Figure \ref{fig:flame_pipeline}, the fine-tuned step is surrounded with dotted lines, indicating that it is not necessary for all frames and is only applied on frames with scan meshes to do further improvement. This strategy is useful for personalization and getting expression prior knowledge for non-neutral frames without scan.  In a nutshell,  the full loss function can be formulated as 
\begin{align}
    \mathcal{L}&= \mathcal{L}_{\text{lmk}} + \mathcal{L}_{\text{scan}}
                + \mathcal{L}_{\text{pix}} + \mathcal{L}_{\text{reg}}
\end{align}
\begin{align}
\small
    \mathcal{L}_{\text{lmk}}&=
    \|\text{lmk}_{2d} - Proj(R\cdot\text{lmk}_{\text{FLAME}(s,e,p)} + t)\|\nonumber\\& +
    \|\text{lmk}_{3d} - R\cdot\text{lmk}_{\text{FLAME}(s,e,p)} - t\| \\
    \mathcal{L}_{\text{scan}}&=
    \min_{i\in\text{scan}}\|v_{i} - R\cdot v_{\text{FLAME}(s,e,p)} - t\| \\
    \mathcal{L}_{\text{pix}} &=
    \|\text{rgb}_{Proj(R\cdot v_{\text{FLAME}(s,e,p)})} \nonumber\\
		&- \text{tex} * (\gamma \cdot \text{SH}(n_{\text{FLAME}(s,e,p)}))\| \\
    \mathcal{L}_{\text{reg}} &=
    \left\|\frac{s}{\sigma_{s}}\right\| + \left\|\frac{e}{\sigma_{e}}\right\| + 
    \left\|\frac{p}{\sigma_{p}}\right\|
\end{align}
where landmark loss includes 2D detected ones and 3D triangulated ones. Scan loss includes the nearest point on scan with each FLAME vertex, which is only calculated at the last frame of each sequence. For rendering, we calculate the RGB value at each float position with bilinear interpolation within the face mask with rendered vertices using face normals $n_{\text{FLAME}}$ and spherical harmonic lighting $SH$. Regularization terms include shape parameter $s$, expression parameter $e$, and poses $p$ for jaw, neck and eyes.

We assume frames of neutral sequences are always neutral (expression parameter $s$ and pose parameter $p$ are zero), sequences with non-neutral expressions start with neutral and end with exaggerated expressions. Dense mesh reconstruction is at least applied on the last frame to generate scan mesh for each expression sequence. The personalization step is inspired by ~\cite{FLAME:SiggraphAsia2017}. It starts with FLAME basis as an initial value, as shown in the left image of Figure \ref{fig:flame_init}, optimizes FLAME parameters, and is fine-tuned with the help of scan mesh to get an accurate face shape template. With a personal template provided, as shown in the middle image of Figure \ref{fig:flame_init}, non-neutral frames' fitting won't optimize shape parameters anymore, and we solve the last frame paired with scan mesh firstly and puts more effort into other parameters to ensure face expression as vivid as the input image. Due to the assumption mentioned above, frames in between the first frame and the last frame are performed with linear interpolation to get a rough initial value, as shown in the right image of Figure \ref{fig:flame_init}. For the purpose of ensuring the annotation to the full extent of accuracy, the human annotators are asked to identify and rectify inaccurate annotation results of FLAME. The annotators must identify and select the incorrect results, and then we provide the necessary refinement to generate the accurate 3D head model.

In addition to FLAME fitting annotation, we also provide the UV texture map as  an extra annotation upon the fitting. Specifically, since it is low quality and has few details, instead of using an albedo map optimized from our fitting pipeline, we take view-dependent texture maps unwrapped from captured images of selected views and composited them together with Poisson blending~\cite{perez2003poisson} to create the final high-quality texture map in Figure \ref{fig:flame_tex}. 

\begin{figure}
    \centering
    \includegraphics[width=1.0\linewidth]{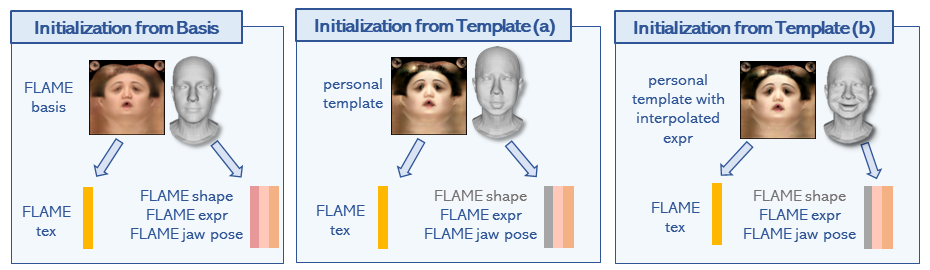}
    \caption{\textbf{Initialization Modes for FLAME Fitting.} There are three initialization modes according to different fitting purposes. Initialization from the basis is designed for getting a personal template. Initialization with the template is to fix shape parameters and do expression fitting, (a) is for frames with scan mesh, (b) is for frames without. }
    \vspace{-0.5cm}
    \label{fig:flame_init}
\end{figure}

\begin{figure}
    \centering
    \includegraphics[width=1.0\linewidth]{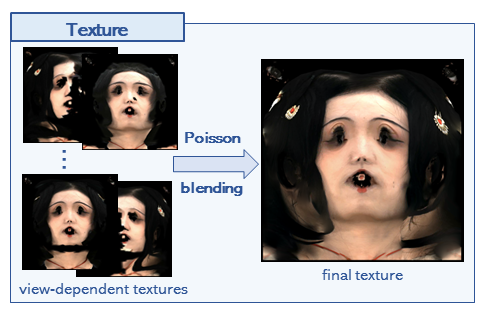}
    \caption{\textbf{Final Texture Map.} View-dependent texture maps are selected and composited together with Poisson blending to create the final full texture map as the UV map annotation.}
    \vspace{-0.5cm}
    \label{fig:flame_tex}
\end{figure}

\begin{figure} 
    \centering
    \includegraphics[width=1.0\linewidth]{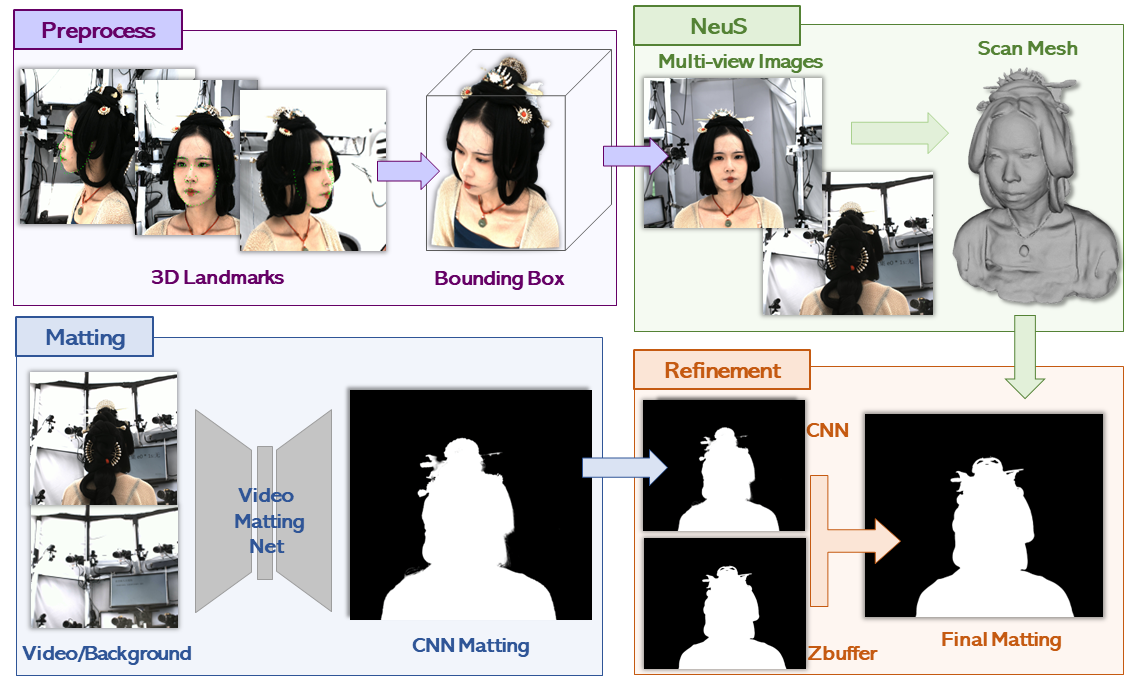}
    \caption{\textbf{Dense Mesh Reconstruction and Matting.} Dense mesh reconstruction is supported by NeuS, it builds models for the subject(foreground) and background separately, bounding box is estimated by robust 3D landmarks for better separation. The final matting result is refined with a Z-buffer value. This is applied for refining the situation when the mask predicted from the video matting network cannot well handle detailed head accessories.}  
    \vspace{-0.5cm}
    \label{fig:scan_matting}
\end{figure}

\begin{figure*}[htb]
    \centering
    \includegraphics[width=1\linewidth]{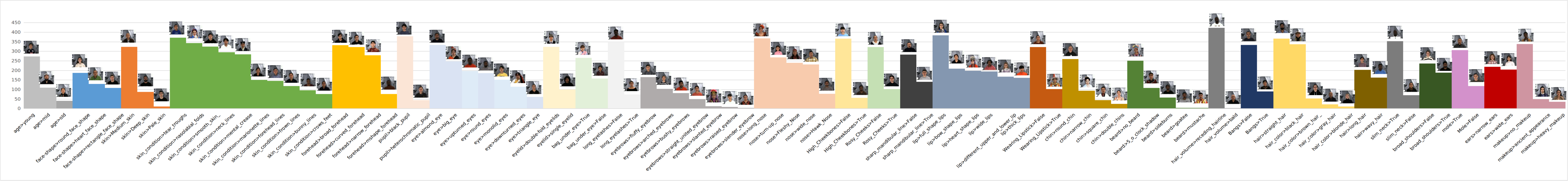}
    \caption{\textbf{Statistical Chart of Static Facial Features.} The properties that lie in the same attribute group of facial features are highlighted in the same color. An exemplar image of each attribute is shown in the corresponding histogram column. We use ``\textgreater'' to split the group and attribute. }  
    \label{fig:att_annot}
\end{figure*}

\subsubsection{Scan and Matting Refinement}
The processing pipeline is illustrated in Figure~\ref{fig:scan_matting}.
\noindent
\textbf{Scan.} Specifically, we apply NeuS~\cite{wang2021neus} to multi-viewed images with known camera intrinsics and extrinsics. In practice, a rigid transformation is estimated from landmarks of a standard FLAME model to target detected 3D landmarks from triangulation. Then the bounding box of the head region is assumed to be $2$ times the bounding box of the FLAME model. We follow the setting assuming that a background NeRF~\cite{mildenhall2020nerf} modeling the rendered results outside the bounding box and a NeuS~\cite{wang2021neus} modeling radiance field inside the bounding box. Both are modeled as an $8$-layer multi-perceptual network (MLP) with skip connections in the $5$-th layer, and the inputs are coded with positional encoding. For each video sequence, we apply this algorithm to the first frame and train from scratch to get the neutral scan mesh. For the following frames, we pick the keyframe where the expression seems to be the most exaggerated, add fine-tune to the static model to get a similar scanned result, where the bounding box is fixed as the first frame. 

\noindent
\textbf{Matting.} As for the matting annotation, a static background is captured before the formal recording of each round. Then, we use a video-based matting method~\cite{DBLP:conf/wacv/LinYSS22} to estimate the foreground map of each image.
To further improve matting accuracy, we additionally tailor the depth information into the pipeline. Concretely, we rasterize the scanned mesh to each camera view, and use this geometry prior to refine the video-based matting estimation, with graphical-based segmentation. Grabcut~\cite{rother2004grabcut} is used with the intersection of both masks as the absolute foreground and areas outside the union with a fixed size of padding as the absolute background. We calculate Bayesian posterior for each pixel as the alpha value. We further employ human annotators to identify and rectify inaccurate annotation results of scan and matting. Then we provide the necessary parameters to generate the accurate dense mesh or manually label the foreground to yield precise matting maps.

\begin{figure}[htb]
    \centering
    \includegraphics[height=0.7\textheight]{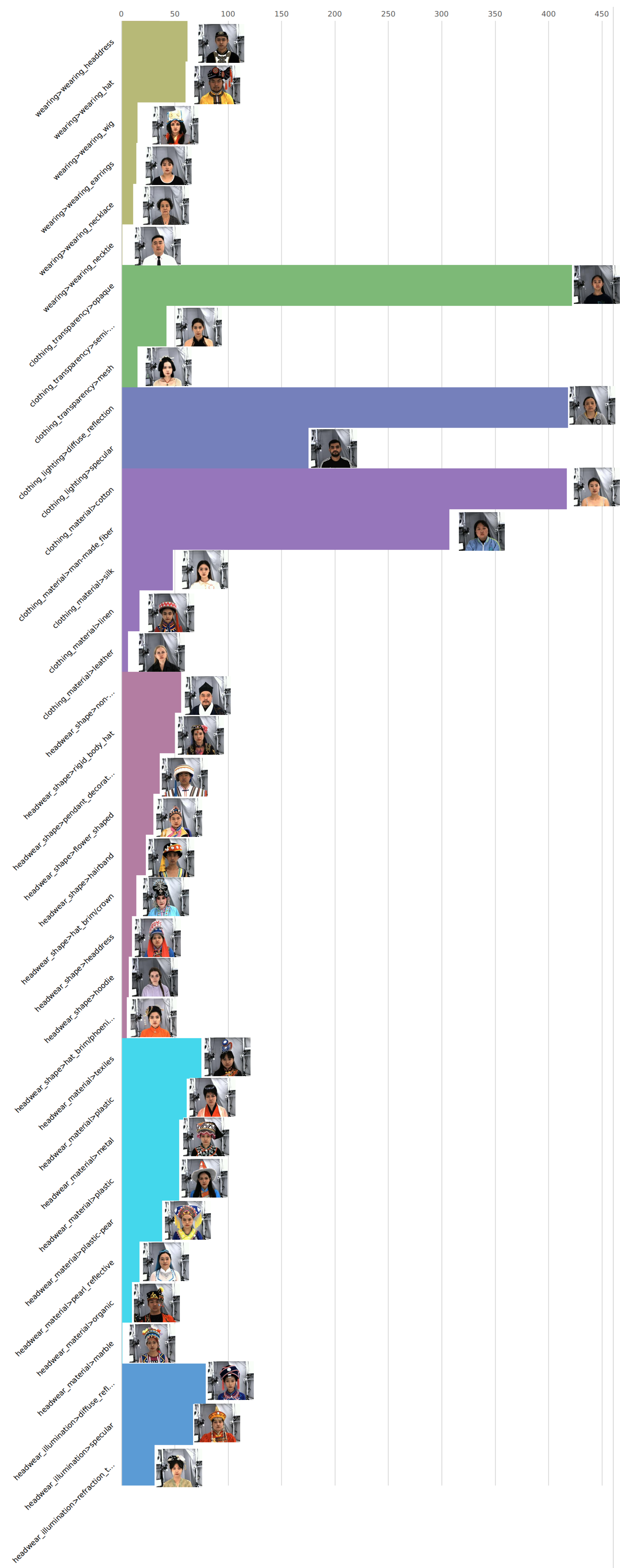}
    \caption{\textbf{Statistical Chart of Static Information of Non-facial Regions.} he properties that lie in the same attribute group of non-facial features are highlighted in the same color. An exemplar image of each attribute is shown in the corresponding histogram column. We use ``\textgreater'' to split the group and attribute. }  
    \label{fig:att_annot2}
\end{figure}

\subsubsection{Text Annotation}
\label{sec:text_annot}
Both static and dynamic text-based descriptions are involved in our text annotation to further facilitate multi-modality research on human head avatar creation. The text combines four types of annotations: \textit{static facial features}, \textit{static information of non-facial regions}, \textit{dynamic facial actions}, and \textit{dynamic video activity descriptions}. With these four aspects of text annotation, we could provide a comprehensive description of each human head to boost various downstream tasks. 

\noindent\textbf{Static Facial Features.} This aspect of text annotation seeks to comprehensively detail attributes of the subject's facial features in facial regions. Based on the text annotation proposed in CelebA~\cite{liu2018celeba}, we further annotate new facial features, with extending the original $40$ annotations to $95$ common fixed types of facial attributes and $2$ non-fixed text-based salient attributes. The fixed facial attributes refer to the universal and frequent properties, which are annotated through pre-defined attribute item selection. The non-fixed attribute provides flexible supplemental additions to the $95$ fixed attributes, which aim at encompassing a broader range of facial depictions and is annotated through natural language. The combination of fixed attribute and non-fixed attribute annotations could outline human faces with more complete and precise text descriptions than the original category definition in CelebA. 

Specifically, the fixed facial attributes and the corresponding example images are illustrated in Figure~\ref{fig:att_annot}. For every attribute, we employ five annotators to vote on whether the collected subjects contain the particular attribute, and the final annotation is determined by the majority decision.  In particular, we carefully analyze common facial traits, and divide these $95$ facial attributes into $28$ major groups, including facial properties like face shape, skin condition, eye shape, eyebrow shape, lip shape, nose shape, hair shape, etc. Each major group of facial features contains several detailed shape attributes. Compare with the original facial attributes of CelebA~\cite{liu2018celeba}, we introduce more facial feature attributes to describe facial features in detail. For instance, CelebA only defines one single label for eyes, namely ~``narrow eyes'', we provide more variant shapes for comprehensive depictions, including ``almond eyes'', ``big eyes'', ``upturned eyes'', ``round eyes'', ``monolid eyes'', ``downturned eyes'' and ``triangle eye''.  More examples like the skin condition, a newly introduced property group, is a significantly conspicuous facial attribute and has been ignored by CelebA. For this group, we describe it with several detailed attributes, containing ``tear troughs'', ``nasolabial folds'', ``neck lines'', ``mental creases'', ``marionette lines'', ``forehead lines'', ``frown lines'', ``bunny lines'', ``crows feet'' and ``smooth skin''. Through such a fine-grained category enrichment, a fixed common types annotation with $95$ attributes of facial attributes is constructed. 

In addition, we provide two non-fixed attributes: \textit{the salient facial feature}, which describes significant attributes of the facial features, and \textit{ the salient features of the makeup}, which depicts the significant features of the makeup styles. The two attributes do not overlap with any of the fixed attributes. We require annotators to observe the overall features of the subject and describe salient features of the subject's face and makeup style in natural language. The annotated descriptions from 5 annotators are collected and manually removed redundant or nonexistent attributes to yield the final annotation. For example, the salient facial attribute of Figure~\ref{fig:exp_capture} is that \textit{she possesses visible collarbones with a mole above the left eyebrow, round pupil, multiple eyelids, slightly flattened eyebrows, pale forehead, and applies light foundation, draws long and thin eyebrows, wears petal-like lipstick with pink eyeshadow and black mascara.} This flexible attribute further complements salient facial features based on subjective observations, including some color, position and shape of facial features, and some attributes not covered by fixed attributes. 

\noindent\textbf{Static Information of Non-Facial Regions.} This aspect of text annotation aims to depict the attributes of the subjects' non-facial regions, such as the tops of outfits and accessories. In addition to the attributes of inherent facial features, we also consider static information of non-facial regions. Since these properties are distinctive to describe different human heads. We focus on the material, shape, color, and lighting conditions of the subjects' wearing accessories. For holistic head rendering, information on non-facial regions is also critical. However, few studies have involved annotation of these parts, with most research focusing solely on the labels of static facial features. While static facial features have been a primary focus for modeling human appearance, additional qualities corresponding to the wearing elements promote photorealism.  Unprecedentedly, we introduced annotations related to these aspects. By including non-facial attributes in our annotation, we provided a broader, and more integrated knowledge to model human heads in their full individual characteristics.

\begin{table}
\centering
\caption{\textbf{Action Units of Expression.} Each of the collected expressions (Exp) is defined as a set of AUs. Please note that *Exp-4 is a left-toward expression while *Exp-5 is a right-toward expression, and they contain the same set of AUs.}
\label{tab:facs}

\resizebox{\linewidth}{!}{
\begin{tabular}{@{}cl@{}}
\toprule
Expression No. & Action Units                                        \\ \midrule
Exp-1         & AU-18, AU-22, AU-25, AU-27, AU-43                   \\
Exp-2         & AU-6, AU-12, AU-13, AU-14, AU-25, AU-26, AU-27      \\
Exp-3         & AU-1, AU-5, AU-25, AU-26, AU-27                     \\
Exp-4*         & AU-4, AU-6, AU-9, AU-11, AU-13, AU-14, AU-17, AU-44 \\
Exp-5*         & AU-4, AU-6, AU-9, AU-11, AU-13, AU-14, AU-17, AU-44 \\
Exp-6         & AU-4, AU-7, AU-9, AU-10, AU-15, AU-25, AU-41        \\
Exp-7         & AU-16, AU-25, AU-26, AU-28                          \\
Exp-8         & AU-13, AU-25, AU-26, AU-27                          \\
Exp-9         & AU-13, AU-17, AU-18, AU-23                          \\
Exp-10        & AU-6, AU-12, AU-13                                  \\
Exp-11        & AU-25, AU-26, AU-27                                 \\ \bottomrule
\end{tabular}
}
\end{table}

\begin{figure}[htb]
    \centering
    \includegraphics[width=1\linewidth]{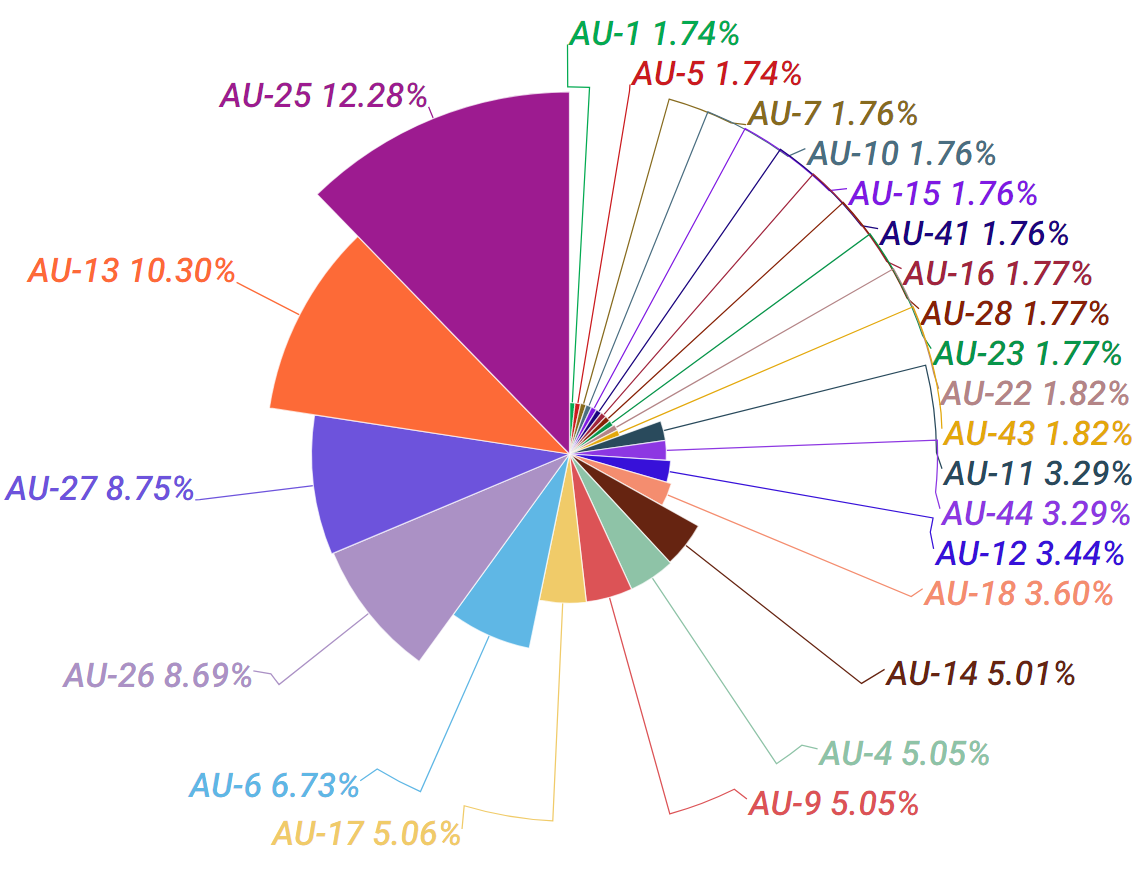}
    \caption{\textbf{Statistical Chart of Dynamic Facial Actions.} We illustrate the proportion of every AU in our text annotation data of dynamic facial actions.}
    \label{fig:facs}
\end{figure}

Similar to static facial features, we provide two types of annotation, including fixed attributes and non-fixed attributes. As shown in Figure~\ref{fig:att_annot2}, the fixed attributes contain $36$ attributes derived from 7 major groups, such as accessories shape, clothing transparency, headwear shape, {\textit{etc}}. For every attribute annotation, we require five annotators to label whether the subject has the attribute. The final annotation of this attribute is voted by the majority choice. Additionally, the annotators are required to describe the non-fixed attributes in natural language. The non-fixed attributes contain $1)$ \textit{ the color of the tops of the outfits}, in which the annotators describe the colors and in the order from large areas to small areas; $2)$~\textit{the color of the head accessories}, in which the annotators mark the colors included in the order from large areas to small areas; and $3)$ \textit{the salient features of the accessories}, in which the annotators describe the significant features of the accessories. For instance, the non-fixed attributes in Figure~\ref{fig:exp_capture} are that $1)$ \textit{her tops of the outfits are yellow and black}, $2)$ \textit{her accessories are multiple colors of golden, white, blue and red}, and $3)$ \textit{she wears an ancient jade pendant and a golden hairpiece with red stones and a blue circlet in the crown}. There are no overlapped descriptions between non-fixed attributes and fixed attributes. The proposed text annotation on static information of non-facial regions involves diverse and rich descriptions for the non-inherent attributes, which could promote text-aware generation with detailed and high-fidelity textures.

\noindent\textbf{Dynamic Facial Actions}
The text annotation of dynamic facial actions refers to explicitly describing the dynamic changes in the local facial features of the collected subjects at each timestamp.
Here, we only focus on the collected expression-related videos and ignore speech-related and wig-related videos because expression-related videos already contain a large number of dynamic changes in local facial features.

Based on Facial Action Coding System, FACs~\cite{ekman1978facial}, facial expression can be described into specific action units (AUs), which are the fundamental facial actions of individual muscles or groups of muscles. The detailed descriptions of each AU can be found in \url{https://www.cs.cmu.edu/~face/facs.htm}. Each of the 11 collected expression categories can be further divided into a set of multiple action units (AUs), as shown in Table~\ref{tab:facs}. We provide AU annotations for each frame of changes in expression videos. As shown in Figure~\ref{fig:facs}, we statistically analyzed the proportion of each AU category in the annotations. AU-25, representing that the lips part, appears most frequently, accounting for 12.28\%, while AU-1, representing that the inner brow raise, appears least frequently, only 1.74\%. The top 3 most prevalent AUs are AU-25 (lips parting), AU-13 (cheek puffing) and AU-27 (mouth stretching), while the least prevalent top 3 AUs are AU-1 (inner brow raising), AU-5 (upper lid raising) and AU-7 (lid tightening). It indicates that our dataset encompasses more extensive mouth movement variations, which are significant facial motions while paying comparatively little attention to subtle brow and lid regions motions.

\noindent\textbf{Dynamic Video Activity Descriptions.} 
The text annotation of dynamic video activity descriptions is video-linguistic annotation and aims to globally describe the overall activity of the subjects in the collected videos in complete sentences. 

To globally describe facial activity with diversity, four annotators were employed to introduce each video action from four different perspectives: dynamic changes in facial actions, dynamic changes in facial state, dynamic changes in facial features, and dynamic changes in facial muscles. We collected videos in three scenarios: expressions (Exp), hairstyles (HS) and speeches (Sp). Thus, each video has a corresponding template, and the annotators describe each video type from the collection templates, allowing us to obtain text descriptions for each video type. The descriptions of the actions performed by the subject can be found in Figure~\ref{fig:text_sentence}. Each type of action has four corresponding descriptions. In particular, for hairstyle videos, we describe wig color, shape, texture, etc., which does not overlap with our previous annotations, since the previous annotations did not involve wigs. For every individual video, providing merely a subject (i.e., ``a man" or ``a female") and integrating this with the relevant template of dynamic action descriptions yields a complete descriptive sentence.

As shown in Figure~\ref{fig:text_sentence}, we provide comprehensive and diverse video activity descriptions composed of user-friendly natural language sentences, which can facilitate video generation or video editing. 

\begin{figure}[htb]
    \centering
    \includegraphics[width=0.95\linewidth]{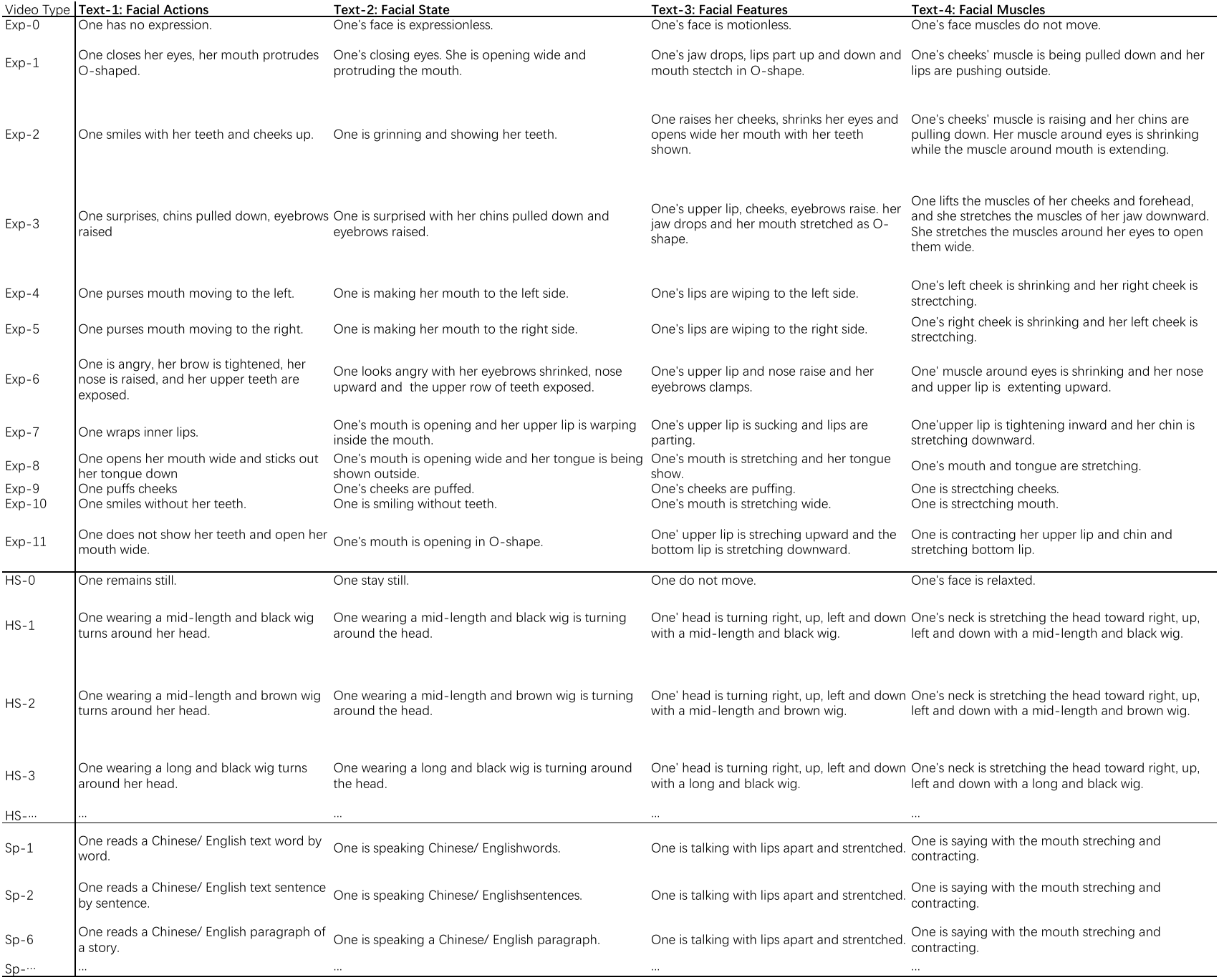}
    \caption{\textbf{Dynamic Video Activity Descriptions.} We provide four perspectives of text descriptions about each video type's activity. Exp refers to expression-based video, HS refers to hairstyle-based video, and SP refers to speech-based video. ``One'' can be replaced by a subject.}
    \label{fig:text_sentence}
\end{figure}

\begin{figure*}[htb]
    \centering
    \includegraphics[width=1.0\linewidth]{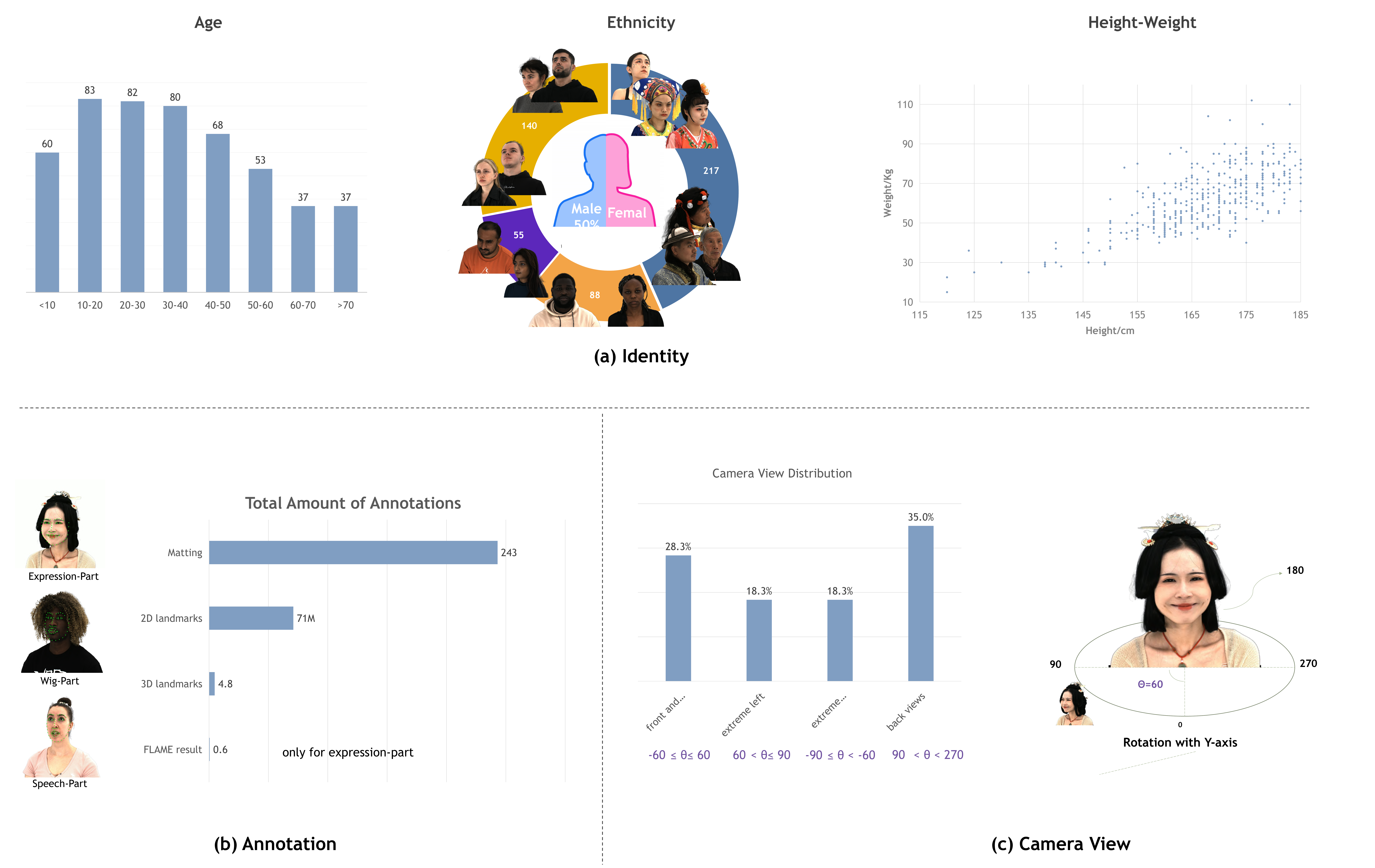}
    \caption{\textbf{General Data Distribution.} The data is summarized in three aspects, identity attributes, annotation, and camera view.}
    \vspace{-10pt}
    \label{fig: statistic 1}
\end{figure*}

\begin{figure*}[htb]
    \centering
    \includegraphics[width=0.95\linewidth]{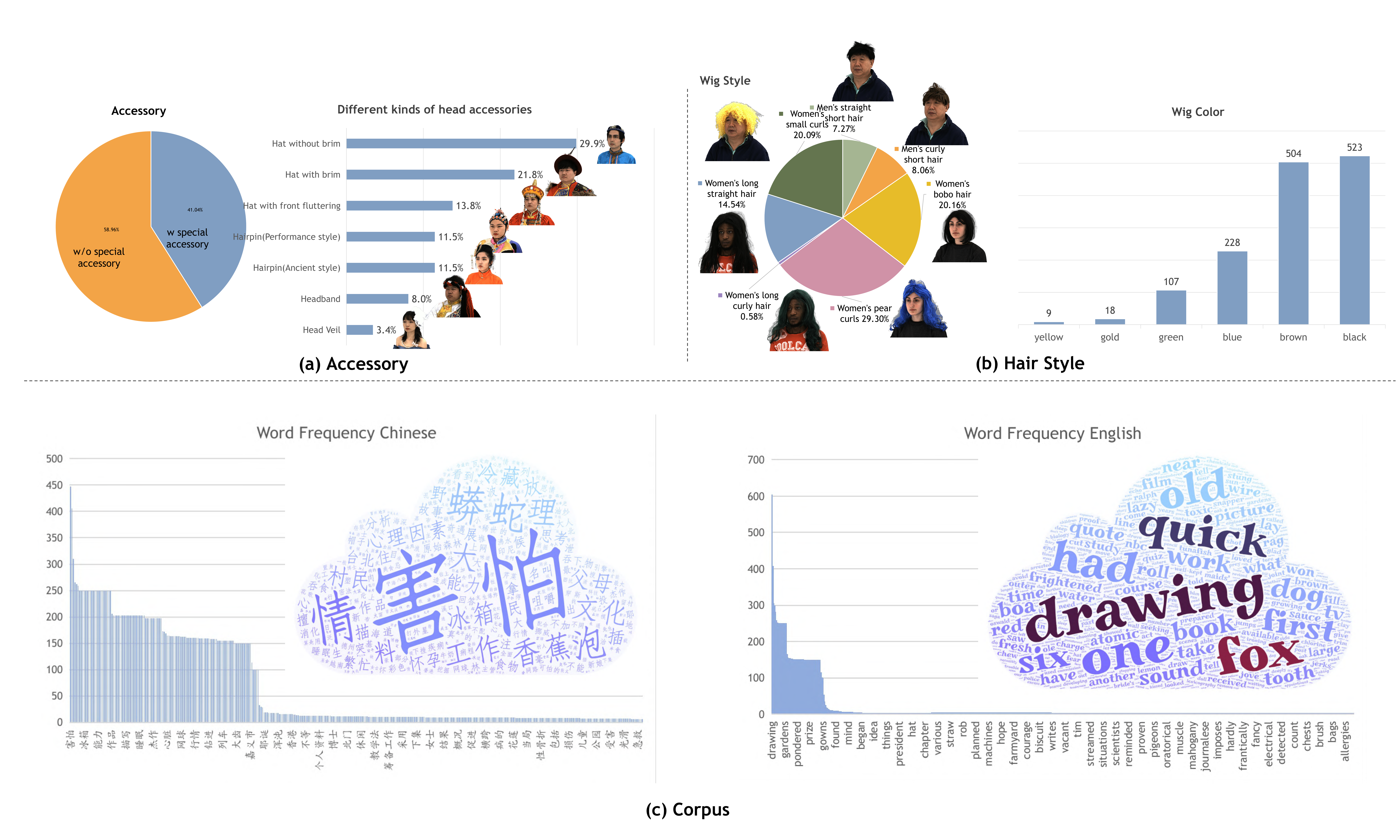}
    \caption{\textbf{Collection Statistic.} We demonstrate the collection statistic on three sides, namely accessory, wig, and corpus.}
    \label{fig: statistic 2}
\end{figure*}

\subsection{Dataset Statistics Details}
\label{sec:statiscs}
Since RenderMe-360 is a large-scale head dataset with multiple data, identity, and annotation, we make the statistic analysis into six aspects as below.

\noindent \textbf{Identity.}
As shown in Figure~\ref{fig: statistic 1} (a), we summarize data of captured identities in four dimensions, including age, height-weight, gender, and ethnicity. The subjects' ages range from  $8$ and $80$ with approximate normal distribution, where teenagers and adults form the major part. A relatively large number of children and the elderly increase diversity of our assets. We show a height-weight distribution map, which indicates a large part of the models is located in height between $155$cm and $185$cm, and weight between $50$kg and $90$kg. Notably, the recorded height and weight data can support the physical nature perception of humans, which is an important question in commonsense reasoning.
Our dataset is gender-balanced and divided into $4$ ethnicities ($217$ Asian, $140$ White, $88$ Black, and $55$ Brown). Ethnic diversity poses significant challenges and helps explore the margin and limitations of head avatar research.

\noindent \textbf{Annotation.}
As mentioned before, we obtain a dataset with more than $243$M frames which are fine-grained annotated. As Figure~\ref{fig: statistic 1} (b) shows, there are three data collection parts of RenderMe-360, including Expression-Part, Wig-Part, and Speech-Part. Since frames in all the collection parts are annotated, there have over $243$M frames with matting, $71$M frames with 2D landmarks, and $4.8$M frames with 3D landmarks. Since only frames in the expression collection are annotated with FLAME, we have $0.6$M FLAME result in total. Besides, we also provide UV maps, AUs, appearance annotation, and text annotation. Rich and multimodal annotation provides more possibilities for downstream research and application.

\noindent \textbf{Camera View.}
Since the POLICY contains $60$ cameras which form $4$ layers, we demonstrate the camera view distribution in Figure~\ref{fig: statistic 1} (c). Camera views are divided into $4$ groups based on rotation angle with the y-axis. Front and mild side views are convenient for face fitting algorithms, extreme left and extreme right views are challenged for landmark detection, while back views are helpful with hair reconstruction.

\noindent \textbf{Accessory.} 
Parts of Asians (about $40$\%) are captured with special clothing and head accessories, while others are not, therefore, distributions of head accessories are only calculated among Asians, which is summarized in Figure~\ref{fig: statistic 2} (a). The high diversity of accessories types, materials and textures presents huge challenges for head rendering and reconstruction.

\noindent \textbf{Hair Style.} 
As shown in Figure~\ref{fig: statistic 2} (b), we have $7$ styles for wigs, $2$ with men’s styles, and $5$ with women’s styles. We randomly sampled about $10$ wigs for captured subjects, wig styles are not specified for gender. $6$ colors are not evenly distributed among each wig. Therefore, subjects captured with black and brown are the majority in our dataset, while yellow color has the least portion. Due to the hair-related benchmark, the complexity of hair structure and the dynamic deformation during large head motion challenge the SOTA methods, and the large hair assets provide a great database for the application of hair rendering and reconstruction as well as the potential research opportunity for cross-identity hairstyle transfer and animation.

\noindent \textbf{Corpus.} 
We calculated the word frequency for Chinese and English separately. From the cloud visualization, word frequency is indicated by the size of each character. The most frequent word ``Hai Pa'' in Chinese appears nearly $450$ times among all sentences, while the least frequent one ``Ji Jiu'' is less than 50. We only summarize the phrased in Chinese, but not single characters like ``de'', ``shi'', ``wo'' and \etc., since there have no specific implications. Among English, the most frequent word, ``Drawing'', occurs more than $600$ times, while the least frequent one ``Ambitious'' is close to 0. The corpus statistic and ``word cloud'' are demonstrated in Figure~\ref{fig: statistic 2} (c). Since our collection contains cross-identity repeated corpus and also different corpus, it is beneficial for the construction of the generalizable talking model.

\section{Benchmarks Details}
\label{sec:benchmark}
\begin{figure}[!htp]
    \centering
    \includegraphics[width=1.0\linewidth]{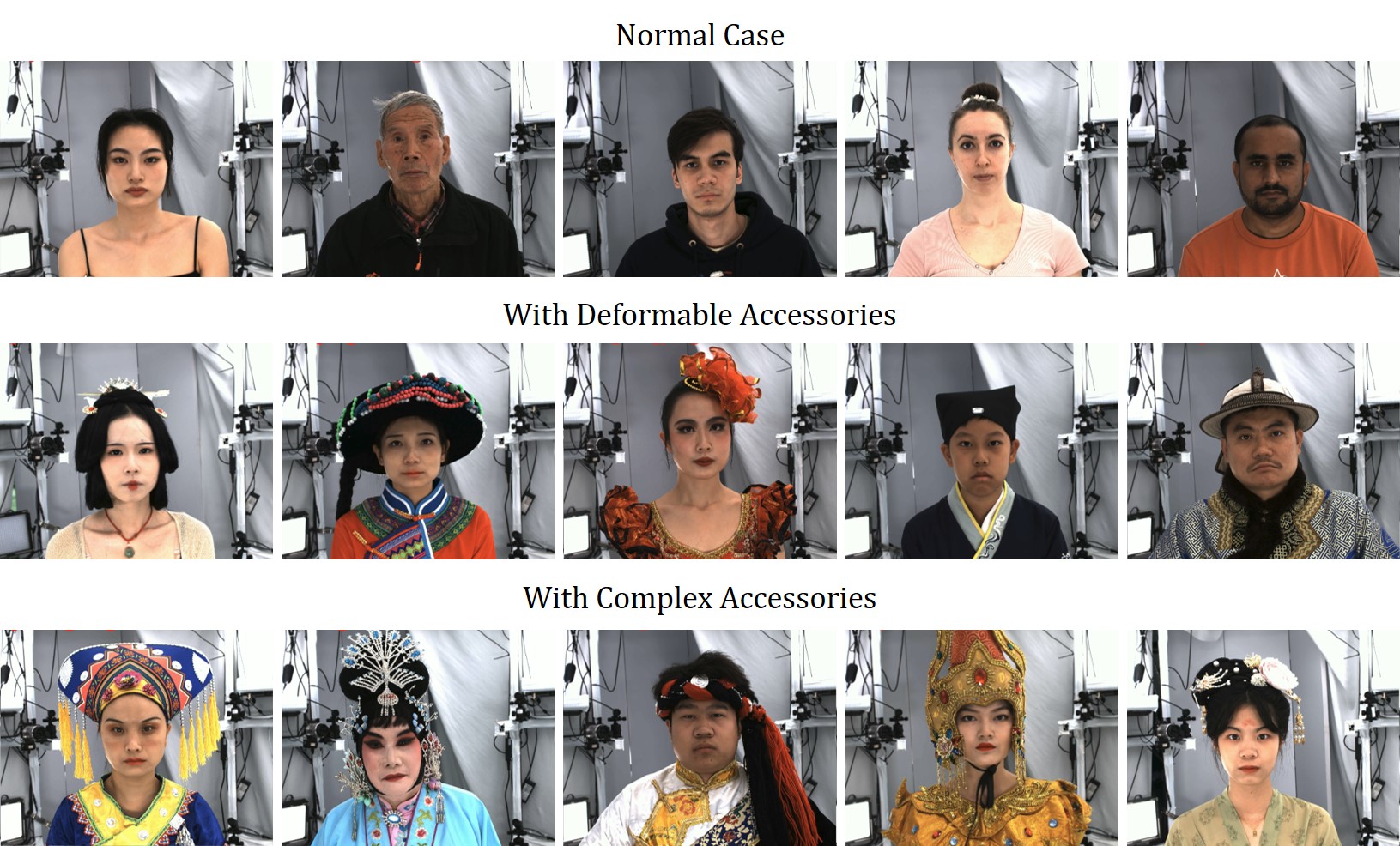}
    \caption{\textbf{Samples in Benchmark Splits.} We create three splits for benchmark evaluation, depending on the accessory difficulty, namely, `Normal Case', `With Deformable Accessories', and `With Complex Accessories'.}
\vspace{-0.5cm}
    \label{fig:benchmarks_preview}
\end{figure}

Based on the RenderMe-360 dataset, We construct a comprehensive benchmark on five critical tasks to showcase potential usage of our data, and reflect the status quo of relative methods. Due to the space limitation, some experiments and settings are not described in the main paper in detail. In this section, we introduce the criterion to divide our dataset splits in the first step. Then we provide a detailed discussion on benchmarks -- $1)$ We analyze the novel view synthesis benchmark with more qualitative results, and additional quantitative ablations. $2)$ We provide additional experiments in the novel expression synthesis benchmark with different training and testing settings from the main paper to value the state-of-the-art methods in more aspects. $3)$ We provide more qualitative visualizations for hair rendering benchmark.$4)$ For the hair editing benchmark, we provide qualitative results over different inversion methods, to serve as the complementary demonstrations to quantitative results shown in the main paper. $(5)$ For the talking head generation benchmark, we provide more experimental details.

\subsection{Benchmark Splits}

When it comes to rendering the human head, different attributes of head performance have a great impact on rendering tasks. For example, the high-frequency texture, detailed geometry, the reflection effects under different materials, and
the accessories which have different deformation caused by human head, all these factors are challenging and crucial for rendering tasks. To conduct a thorough evaluation of state-of-the-art methods, we split benchmark data for head-centric rendering tasks, with spanning difficulties in the hierarchy. Figure~\ref{fig:benchmarks_preview} shows a preview of split data samples. Concretely, we are following the defined rules to split data: $(1)$ Normal Case. Normal cases are identities without any accessories; $(2)$ With Deformable Accessories. Identities who wear deformable accessories, \eg, hair band, normal hat, \etc.$(3)$ With Complex Accessories. Identities have accessories with sophisticated structures or textures, \eg, gauze kerchiefs, complex earrings, or hats with pendants. For each task, we sample data from these three groups with different sampling principles, according to the characteristics of specific tasks. Please refer to the corresponding sections for more details.

\subsection{Novel View Synthesis}\label{nvs-subb}

\begin{figure*}
    \centering
    \vspace{-20pt}
    \includegraphics[width=1.0\linewidth]{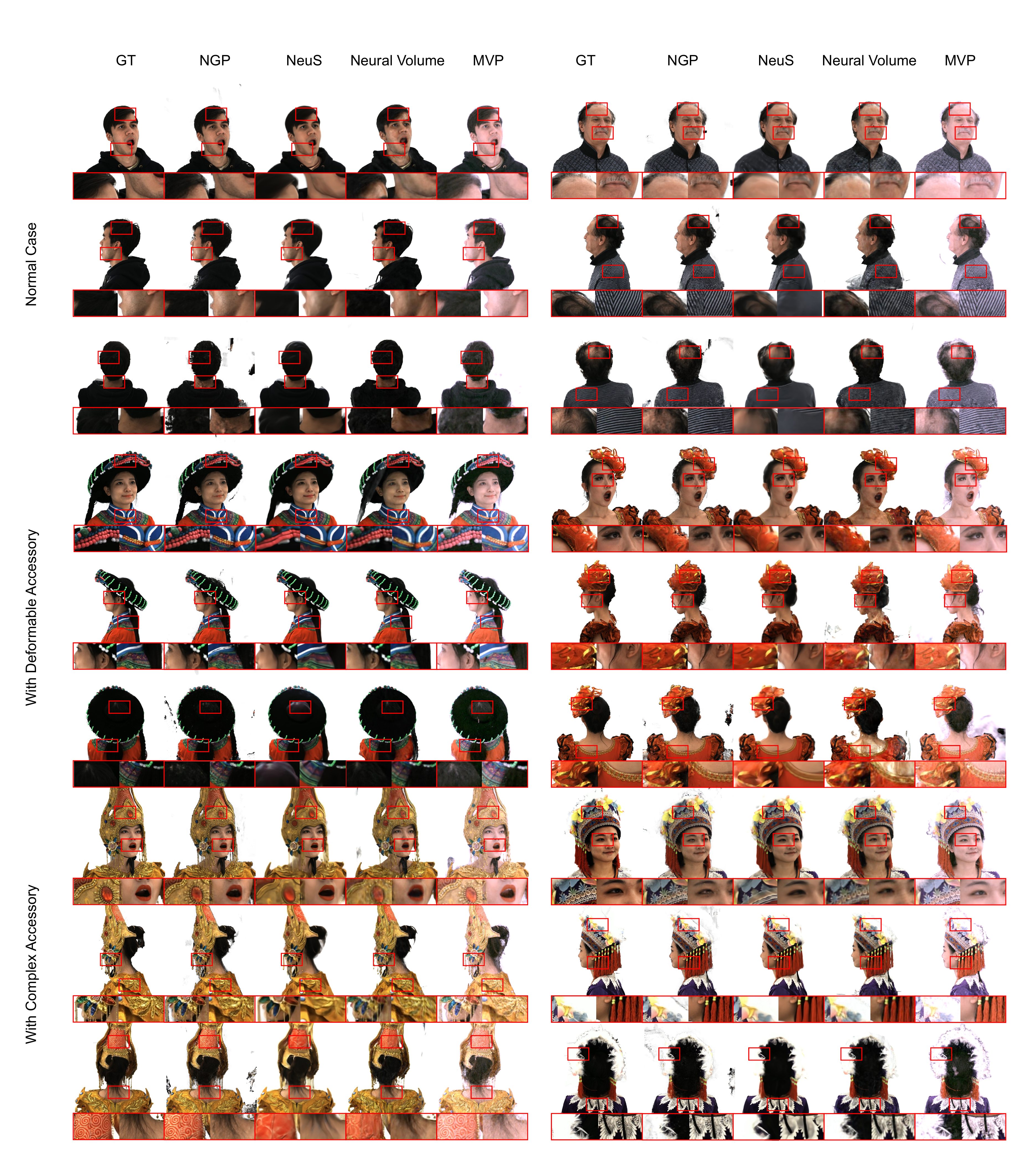}
    \caption{\textbf{Illustration of Qualitative Novel View Synthesis (\textit{\#Protocol-1}).} We sample two subjects in each data split and show the novel view synthesis results in three different test views (frontal, side, back) among four methods. NeuS performs well with almost no surrounding noise but has a much smoothing surface, while Instant-NGP produces a lot of surrounding noise and can recover some high-frequency parts. MVP renders lighter and more refined results, and Neural Volume renders skins mostly with many artifacts.}
    \vspace{-0.5cm}
    \label{fig:nvs}
\end{figure*}

\noindent
\textbf{Detailed Settings.}
As mentioned in the main experiment part, for {\textit{\#Protocol-1}} we evaluate the performance of novel view synthesis among four state-of-the-art methods. Specifically, we select two expressions from each subject, which means we train $40$ models for Instant-NGP~\cite{muller2022instant} and NeuS~\cite{wang2021neus} respectively, and $20$ models for MVP~\cite{lombardi2021mixture} and NV~\cite{lombardi2019neural} respectively.  Note that two expression sequences of one identity are trained with same configuration. For each model of Instant-NGP and NeuS,  we have $38$ camera view images for training and $22$ camera view images for testing, while the whole sequences of the selected expressions, which has in total about $8000$ frames of $38$ training views, are fed into the training of MVP and NV. For preprocessing, images are resized and matted to $512 \times 512$ with white background. Note that to get more stable rendering results, we do not resize the image and use a black background for Instant-NGP. We train $30k$ iterations for Intant-NGP to get sufficient convergence of the model, $200k$ iterations for MVP, and $50k$ iterations with batch size $16$ for NV. The other settings of these four methods are as same as the default implementations in~\cite{muller2022instant, wang2021neus, mvp, lombardi2019neural}.

\noindent
\textbf{Additional Qualitative Results.}
The qualitative result is shown in Figure~\ref{fig:nvs}, all four methods function normally in reconstructing the selected subjects, but with different performances. For the normal case, we mainly focus on high-frequency parts like hair and beard. As zoom-in pictures in the first three rows show, NeuS and Neural Volume can reconstruct the shape of the head and most facial features, but fail to render hair and beard in detail. Instant-NGP and MVP perform well in hairiness, which can be seen in relatively "high resolution", but there is still a gap between rendered image and ground truth. For the subjects with deformable accessories, we pay attention to the accessories with different textures. As demonstrated in the subject in the middle left, NeuS fails to reconstruct the bead-like shape of the fabric hat, smoothes and forms long stripes, indicating its disability to recover objects with complex textures. Focusing on the subject in the middle right, Neural Volume produces many artifacts in the neck, eyes and flower-like semi-transparent accessory. Finally, for the identities with complex accessories, we observe that Instant-NGP and MVP can render rigid or non-rigid accessories, like pendants, gemstones, feathers, and fabric slings, with high-frequency texture results. Scattered hair on the skin is failed to synthesize properly in all methods.

Generally speaking, difficulty of rendering increases when subjects wear complex accessories with complicated textures and various materials, however, there is no large gap in reconstructed results between these three test sets. When turning to a side-by-side comparison, Instant-NGP reconstructs the identity with a lot of surrounding noise, especially in the back views due to a relatively less proportion in the training, while rigid body accessories with high-frequency parts can be rendered well without much smoothing. Rendered results from NeuS almost have no noise and artifacts, showing superior performance than other methods, but NeuS fails to recover the most high-frequency parts in the head or accessories. Neural Volume shows lots of artifacts in the face or neck. By applying MVP, identities are reconstructed well with head and high-frequency accessories, but results are lighter compared to ground truth and ones from other methods. All methods can not handle complex textures completely, so their performance needs to be improved.

\begin{table}[htb]
\begin{center}
\resizebox{0.475\textwidth}{!}{
\begin{tabular}{c|c|cccc}
\toprule[1.5pt]
\textbf{Split} & \textbf{Metrics} & \textit{NGP}~\cite{muller2022instant} &\textit{NeuS}~\cite{wang2021neus} & \textit{NV}~\cite{lombardi2019neural} & \textit{MVP}~\cite{lombardi2021mixture} \\

\midrule
\multirow{3}{*}{\begin{tabular}{c} \textbf{Cam Split 0} \\\relax \makecell{\\} \textbf{[train 56, test 4]} \end{tabular}} &
   PSNR$\uparrow$      & \best{24.85} & 23.12 & \worst19.1 & 24.78 \\
~  & SSIM$\uparrow$    & 0.848 & 0.868 & \worst0.739 & \best{0.913} \\
~  & LPIPS$\downarrow$ & 0.23  & 0.16  & \worst0.3 & \best{0.12} \\
\midrule
\multirow{3}{*}{\begin{tabular}{c} \textbf{Cam Split 1} \\\relax \makecell{\\} \textbf{[train 38, test 22]} \end{tabular}} &
    PSNR$\uparrow$      & 21.5 & 23.37 & \worst{18.53} & \best{24.75} \\
~  & SSIM$\uparrow$    & 0.789 & 0.874 & \worst{0.699} & \best{0.922} \\
~  & LPIPS$\downarrow$ & \worst {0.37}  & 0.16  & 0.35 & \best{0.11} \\
\midrule
\multirow{3}{*}{\begin{tabular}{c} \textbf{Cam Split 2} \\\relax \makecell{\\} \textbf{[train 26, test 34]} \end{tabular}} &
   PSNR$\uparrow$      & 19.94 & \best{22.18} & \worst17.45 & 21.71 \\
~  & SSIM$\uparrow$    & 0.749 & \best{0.855} & \worst0.696 & 0.832 \\
~  & LPIPS$\downarrow$ & \worst{0.38}  & \best{0.17}  & 0.36 & 0.18 \\
\bottomrule[1.5pt]
\end{tabular}
}

\caption{\textbf{Ablation Study of Camera Split (\textit{\#Protocol-2})}. We set up the experiments with three camera splits and four methods.}
\label{tab:case-specific-novel-view-camera-ablation}
\vspace{-0.5cm}
\end{center}
\end{table}

\begin{figure*}[t]
    \centering
    \includegraphics[width=1.0\linewidth]{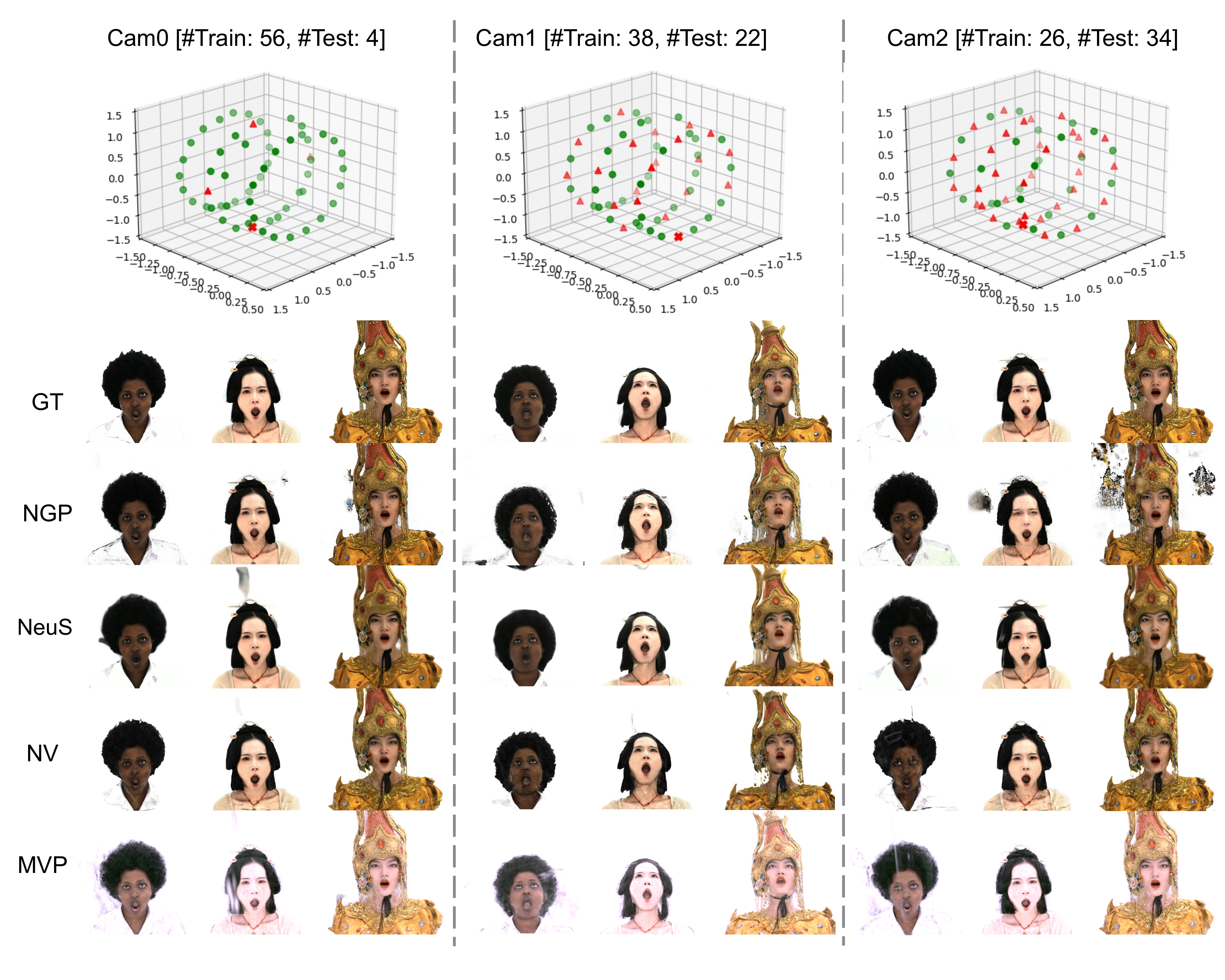}
    \vspace{-0.7cm}
    \caption{\textbf{Illustration of Camera Split Ablation (\textit{\#Protocol-2}).} We select and visualize three different camera settings, which are visualized on the top side of the figure. Green circles stand for training views, red triangles stand for testing views. We demonstrate three subjects in different data groups rendered with same expression. The visualized novel camera views are marked as {\color{Red}\protect\usym{2717}} in the camera split visualization.}
    \label{fig:case-specific-novel-view}
    \vspace{-0.2cm}
\end{figure*}

\subsubsection{Camera Split Ablation for Single ID NVS}\label{camera ablation-single}

\noindent
\textbf{Settings.}
For {\textit{\#Protocol-2}}, in order to do ablation experiment with various training and testing camera splits on rendering results, we design $3$ kinds of camera distribution and retrain the above methods, comparing the metrics. Three kinds of camera splits contain `train 56, test 4', which means most of the camera views are used in training, `train 38, test 22', which is the original distribution, `train 26, test 34', which means more testing views than training views, and all testing views in $3$ splits are uniformly distributed. We select $3$ representative subjects from the above-mentioned subset, and $1$ from each predefined split. The training settings are the same as in Section~\ref{nvs-subb}, except for the distribution of the training views.

\noindent
\textbf{Results.}
The quantitative result is shown in Table~\ref{tab:case-specific-novel-view-camera-ablation}. As the number of training views decreases, a decline in the metrics appears in Instant-NGP~\cite{muller2022instant}. Interestingly, when adding up the number of training views from $38$ to $56$, the performance of the other three methods remains roughly consistent, which indicates the number of training cameras above a certain threshold may not play a key role in performance. When we decrease the number of training views to $26$, all methods have a decline of metrics, and NeuS~\cite{wang2021neus} performs relatively better.

As the demonstration of qualitative result in Figure~\ref{fig:case-specific-novel-view}, there is no large gap in the visual result between Cam0 and Cam1 in all three subjects. For Instant-NGP~\cite{muller2022instant}, more details on accessories are reconstructed as more training views provided, while with fewer training views, more noise and artifacts occur on the face and the surrounding area.  For NV~\cite{lombardi2019neural}, artifacts also gets more when fewer views are involved into training , and it smooths the high-frequency details in all three settings. There is not much difference among three camera settings for MVP~\cite{lombardi2021mixture} and NeuS~\cite{wang2021neus}, but they fail to render high-frequency details with fewer training cameras and generate artifacts as well.

\begin{figure*}
    \centering
    \includegraphics[width=1\linewidth]{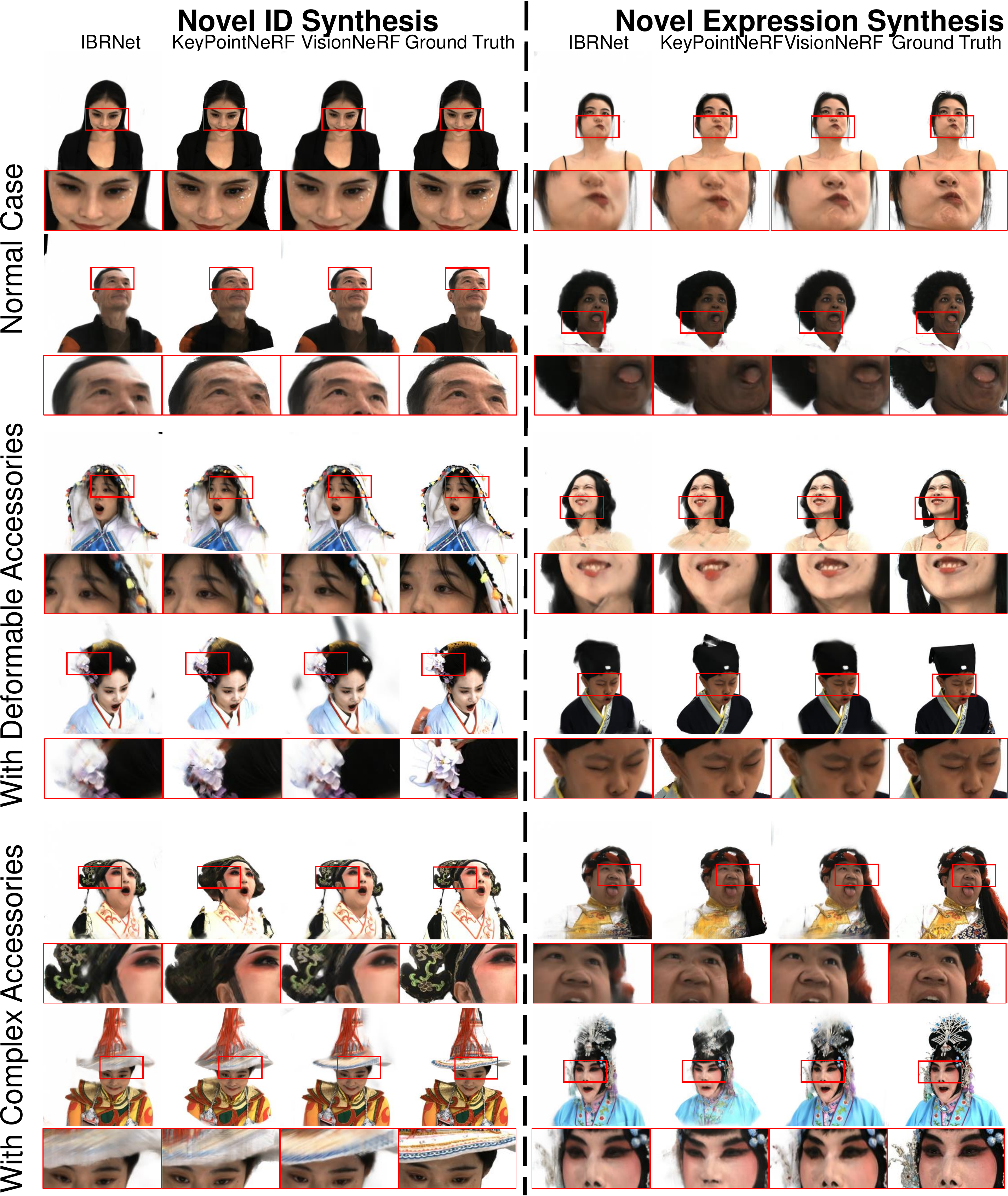}
    \caption{\textbf{Qualitative Results of Generalizable Novel View Synthesis (\textit{\#Protocol-1\&2}).} We illustrate some qualitative results of the generalizable methods, including IBRNet, KeypointNeRF, and VisionNeRF in two different settings, namely synthesizing the novel identifies and synthesizing the novel expressions. Two samples for a case are shown, and the regions in red boxes are zoomed in for better comparison. }
    \label{fig:generlization}
\end{figure*}

\subsubsection{Generalizable NVS}\label{sec:masked_generalizable}
\noindent \textbf{Detailed Settings.} As mentioned in the main paper, we train all models in both protocols with $10$ expressions performed by $187$ identities. For {\textit{\#Protocol-1}} we evaluate novel view synthesis on two unseen expressions on a subset of the training identities. Specifically,  we select $20$ identities in total -- $10$ normal cases, $5$ with deformable accessories, and $5$ with complex accessories. For {\textit{\#Protocol-2}}, $20$ unseen identities are tested with the same splitting strategy. Noted that during training and testing, three source views are used in all experiments, and we crop and resize the source and target views to the $512\times 512$ resolution, and render the images with white background.

\noindent \textbf{Additional Results.} Recall that, in the main paper, we find that KeypointNeRF~\cite{mihajlovic2022keypointnerf} achieves good visual quality while getting the worst quantitative results among all generalizable methods. We discuss the possible reasons behind the phenomenon in the main paper, where the major miss-alignment comes from the non-facial parts, like body parts of the rendered images(such as missing shoulders). Since KeypointNeRF~\cite{mihajlovic2022keypointnerf} tends to anchor the geometry using the relative encoding of facial key points, the body part with no keypoint encoding tends to reconstruct the intersection region from source views. Here, we further provided a quantitative demonstration from another perspective.  Concretely, we re-compute the benchmark results in Tab.~\ref{tab:mask_region} under a different masked region. In the main paper, we calculate metrics of rendered raw full images compared with ground truth. Here, in  Tab.~\ref{tab:mask_region}, we only calculate the regions that KeypointNeRF could render. As shown in the Table, The PSNR results of all methods get higher under this new setting, and KeypointNeRF~\cite{mihajlovic2022keypointnerf} outperforms IBRNet~\cite{wang2021ibrnet} and VisionNeRF~\cite{lin2023visionnerf} in SSIM and LPIPS, which accords with our visual observation.

\begin{table}[!htp]
\begin{center}
\resizebox{0.95\linewidth}{!}{
\begin{tabular}{c|c|c|ccc} \hline
\textbf{Train Setting} & {\textbf{Test Setting}}  
& {\textbf{Methods}}
& \multicolumn{1}{c}{PSNR$\uparrow$} & \multicolumn{1}{c}{SSIM$\uparrow$} & \multicolumn{1}{c}{LPIPS$\downarrow$*}\\ \hline
\multirow{6}*{Fixed Views} & \multirow{3}*{Fixed Views} &
IBRNet~\cite{wang2021ibrnet}& \worst23.70 & \worst0.889 & 135.16 \\
& & VisionNeRF~\cite{lin2023visionnerf} & 24.32 & 0.893 & \worst139.27 \\
& & KeypointNeRF~\cite{mihajlovic2022keypointnerf} & \best24.75 & \best0.901 & \best103.78\\
\cline{2-6}
& \multirow{3}*{Random Views} &
IBRNet~\cite{wang2021ibrnet} & \worst22.25 & \best0.895 & \worst157.96 \\
& & VisionNeRF~\cite{lin2023visionnerf} & \best22.58 & 0.874 & 157.54 \\
& & KeypointNeRF~\cite{mihajlovic2022keypointnerf} & 22.40 & \worst0.861 & \best143.265 \\
\hline
\multirow{6}*{Random Views} & \multirow{3}*{Fixed Views} &
IBRNet~\cite{wang2021ibrnet} & \worst24.84 & 0.903 & 102.57 \\
& & VisionNeRF~\cite{lin2023visionnerf} & \best25.80 & \worst0.902 & \worst118.72 \\
& & KeypointNeRF~\cite{mihajlovic2022keypointnerf} & 25.12 & \best0.910 & \best85.39 \\
\cline{2-6}
& \multirow{3}*{Random Views} & 
IBRNet~\cite{wang2021ibrnet}& 24.24 & 0.895 & 102.50  \\
& & VisionNeRF~\cite{lin2023visionnerf} & \worst23.11 & \worst0.879 & \worst149.62\\
& & KeypointNeRF~\cite{mihajlovic2022keypointnerf} & \best24.715 & \best0.890 & \best85.94\\
\hline
\end{tabular}
}
\vspace{-0.5cm}
\end{center}
\caption{\textbf{Masked Results on Generalizable NVS.} We re-calucuated the overall metrics on masked images in Tab~\ref{tab:unseen_person}.}
\vspace{-0.5cm}
\label{tab:mask_region}
\end{table}

\subsection{Novel Expression Synthesis}\label{nes-supp}

\noindent
\textbf{Additional Settings.}
As mentioned in the main experiment part, we evaluate the performance of novel expression synthesis among three state-of-the-art methods, namely NeRFace~\cite{Gafni_2021_CVPR}, IM Avatar~\cite{zheng2022avatar} and Point Avatar~\cite{zheng2022pointavatar}. Here we elaborately discuss the experiments for {\textit{\#Protocol-2}}, in which we select the same $20$ identities to form the benchmark data. We use $2$ sequences of verbal (about $1700$ to $2000$ frames) for training, another $1$ unseen verbal sequence and $11$ expression sequences (exclude the natural expression) for testing. All data samples used in \textit{\#Protocol-1\&2} are resized and matted to $512 \times 512$ with white background. We train $1000k$ iterations for NeRFace, $100$ epochs for IM Avatar, $65$ epochs for Point Avatar. We keep other training configurations the same as the default one, whose details are referred to ~\cite{Gafni_2021_CVPR, zheng2022avatar, zheng2022pointavatar}. All methods are evaluated in PSNR, SSIM, LPIPS, and L1 Distance, similar to ~\cite{zheng2022pointavatar}.

\noindent
\textbf{Additional Results.}
The quantitative result is shown in Table~\ref{tab:Case_Specific_Novel_Pose_Synthesis}. We find that Point Avatar~\cite{zheng2022pointavatar} achieves the best performance on the ~`Speech' set in terms of the average for `PSNR', `SSIM', `LPIPS', while NeRFace~\cite{Gafni_2021_CVPR} performs relatively better on the expression test data in total. Since the official implementation of IM Avatar is unstable in training, we can only show the results with the intermediate saved checkpoint. This contributes to IM Avatar's underperforming over other methods by a large margin. There exists a clear gap in the quantitative result between the speech and expression data in IM Avatar~\cite{zheng2022avatar} and Point Avatar~\cite{zheng2022pointavatar}. We attribute this difference to a different distribution of data. Since the speech data is mostly interpolation data, and the expression data tends to be extrapolation data. In addition, the qualitative result provides pieces of evidence from another perspective, which are shown in Figure~\ref{fig:nes}. IM Avatar collapses in the mouth parts and fails in detail synthesis (such as hair, and accessories). PointAvatar shows a high-quality performance in generating a 3D avatar, which reconstructs tiny strands of hair, while suffering from dynamic unseen expressions. NerFace also shows a strong ability to generate a 3D avatar that can extrapolate to simple unseen expressions. These methods all perform fine when interpolating into another verbal video, whereas struggle with extrapolation like Speech-to-Expression.

\begin{figure}[htb]
    \centering
    \includegraphics[width=0.95\linewidth]{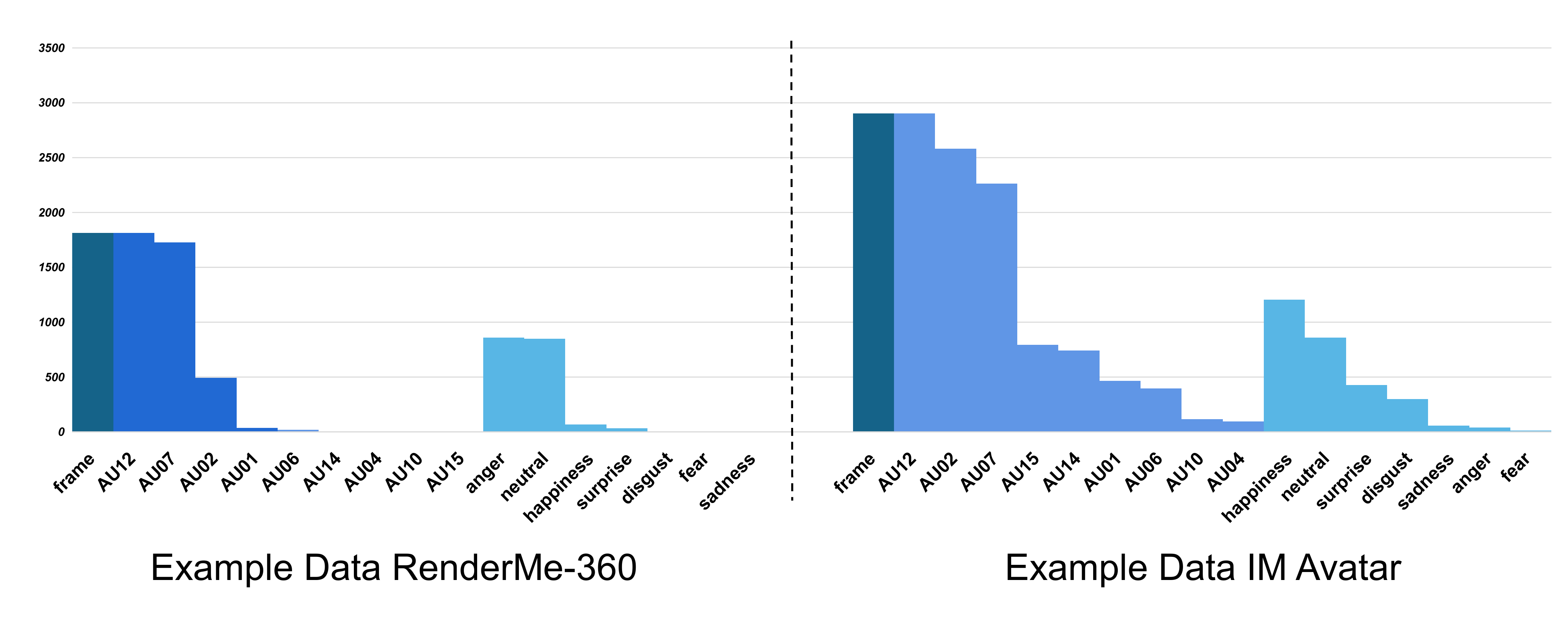}
    \caption{\textbf{Comparison of Training Data Between RenderMe-360 and IMavatar.} We summarize the frames, AUs, head poses, and expressions between the example data from RenderMe-360 and data from IMavatar.}
    \label{fig:nes_diff_au}
\end{figure}

\begin{figure}[htb]
    \centering
    \includegraphics[width=0.95\linewidth]{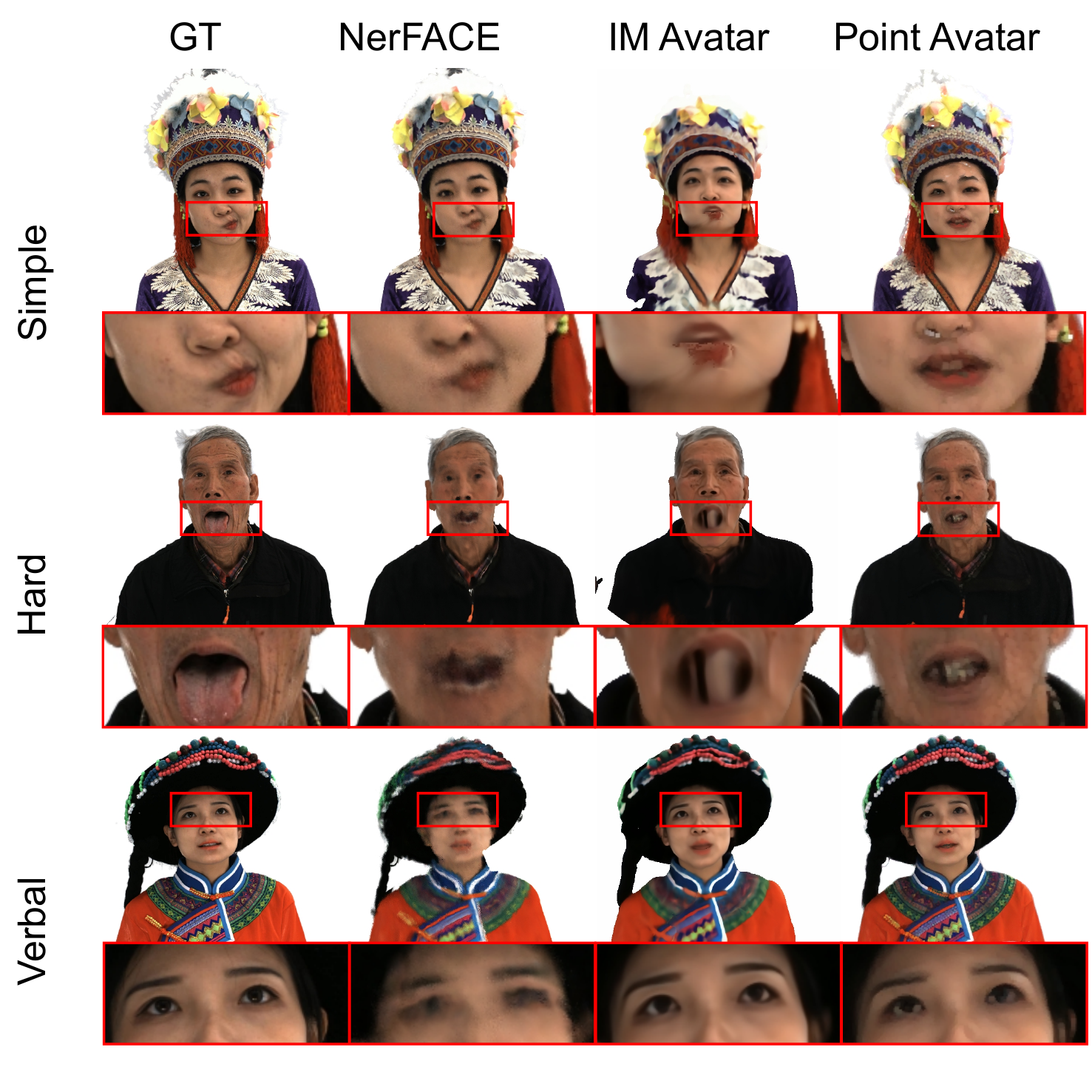}
    \caption{\textbf{Illustration of Novel Expression Synthesis (\textit{\#Protocol-2}).} We select three different identities from different levels of difficulty. The first line is the simple expression, the middle line is the hard expression and the last line is the interpolation result of another verbal video. }
    \label{fig:nes}
\end{figure}

We also perform the ablation experiments that trained with different FLAME fitting parameters, as shown in the last two rows of Table~\ref{tab:Case_Specific_Novel_Pose_Synthesis}. Specifically, DECA applies a model-based single-view fitting process, while our annotation pipeline designs a multi-view fitting process with the supervision of corresponding scan and images. We quantitatively compare the fitting quality, by calculating the facial landmark distance metric, which stands for the fitting error and reflects the quality of the expression parameters. For $99.3$\% of the data, the fitting result from our pipeline has better fitting quality. We further calculate the L2 difference of the shape parameter from the mean face to aligned identities, and obtain the result ($14.115$ in our pipeline, compared to $2.77$ from DECA). This phenomenon reflects that DECA tends to produce results converging to the mean face. 

\begin{table}[htb]
\begin{center}
\resizebox{0.475\textwidth}{!}{
\begin{tabular}{c|c|cccc}
\toprule[1.5pt]
\textbf{Method} & \textbf{Split} & $L_{1}$ $\downarrow$ &  PSNR $\uparrow$ & SSIM $\uparrow$ & LPIPS $\downarrow$ \\
\midrule
\multirow{3}{*}{\begin{tabular}{c} \textbf{NerFace~\cite{Gafni_2021_CVPR}} \end{tabular}} 
                  & \textbf{EN} & 0.0338 & 22.23 & 0.826 & 0.1264 \\
~                 & \textbf{EH} & 0.0369 & 21.4 & 0.815 & 0.1351 \\
~                 & \textbf{S} & 0.03 & 20.51 & 0.848 & 0.1499 \\
\midrule
\multirow{3}{*}{\begin{tabular}{c} \textbf{IM Avatar~\cite{zheng2022avatar}} \end{tabular}} 
                  & \textbf{EN} & 0.148 & \worst14.45 & 0.723 & 0.2751 \\
~                 & \textbf{EH} & \worst0.1522 & 14.5 & \worst0.718 & \worst0.2812 \\
~                 & \textbf{S} & 0.071 & 20.61 & 0.828 & 0.1754 \\
\midrule
\multirow{3}{*}{\begin{tabular}{c} \textbf{PointAvatar~\cite{zheng2022pointavatar} } \end{tabular}} 
                  & \textbf{EN} & 0.01 & 21.99 & 0.854 & 0.1097 \\
~                 & \textbf{EH} & 0.0103 & 21.83 & 0.852 & 0.1112 \\
~                 & \textbf{S} & \best0.0032 & \best26.95 & \best0.917 & \best0.0598 \\
\midrule
\multirow{3}{*}{\begin{tabular}{c} \textbf{PointAvatar~\cite{zheng2022pointavatar} } \\\relax \makecell{\\} (with DECA~\cite{DECA:Siggraph2021}) \end{tabular}} 
                  & \textbf{EN} & 0.0093 & 22.68 & 0.861 & 0.103 \\
~                 & \textbf{EH} & 0.0099 & 22.3 & 0.856 & 0.107 \\
~                 & \textbf{S} & 0.0034 & 26.83 & 0.914 & 0.0607 \\

\bottomrule[1.5pt]
\end{tabular}
}
\caption{\textbf{Novel Expression Synthesis (\textit{\#Protocol-2}).}  We evaluate three methods on the novel expression synthesis task on different splits of RenderMe-360. \textbf{EN}: Normal Expression, \textbf{EH}: Hard Expression, \textbf{S}: Speech.}
\vspace{-4ex}
\label{tab:Case_Specific_Novel_Pose_Synthesis}
\end{center}
\end{table}

\begin{figure}[htb]
    \centering
    \includegraphics[width=0.95\linewidth]{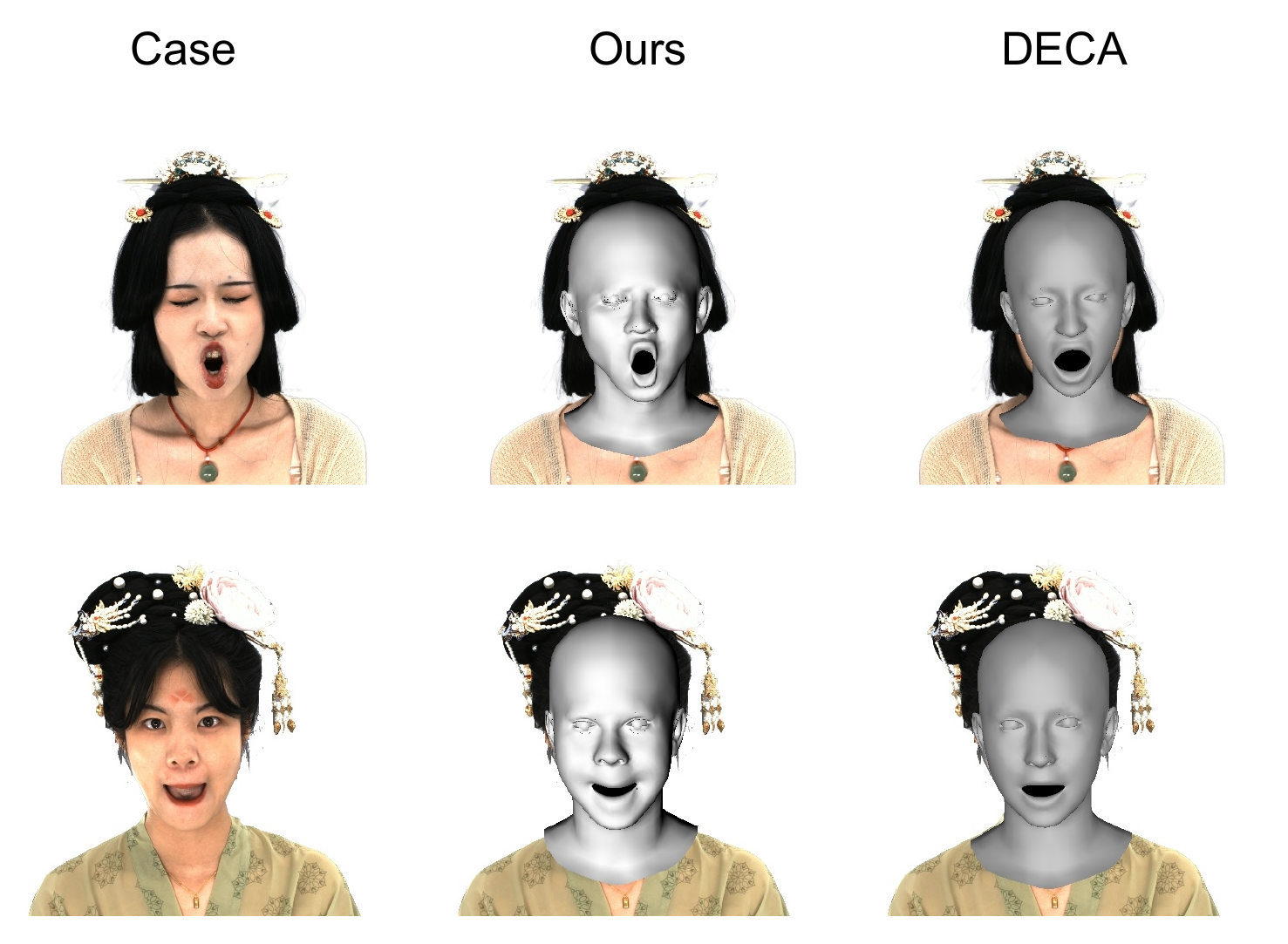}
    \caption{\textbf{Examples for Comparison of Different FLAME Fitting Quality.} We compare and visualize FLAME fitting results from RenderMe-360 and DECA. DECA is the processing pipeline of the official implementation of IM Avatar and Point Avatar.}
    \label{fig:nes_flame_compare}
    \vspace{-0.5cm}
\end{figure}

\begin{figure*}[htb]
    \centering
    \includegraphics[width=0.97\linewidth]{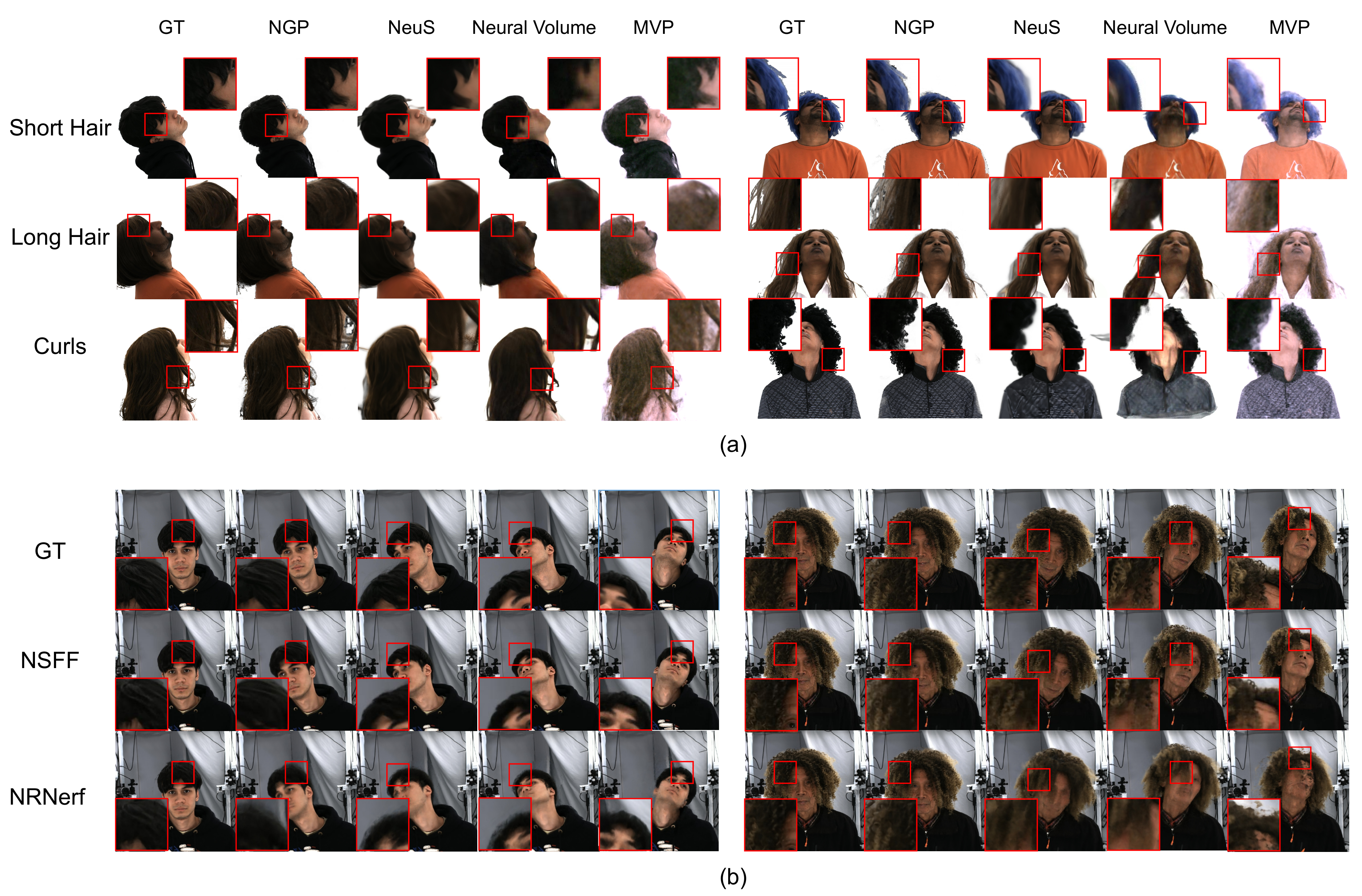}
    \vspace{-0.3cm}
    \caption{\textbf{Illustration of Hair Rendering.} (a) We show subjects in three kinds of hairstyles, and for the dynamic rendering methods (NV and MVP), we demonstrate the same frame as the static rendering methods. (b) We select keyframes of the sequence (novel inter-timestamp). Better zoom in for more details.}
    \label{fig:hair_render}
\end{figure*}

We further sample and visualize the FLAME result between two methods in Figure~\ref{fig:nes_flame_compare}. Our produced results mimic the motion of the mouth and eyes better, and cover richer details in geometry. Interestingly, a better FLAME fitting result does not contribute too much performance boost on Point-Avatar. As shown in the table, Point-Avatar trained with better FLAME parameters performs slightly better on the conversation sequences, but lags behind on intentional expression sequences. We guess the possible reason lies in the characteristics of the training and testing data. Compared with the training data used in the original paper (two of the subjects used in  Point-Avatar are from IMavatar's dataset), our conversation sequences are more challenging for Speech-to-Expression settings (\textit{i.e.,}, EN, EH in the Table ~\ref{tab:Case_Specific_Novel_Pose_Synthesis}). As shown in Figure~\ref{fig:nes_diff_au}, the facial attributes of our data are more challenging, as the main changes are around the mouth and fewer expressions pop up during the speech sequence. This leads to a larger distribution gap between training and testing scenarios. Moreover, since our FLAME pipeline produces better-aligned results in expression parts that are far away from the mean face (Figure~\ref{fig:nes_flame_compare}), the trained model struggles with these out-of-distribution cases, and has relatively lower metric performances than the ones trained on the FLAME version that is inaccurate but smooth across the sequence. 

\subsection{Additional Results of Hair Rendering}\label{hair-rendering-supp}

Following the discussion of hair rendering in the main part, we also demonstrate the qualitative result in Figure~\ref{fig:hair_render}. Figure (a) shows the visualization among methods under NVS track of static hair rendering and dynamic hair rendering. With the increase of the hair geometry complexity, we do not observe an obvious quality degradation of the hair rendering, while the corresponding metrics have a declining trend. We guess the main difference is on thin hair strand, which is the main challenge during hair rendering. As the complexity of the hairstyle increases, more hair strands spread out around the head (this can be discovered from the zoom-in area in the Figure), which are partially dismissed or smoothed during the rendering, causing degradation of metrics. Comparing the visualization of $4$ methods, we found some method-specific characteristics. Instant-NGP~\cite{muller2022instant} reconstructs the hair geometry not perfectly, but relatively well among four methods, since most of the diffusing hair strands can be reconstructed. We guess the multi-resolution data structure from NGP helps model the fine-grained geometry details. NeuS~\cite{wang2021neus} produces overall correct geometry, but strongly smooths the hair. Specifically, in the 'curls' scenario, all the curly hairs are smoothed to form a general shape, which losses edge details. This is reasonable, as the SDF-based representation has advantages in modeling single-contour objects, but struggles with multiple contours objects, especially with thin structures. Neural Volume~\cite{lombardi2019neural} produces lots of smoothness and blur, and most of the thin hair parts are dismissed, observed from the visualization. Since we feed the whole sequences with large motion into the model, it seems that Neural Volume can not handle this scenario. MVP~\cite{mvp} can preserve the hair details, but from all observed results, there are always artifacts surrounding the whole hair area. One possible reason is the size and quantity limitation of the volumetric primitives in the training procedure. As thin geometry, the hair parts need thousands of small primitives for high-quality representation, which requires great demands on training and is not training-friendly. A special primitive design is needed to be applied for hair rendering to improve performance.

In Figure (b) we show the time-interpolation results of two methods. NSFF~\cite{2021nsff} has better performance than NR-NeRF~\cite{2021nrnerf} in different hairstyles. For the head motion, NSFF preserves most of the strand details regardless of the motion blur, while NR-NeRF produces more blur and artifacts in the hair areas and face. The possible reason is that NSFF builds the structure correspondences among timestamps, which can be helpful for thin structure modeling. To improve the modeling capability of the deformable scenario, NR-NeRF introduces per-frame learned latent code, which may lead to smoothness and blurring with the interpolation of the latent code between two timestamps.

\subsection{Hair Editing}
\begin{figure*}[htb]
    \centering
    \includegraphics[width=1\linewidth]{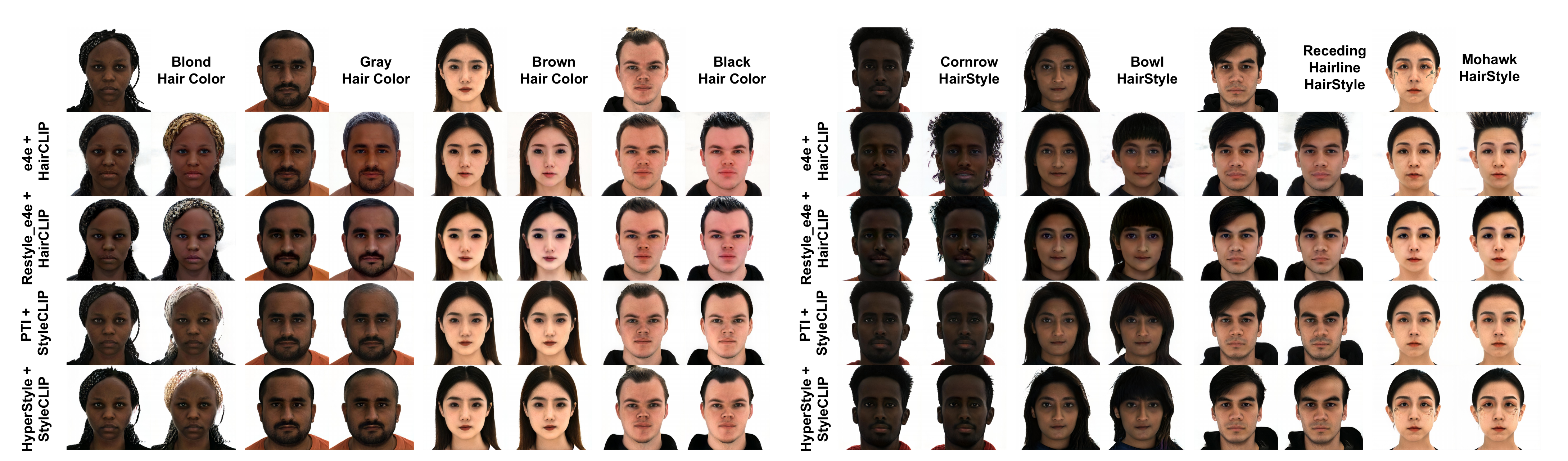}
    \caption{\textbf{Illustration of Qualitative Face Inversion and Hair Editing.} For each identity, we show the aligned face, the text reference, and four combinations of face inversion and further hair manipulation. The HairCLIP model works well for hair color and hairstyle manipulation, but identity maintenance needs to be improved. PTI and HyperStyle's inversion results are relatively satisfactory, but further StyleCLIP editing is worse than HairCLIP. }
    \label{fig:hair_clip}
\end{figure*}

\noindent
\textbf{Detailed Settings.} We first crop the original $2448\times2048$ images to $2048\times2048$ and then use the alignment code from PTI~\cite{roich2021pivotal} to do the crop and align. For the following HairCLIP and StyleCLIP editing with different inversion methods, we use open-source pre-trained models and inference code without any further training or fine-tuning. 
The reference text of hairstyle follows the definition of HairCLIP~\cite{wei2022hairclip}. 

\noindent
\textbf{Additional Qualitative Results.}
Figure~\ref{fig:hair_clip} shows the results of qualitative face inversion and hair manipulation on the normal split from the neutral expression subset of RenderMe-360. 
Based on the inversion results, PTI and HyperStyle can preserve more details such as face shape and hair texture compared to e4e and Restyle\_e4e, which is consistent with the metrics presented in Table ~\ref{tab:inversion}.
In terms of editing results, e4e+HairCLIP, which is specifically designed for hairstyle and hair color editing, performs well on both inputs. However, the other three methods may have some undesired editing effects when accepting arbitrary text inputs. For example, PTI+StyleCLIP may produce no hair when our reference text is gray hair, and HyperStyle+StyleCLIP may not generate desired cornrow hairstyles. 
In summary, the e4e+HairCLIP model has a good effect on hair editing, but identity maintenance limited by the inversion methods which needs to be improved. 
On the other hand, although the inversion results of PTI and HyperStyle are superior compared with e4e and Restyle\_e4e, the further text-based editing results following StyleCLIP are not equally satisfactory. 

\subsection{Talking Head Generation}
\noindent
\textbf{Detailed Settings.} Following AD-NeRF~\cite{guo2021ad}, we first convert videos to $450\times450$ resolution and we trim one second from the beginning and the end of each video to eliminate the interference from hitting board at the start and the end of recording. Then we use $90\%$ frames for training and the remaining for testing. We process each video segment separately, and the video data for each identity has an average length of 6,018 frames at 25 fps. 
To obtain more accurate training data, we utilize the landmark detection model from our data processing pipeline and use the same number of corresponding landmarks at the corresponding positions. Additionally, we use our own pipeline to obtain more accurate parsing results in the face parsing step. We utilize the open-source code of AD-NeRF and the code provided by the author of SSP-NeRF for training and testing. The results we present are generated by models trained for $400k$ iterations using the corresponding official default configurations.

\section{Applications of RenderMe-360}
There are a large number of down-streaming applications that could be enabled by our RenderMe-360 dataset, but have not been included in our current benchmark, such as $1)$ head generation, $2)$ image/video-based face reenact, and $3)$ cross-modal new avatar generation. Below, we demonstrate a specific task, Text to 3D Head Generation, which preliminarily reveals the broad possibilities of RenderMe-360 in abundant down-streaming applications.

\label{sec:application}
\begin{figure}
    \centering
    \includegraphics[width=1.0\linewidth]{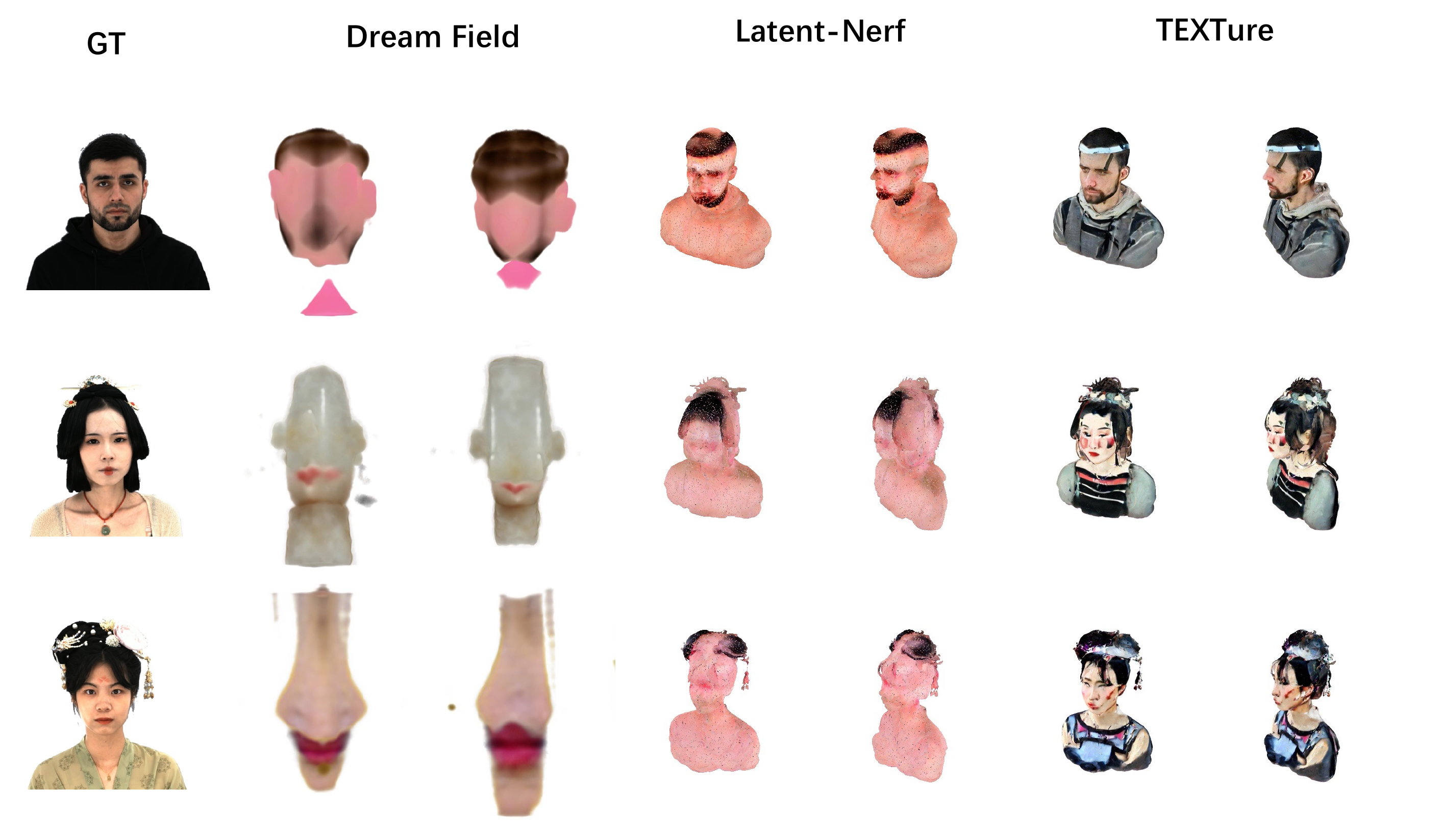}
    \caption{\textbf{Text-based Application.} We select three identities and generate the result with the same text prompt, while Latent-NeRF and TEXTure additionally use the scan as geometry prior. TEXTure performs best among these three methods, and the remaining two methods are not robust in human head scenarios. }
    \label{fig:application}
\end{figure}

\subsection{Text to 3D Head}

We apply our data on three typical Text to 3D Generation pipelines, \ie, Dream Fields ~\cite{jain2021dreamfields}, Latent-NeRF~\cite{metzer2022latent}, and TEXTure~\cite{richardson2023texture}. Although these methods are all general-object-centric, they are distinctive in different aspects. Specifically, Dream Fields uses NeRF to implicitly represent 3D object, and optimize the radiance fields with CLIP guidance. Latent-NeRF brings the NeRF into latent space, and guides the generation with both text and proxy geometry. TEXTure requires a precise mesh alongside the text prompt, to serve as input. It leverages a pre-trained depth-to-image diffusion to iteratively inpaint the 3D model.

 We select three identities from RenderMe-360 with different head characteristics. The first row in Figure~\ref{fig:application} is the simplest sample without any makeup or extra accessories. The second row is a bit complicated, we select it from the set ~'With Deformable Accessories'. The last row shows the sample in the most complicated set, in which we can see the subject has unique makeup and wears complex accessories. We use the corresponding text annotation of the samples to serve as the prompt input, which covers distinguishing descriptions of human heads in fine-grained details. We follow the original setting of the three methods, in which the scan annotation for each identity sample is used in Latent-NeRF and TexTure.
 
 As shown in Figure~\ref{fig:application}, TEXTure can generate more reasonable results than the other two methods. The reasons are two folds. First, it only needs to learn a representation that relates to texture, and geometrically wrap the texture into a 3D mesh to generate the 3D head. Second, it uses depth-to-image diffusion, which can generate high-quality 2D head images. In contrast, Dream Fields can not produce a complete 3D head with text prompt only. Latent-NeRF can not produce fine-grained texture, although it also uses geometry prior and text prompt as TEXTure. We infer that is because it cannot well embed the text prompt into the neural implicit rendering field during training. In a nutshell, this toy example showcases several interesting suggestions for future researches on Text-to-3D-Head: $1)$ With the rich annotations of RenderMe-360, it is possible to generate a high-fidelity 3D head avatar corresponding to text prompts. $2)$ There might be a bottleneck in using text to describe complex geometry, which might be one of the reasons why current text-to-3D paradigms struggle to generate realistic human-centric 3D targets. $3)$ As our data annotation covers multiple modalities and dimensions, it allows the researchers to explore new paradigms with different prompt conditions. 

\end{document}